\documentclass[11pt,a4paper]{article}
\usepackage[T1]{fontenc}
\usepackage[utf8]{inputenc}
\usepackage{lmodern}
\usepackage{amsmath,amssymb,amsthm}
\usepackage{authblk}
\usepackage[margin=27mm]{geometry}
\usepackage{booktabs,needspace}
\usepackage{array}
\usepackage{float}
\usepackage{tikz}
\usepackage{microtype}
\usepackage[colorlinks=true,linkcolor=blue,citecolor=blue,urlcolor=blue]{hyperref}
\hypersetup{
  pdftitle={Boundary Geometry Determines Optimal Shallow ReLU
    Approximation under Weighted Coefficient Constraints},
  pdfauthor={Jia-Qi Yang and Lei Shi}
}
\numberwithin{equation}{section}
\newtheorem{theorem}{Theorem}[section]
\newtheorem{lemma}[theorem]{Lemma}
\newtheorem{proposition}[theorem]{Proposition}
\newtheorem{corollary}[theorem]{Corollary}
\theoremstyle{definition}
\newtheorem{definition}[theorem]{Definition}
\theoremstyle{remark}
\newtheorem{remark}[theorem]{Remark}
\allowdisplaybreaks[2]
\title{Boundary Geometry Determines Optimal Shallow ReLU
Approximation under Weighted Coefficient Constraints$^\dag$\footnotetext{$\dag$~The work of Lei Shi is supported by the National Natural Science Foundation of China under Grant Nos.~12171093 and 12571099. Email addresses: jqyang24@m.fudan.edu.cn (J.-Q. Yang), leishi@fudan.edu.cn (L.~Shi). The corresponding author is Lei Shi. }}
\author{Jia-Qi Yang}
\author{Lei Shi}
\affil{School of Mathematical Sciences and Shanghai Key Laboratory for
	Contemporary Applied Mathematics, Fudan University, Shanghai 200433, China.}

\date{}
\begin{document}
\maketitle

\begin{abstract}
We study how boundary geometry affects $L^q$ approximation by shallow ReLU neural networks on convex bodies $K\subset\mathbb R^d$. The target classes are variation unit balls with weights $h_z\theta_z^\eta$, where $h_z$ and $\theta_z$ denote the activation cap height and relative volume. We establish upper and lower bounds for arbitrary convex bodies. For $2\le q<\infty$ and $\eta<1/q$, we determine the optimal polynomial approximation exponents for all planar convex bodies and their products with polytopes, including cubes and circular cylinders. For $C^2$ convex bodies, the classification holds for $\eta\le1/q$ in a dimension-dependent range of finite $q\ge2$ extending beyond $2$. When all principal curvatures are positive, the result extends to every finite $q\ge2$.  For these classes, a single geometric growth exponent associated with small boundary caps captures the dependence on $K$. We also identify sharp weight thresholds for retaining the classical approximation exponent and improve previous weights on Euclidean balls while preserving the classical $L^2$ approximation rate. For finite networks, the upper bounds are realized by compression using only original neurons, without increasing the weighted coefficient sum. These results show that boundary geometry can change the optimal exponent even when dimension, error norm, and weighting rule are fixed.
\end{abstract}

\noindent\textbf{Keywords:}
Neural networks;
Approximation rates;
Variation spaces;
Convex geometry.

\noindent\textbf{MSC codes:}
 41A25, 52A20, 68T07.
\section{Introduction}
\label{sec:introduction}
\subsection{Shallow Neural Network Approximation}
Shallow neural networks with activation function
$\sigma:\mathbb R\to\mathbb R$ are used in supervised learning
and form a class of approximants for multivariate functions.
For positive integers $d$ and $n$, let $\Sigma_n^d(\sigma)$
denote the class of functions on $\mathbb R^d$ represented
by networks with at most $n$ hidden neurons:
\begin{equation}\label{eq:intro-network-class}
 \Sigma_n^d(\sigma)
 =
 \left\{
   \sum_{j=1}^{n}a_j\sigma(\omega_j\cdot x+b_j):
   a_j,b_j\in\mathbb R,\ \omega_j\in\mathbb R^d
 \right\},
\end{equation}
with the convention that $\Sigma_0^d(\sigma)=\{0\}$.

Let $K\subset\mathbb R^d$, $d\ge2$, be a convex body,
that is, a compact convex set with nonempty interior,
and let $\mu_K=|K|^{-1}\,dx$ denote the normalized volume
measure on $K$.
For $1\le q<\infty$, let
$\mathcal F\subset L^q(\mu_K)$ be a bounded class of
functions to be approximated. Define its worst-case best
approximation error on $K$ by
\begin{equation}\label{eq:intro-best-error}
 \mathcal E_n^K(\mathcal F;\sigma,q)
 :=
 \sup_{f\in\mathcal F}
 \inf_{S\in\Sigma_n^d(\sigma)}
 \|f-S\|_{L^q(\mu_K)},
 \qquad n\ge0.
\end{equation}
This quantity measures how well networks with at most
$n$ hidden neurons can approximate the target class in the
worst case. To describe how this error decays as $n$ increases, define the optimal polynomial approximation exponent by \begin{equation}\label{eq:intro-optimal-exponent}
 r_*^K(\mathcal F;\sigma,q)
 :=
 \sup\left\{
 r\ge0:
 \sup_{n\ge0}(n+1)^r
 \mathcal E_n^K(\mathcal F;\sigma,q)<\infty
 \right\}.
\end{equation}
Early work on shallow neural network approximation derived upper 
bounds from a first-moment condition on the Fourier transform
\cite{Barron1993} and from Sobolev regularity
\cite{Mhaskar1996}.
The Fourier viewpoint extends to spectral Barron spaces,
with approximation rates depending on the smoothness index
\cite{SiegelXu2022Spectral}.


A class of target functions studied in nonlinear
approximation is the variation unit ball generated by
a bounded dictionary $\mathcal D\subset L^q(\mu_K)$
\cite{KurkovaSanguineti2001,SiegelXu2023Variation}.
It is defined by
\[
\mathbb B(\mathcal D)
 :=
 \overline{
 \left\{
   \sum_{i=1}^{N}c_i g_i:
   N\in\mathbb N,\ 
   g_i\in\mathcal D,\  c_i\in\mathbb R,\ 
   \sum_{i=1}^{N}|c_i|\le1
 \right\}
 }^{\,L^q(\mu_K)}.
\]
The number of terms $N$ is unrestricted,
in contrast to the prescribed approximation width $n$. The associated variation space is
$\bigcup_{R>0}R\,\mathbb B(\mathcal D)$. We refer to the elements of $\mathcal D$ as atoms.

For shallow neural networks, the dictionary can be chosen
as a bounded family of single-neuron ridge functions
$x\mapsto\sigma(\omega\cdot x+b)$ restricted to $K$.
Every $n$-term linear combination from such a dictionary
is then the restriction to $K$ of a network in
$\Sigma_n^d(\sigma)$.
Such $\ell^1$ constraints on the output coefficients also appear in convex formulations of shallow neural network learning, yielding
approximation and statistical estimation bounds \cite{Bach2017}.
Related variation spaces have been developed for nonlinear
operators between Hilbert spaces, providing target classes
for which approximation and generalization bounds have
been established \cite{YangShi2026}.

For general bounded dictionaries in $L^2(\mu_K)$,
Maurey's sampling argument gives the classical $n^{-1/2}$
approximation bound \cite{Pisier1981}.
For every $n\ge1$, each
$f\in\mathbb B(\mathcal D)$ admits an approximant
$f_n=\sum_{j=1}^{n}a_jg_j$, with $g_j\in\mathcal D$,
satisfying
\[
 \|f-f_n\|_{L^2(\mu_K)}
 \le C R_{\mathcal D}n^{-1/2},
 \qquad
 \sum_{j=1}^{n}|a_j|\le1,
\]
where $R_{\mathcal D}
=\sup_{g\in\mathcal D}\|g\|_{L^2(\mu_K)}$
and $C$ is an absolute constant.
The $n^{-1/2}$ rate is sharp for dictionaries formed
by orthonormal bases in infinite-dimensional Hilbert
spaces \cite{KurkovaSanguineti2001}.
Greedy approximation methods provide constructive
bounds of this order
\cite{Jones1992,DeVoreTemlyakov1996,
BarronCohenDahmenDeVore2008}. 

The general $n^{-1/2}$ approximation bound can be improved
by exploiting additional structure of neural network
dictionaries.
For certain sigmoidal activation functions, covering
estimates and randomized approximation yield the $L^2$
rate $n^{-1/2-1/(2d)}$ on the corresponding variation
unit balls \cite{Makovoz1996}. For the ReLU$^k$ activation $\sigma_k(s)=(s_+)^k$,
where $s_+=\max\{s,0\}$ and $k\ge1$ is an integer,
the variation unit ball generated by ridge atoms
$(\xi\cdot x-t)_+^k$, with $|\xi|=1$ and cutting hyperplanes intersecting $K$, admits the
$L^2(\mu_K)$ approximation rate
$n^{-1/2-(2k+1)/(2d)}$. This polynomial exponent is optimal
\cite{SiegelXu2024,Siegel2025}.
The same upper rate holds in $L^\infty(K)$
\cite{Siegel2025}.
For a target given by a finite expansion, the approximant
can be chosen using only the original atoms, without
increasing the absolute coefficient sum
\cite[Proposition~1]{Siegel2025}. 
In particular, for the ReLU variation unit ball $\mathcal F$,
the $L^2$ lower bounds and $L^\infty$ upper bounds yield $r_*^K(\mathcal F;\sigma_1,q)=(d+3)/(2d)$
for every $2\le q<\infty$.


The ReLU$^k$ variation space connects coefficient
constraints with function classes defined through the
Radon and Fourier transforms.
Its norm admits a Radon-BV characterization modulo
polynomials of degree at most $k$
\cite{OngieEtAl2020,ParhiNowak2021,SiegelXu2023Variation}.
The spectral Barron space of order $k+1$ embeds
continuously into this variation space
\cite{KlusowskiBarron2018,Xu2020,SiegelXu2024},
so the preceding upper approximation rates also hold
on the spectral Barron unit ball.

\subsection{Weighted Variation Spaces and Domain Geometry}
The classical ReLU variation unit ball uses the unweighted
coefficient constraint $\sum_{i=1}^{N}|c_i|\le1$.
Domain-dependent weights were introduced to incorporate
the geometry of $K$ into this constraint
\cite{DeVoreNowakParhiSiegel2025}. 

To describe the relevant geometry, define the support
function $H_K$ and the set $Z_K$ of admissible neuron
parameters by
\[
\begin{aligned}
 H_K(\xi)&:=\max_{x\in K}\xi\cdot x,\\
 Z_K&:=\left\{
 (\xi,t)\in\mathbb R^d\times\mathbb R:
 |\xi|=1,\ -H_K(-\xi)\le t<H_K(\xi)
 \right\}.
 \end{aligned}
\]
The set $Z_K$ parametrizes nonzero ReLU atoms whose cutting hyperplanes meet $K$. For $|\xi|=1$ and $(\xi,t)\notin Z_K$, the neuron output is either zero or affine on $K$. For $z=(\xi,t)\in Z_K$, define the ReLU atom and its
activation cap by
\[
\begin{aligned}
 \phi_z(x)&:=(\xi\cdot x-t)_+,\qquad x\in K,\\
 C_z&:=K\cap\{x:\xi\cdot x\ge t\}.
\end{aligned}
\]
The cap has height and relative volume
\[
h_z:=H_K(\xi)-t,
 \qquad
 \theta_z:=\mu_K(C_z)=\frac{|C_z|}{|K|}.
\]
Since $|\xi|=1$, $h_z$ is the distance between the
cutting hyperplane and the supporting hyperplane
$\xi\cdot x=H_K(\xi)$.
It also equals $\|\phi_z\|_{L^\infty(K)}$, and $\theta_z$
is the relative volume of the region where $\phi_z>0$.

On the Euclidean unit ball $B^d$, the weight introduced in \cite{DeVoreNowakParhiSiegel2025} is
\[
w_{\mathrm{prev}}(z)
 :=h_z^{(d+2)/4}=(1-t)^{(d+2)/4}.
\]
The corresponding weighted variation unit ball is
\[
 \mathcal B_{w_{\mathrm{prev}}}(B^d)
 :=
 \mathbb B\!\left(
 \left\{
 \frac{\phi_z}{w_{\mathrm{prev}}(z)}:
 z\in Z_{B^d}
 \right\}
 \right),
\]
where the closure is taken in $L^2(\mu_{B^d})$.
The associated weighted variation space is
$\bigcup_{R>0}R\mathcal B_{w_{\mathrm{prev}}}(B^d)$.\footnote{
Equivalently, the unit ball consists of Bochner integrals
$f=\int_{Z_{B^d}}(\phi_z/w_{\mathrm{prev}}(z))\,d\nu(z)$
in $L^2(\mu_{B^d})$, where $\nu$ is a finite signed Radon
measure of total variation at most one.
See \cite[Section~4, Eq.~(4.11)]{DeVoreNowakParhiSiegel2025}.
} 
By \cite[Theorems~5.2 and~6.1]{DeVoreNowakParhiSiegel2025},
for every $R>0$ and $n\ge1$, each
$f\in R\mathcal B_{w_{\mathrm{prev}}}(B^d)$ admits an approximant $S_n$ with at most $n$ ReLU atoms satisfying
\[
\|f-S_n\|_{L^2(\mu_{B^d})}
 \lesssim_d R n^{-(d+3)/(2d)}.
\]
The weighted space strictly contains the ordinary variation
space ($w\equiv1$), yet its unit ball retains the classical
$L^2$ approximation rate.

The constructions in \cite{DeVoreNowakParhiSiegel2025}
are adapted to the geometry of Euclidean balls and cubes.
The upper and lower bounds in
\cite{SiegelXu2024,Siegel2025} imply that, for $2\le q<\infty$, the unweighted ReLU variation unit ball
has optimal polynomial approximation exponent $(d+3)/(2d)$ on every convex body in $\mathbb R^d$. Keeping $d$ and $q$ fixed, we ask whether this independence
of domain shape persists for weighted variation unit balls
defined by a common weighting rule.
If not, which geometric quantities determine the optimal
exponent?
For a given domain, such a characterization would also
determine how far the coefficient constraint can be
relaxed while retaining the classical exponent.

The choice of weights is guided by the atom norm estimate
in Proposition~\ref{geom:atoms}:
\[
 \|\phi_z\|_{L^q(\mu_K)}
 \asymp_{d,q}h_z\theta_z^{1/q},
 \qquad 1\le q<\infty.
\]
For $\eta\in\mathbb R$, we consider the weights and
coefficient constraint
\[
 w_{K,\eta}(z)=h_z\theta_z^\eta,
 \qquad
 \sum_{i=1}^{N}|c_i|w_{K,\eta}(z_i)\le1.
\]
On $B^d$, the choice $\eta=(d-2)/(2(d+1))$
gives $w_{B^d,\eta}\asymp_d w_{\mathrm{prev}}$
(see Section~\ref{sec:comparison}). At $\eta=1/q$, the weight $w_{K,\eta}(z)$ is comparable to $\|\phi_z\|_{L^q(\mu_K)}$. The normalized atoms $\phi_z/w_{K,\eta}(z)$ form a bounded dictionary in $L^q(\mu_K)$ precisely when
$\eta\le1/q$. Within this range, increasing $\eta$ weakens the
coefficient constraint because $0<\theta_z\le1$. The corresponding variation unit balls therefore
form an increasing family.


\subsection{Main Contributions}

\noindent\emph{Geometric classification of optimal exponents.}
For $2\le q<\infty$ and $\eta<1/q$, we determine the
optimal polynomial approximation exponent for all planar
convex bodies, their products with fixed convex polytopes,
and invertible affine images of these domains
(Theorem~\ref{main:intrinsic}).
For convex bodies with $C^2$ boundary, the classification holds for $\eta\le1/q$ in an explicit dimension-dependent range of finite $q\ge2$ (Theorem~\ref{main:c2}).
If all principal curvatures are positive, the classification
holds for every $2\le q<\infty$ and $\eta\le1/q$
(Theorem~\ref{main:curved}).

In these cases, the dependence on $K$ is governed by
$\overline\kappa(K)$, the upper polynomial growth exponent
of the effective cap count at small volume scales
(see~\eqref{intr:eq-kappa}).
For example, when $d=3$, $q=2$, and $\eta=2/5$, the
optimal exponents are $1$ on a cube, $4/5$ on a
circular cylinder, and $7/10$ on a ball.
Table~\ref{tab:geometric-exponents} summarizes the
optimal exponents for representative convex bodies.

\noindent\emph{Preserving the classical exponent.}
The optimal exponent formulas determine exactly which
values of $\eta$ preserve the classical polynomial exponent.
On $B^d$, the classical $L^2$ approximation rate is
retained for a nonempty interval of $\eta$, while the
corresponding weights permit arbitrarily larger
coefficients on sufficiently small caps than under
$w_{\mathrm{prev}}$ (Section~\ref{sec:comparison}).

\noindent\emph{General bounds and network compression.}
We establish approximation upper bounds on arbitrary
convex bodies for $1\le q<\infty$ and $\eta\le1/q$
(Theorem~\ref{main:randomized}), and lower bounds for
$2\le q<\infty$ (Theorem~\ref{geo:interior-lower}).
We also obtain results of independent interest in convex geometry: bounds on the volumes of cap unions and the pointwise overlap of covering caps, with
constants depending only on $d$
(Propositions~\ref{rand:visibility}
and~\ref{rand:multiplicity}). Moreover, our approximation constructions compress
any given finite network to at most $n$ original atoms,
with adjusted coefficients and no increase in the
weighted coefficient sum.

The remainder of the paper is organized as follows.
Section~\ref{sec:setting} introduces the weighted variation
spaces and cap geometry, and Section~\ref{sec:main-results}
presents the main results.
Section~\ref{sec:comparison} discusses their consequences
and compares them with previous work, followed by an
overview of the proofs in Section~\ref{sec:proof-overview}.
The general upper bounds are proved in
Section~\ref{sec:randomized-proofs}.
Section~\ref{sec:lower-classification} establishes the lower
bounds and the results for $C^2$ convex bodies, while
Section~\ref{sec:geometric-proofs} develops the geometric
estimates needed for the remaining upper bounds.
Auxiliary arguments are collected in the appendix.

\section{Problem Formulation}
\label{sec:setting}



Subsection~\ref{Weighted spaces} gives the precise definitions
of the weighted variation spaces and approximation errors.
Subsection~\ref{sec:geometric-guide} introduces the geometric
quantities used in the main results.

\subsection{Approximation Errors and Weighted Variation Spaces}
\label{Weighted spaces}

Let $K\subset\mathbb R^d$, $d\ge2$, be a convex body,
and retain the notation from Section~\ref{sec:introduction}.
For $z=(\xi,t)\in Z_K$, recall
\begin{equation}\label{eq:domain-atoms}
 \begin{aligned}
 \phi_z(x)&=(\xi\cdot x-t)_+,
 & h_z&=H_K(\xi)-t,\\
 C_z&=K\cap\{x:\xi\cdot x\ge t\},\qquad\qquad\qquad
 & \theta_z&=\mu_K(C_z).
 \end{aligned}
\end{equation}
At $t=-H_K(-\xi)$, the atom is affine on $K$ and
$C_z=K$; at the excluded endpoint $t=H_K(\xi)$,
the atom would vanish identically. Write $X_q(K)=L^q(\mu_K)$ for $1\le q<\infty$ and
$X_\infty(K)=C(K)$ with the uniform norm.
When the domain is clear, $\|\cdot\|_q$ denotes the
corresponding norm.

For a positive weight $w:Z_K\to(0,\infty)$, define the
class of finite networks with weighted coefficient sum
at most one by
\[
 \mathcal A_w(K)
 =
 \left\{
   \sum_{i=1}^N c_i\phi_{z_i}:
   N<\infty,\ c_i\in\mathbb R,\ z_i\in Z_K,
   \sum_{i=1}^N|c_i|w(z_i)\le1
 \right\}.
\]

For each integer $n\ge0$, write
\begin{equation}\label{eq:network-class}
 \Sigma_n^K
 :=\{S|_K:S\in\Sigma_n^d(\sigma_1)\},
 \qquad \sigma_1(s)=s_+,
\end{equation}
where $\Sigma_n^d(\sigma)$ is defined in
\eqref{eq:intro-network-class}.
For $1\le q\le\infty$, define the worst-case approximation
error, allowing the value $+\infty$, by
\begin{equation}\label{eq:domain-error}
 \mathcal E_n^K(w;q)
 =
 \sup_{f\in\mathcal A_w(K)}
 \inf_{S\in\Sigma_n^K}\|f-S\|_{X_q(K)}.
\end{equation}
The target width $N$ is unrestricted, while $n$ bounds
the width of the approximant.
The infimum allows arbitrary neuron parameters.

Recall the weights
\begin{equation}\label{eq:intrinsic-weight}
 w_{K,\eta}(z)=h_z\theta_z^\eta.
\end{equation}
At $\eta=0$, the weight is the maximum amplitude of the atom:
$w_{K,0}(z)=h_z=\|\phi_z\|_\infty$.
Since $0<\theta_z\le1$, increasing $\eta$ decreases the
weights of atoms with small caps, allowing larger
coefficients under the same constraint.

By Proposition~\ref{geom:atoms}, with the convention
$1/\infty=0$, the normalized atoms
$\phi_z/w_{K,\eta}(z)$ are uniformly bounded in $X_q(K)$
if and only if $\eta\le1/q$.
Their $X_q(K)$ norms are comparable to
$\theta_z^{1/q-\eta}$, and hence tend to zero as
$\theta_z\downarrow0$ when $\eta<1/q$.
At the critical weight $\eta=1/q$,
$w_{K,1/q}(z)\asymp_{d,q}\|\phi_z\|_q$. For the weights $w_{K,\eta}$, the associated approximation
errors are also invariant under invertible affine changes
of variables (see Proposition~\ref{geom:atoms}).


For $1\le q\le\infty$ and $\eta\le1/q$, define the optimal
polynomial decay exponent of the approximation error by
\begin{equation}\label{intr:eq-order}
 r_*(K,\eta,q)
 =
 \sup\left\{r\ge0:
 \begin{array}{l}
 \mathcal E_n^K(w_{K,\eta};q)\le C_r(n+1)^{-r}
 \text{ for all }n\ge0\\
 \text{with some constant }C_r<\infty
 \end{array}
 \right\}.
\end{equation}
Here $C_r$ may depend on $K,\eta,q,r$, but not on $n$.
Multiplicative logarithmic factors do not affect $r_*$,
and the defining error bound need not hold with $r=r_*$.

We next introduce the weighted variation space generated
by the coefficient-constrained finite-network class
$\mathcal A_w(K)$.

\begin{definition}[Weighted variation spaces]
Let $1\le q\le\infty$ and let $w:Z_K\to(0,\infty)$
be a positive weight.
If the \emph{normalized dictionary}
$\{\phi_z/w(z):z\in Z_K\}$ is bounded in $X_q(K)$,
define the weighted variation unit ball by
\[
 \mathbb B_w^{(q)}(K)
 :=\overline{\mathcal A_w(K)}^{\,X_q(K)}.
\]
The associated \emph{weighted variation space} is
\begin{equation}\label{eq:space-definition}
 \mathcal V_w^{(q)}(K)
 :=\bigcup_{R>0}R\mathbb B_w^{(q)}(K),
\end{equation}
with norm
\begin{equation}\label{eq:variation-norm}
 \|f\|_{\mathcal V_w^{(q)}(K)}
 :=\inf\{R>0:f\in R\mathbb B_w^{(q)}(K)\}.
\end{equation}
\end{definition}

Equivalently, $\|f\|_{\mathcal V_w^{(q)}(K)}$ is the
infimum of $R>0$ for which $f$ is an $X_q(K)$ limit
of finite networks with representations satisfying
$\sum_i|c_i|w(z_i)\le R$.
The space $\mathcal V_w^{(q)}(K)$ is Banach, with
$\mathbb B_w^{(q)}(K)$ as its closed unit ball.
The completeness argument and the proof of the following
proposition are given in
Appendix~\ref{app:closure-allocation}.

\begin{proposition} 
\label{prop:closure-exponent}
Let $w:Z_K\to(0,\infty)$ and $1\le p\le q\le\infty$.
If the normalized dictionary
$\{\phi_z/w(z):z\in Z_K\}$ is bounded in $X_q(K)$, then
\[
 \mathbb B_w^{(p)}(K)=\mathbb B_w^{(q)}(K),
 \qquad
 \mathcal V_w^{(p)}(K)=\mathcal V_w^{(q)}(K),
\]
and the two variation norms agree.
For $q=\infty$ and $p<\infty$, these identities are
understood through the natural embedding
$C(K)\hookrightarrow L^p(\mu_K)$.
\end{proposition}

Thus, if the normalized dictionary is bounded in $X_q(K)$,
the closure defining $\mathbb B_w^{(q)}(K)$ agrees with
the $L^1(\mu_K)$ closure of $\mathcal A_w(K)$.
For such weights, Lemma~\ref{lem:closure} also shows that
the supremum in \eqref{eq:domain-error} is unchanged when
$\mathcal A_w(K)$ is replaced by $\mathbb B_w^{(q)}(K)$. For $1\le q<\infty$, the definitions in
\eqref{eq:intro-best-error} and
\eqref{eq:intro-optimal-exponent} give
\[
 \begin{aligned}
 \mathcal E_n^K(w;q)
 &=\mathcal E_n^K(\mathbb B_w^{(q)}(K);\sigma_1,q),\\
 r_*(K,\eta,q)
 &=r_*^K(\mathbb B_{w_{K,\eta}}^{(q)}(K);\sigma_1,q),
 \qquad \eta\le1/q.
 \end{aligned}
\]

\subsection{The Geometry of Small Caps}
\label{sec:geometric-guide}

The weight $w_{K,\eta}$ depends on the height and relative
volume of each individual activation cap. For the
approximation of sums of atoms, however, the distribution
of these caps across the boundary of $K$ also matters.
Caps of comparable small volume may occur in many different
directions, and their union can cover a much larger region
than any single cap. We begin by describing this collective
coverage through the wet part of $K$.

\begin{definition}[Wet part and floating depth]
\label{def:wet-part-depth}
The \emph{wet part} at threshold $\tau$, defined as the union
of all caps with $\theta_z\le\tau$, has normalized volume
\[
 W_K(\tau)=\mu_K\!\left(\bigcup_{\theta_z\le\tau}C_z\right).
\]
The \emph{floating depth} of $x\in K$ is the smallest relative
cap volume needed to contain $x$, expressed as
\[
 v_K(x)=\inf\{\theta_z:x\in C_z\}.
\]
\end{definition}

\begin{figure}[htbp]
\centering
\begin{tikzpicture}[
 x=1.3cm,y=1.3cm,font=\small,
 line cap=round,line join=round,>=stealth
]
 \begin{scope}
  \clip (0,0) circle (1.15);
  \fill[blue!15] (-1.2,.45) rectangle (1.2,1.2);
 \end{scope}

 \draw[thick] (0,0) circle (1.15);
 \draw (-1.30,.45)--(1.48,.45);
 \node[anchor=east] at (-1.40,.45) {$\xi\cdot x=t$};

 \draw[dashed] (0,1.15)--(1.48,1.15);
 \draw[<->] (1.40,.47)--(1.40,1.13);
 \node[right] at (1.43,.80) {$h_z$};

 \draw[->] (-1.32,.72)--(-1.32,1.13);
 \node[above] at (-1.32,1.13) {$\xi$};

 \node at (0,.74) {$C_z$};
 \node at (.10,-.01) {$\phi_z=0$};
 \node at (-.65,-.57) {$K$};
 \node at (0,-1.52) {(a) Activation cap};

 \begin{scope}[shift={(4.6,0)}]
  \fill[blue!15,even odd rule]
   (0,0) circle (1.15)
   (0,0) circle (.9);

  \draw[thick] (0,0) circle (1.15);
  \draw[dashed,black!60] (0,0) circle (.9);

  \foreach \a in {0,90,210} {
   \begin{scope}[rotate=\a]
    \draw[black!75] (-.715891,.9)--(.715891,.9);
   \end{scope}
  }

  \node[anchor=west,align=left] (wetlabel) at (1.48,.10)
   {Wet part\\at threshold $\tau$};
  \draw[->] (wetlabel.west)--(0.98,.22);

  \node at (0,-1.52) {(b) Wet part};
 \end{scope}
\end{tikzpicture}
\caption{Activation caps and the wet part.
In (a), $C_z$ has relative volume $\theta_z$, and its height
$h_z$ equals the maximum of $\phi_z$.
In (b), the shaded annulus is the union of all caps of the disk
with relative volume at most $\tau$.
Three representative cutting lines are shown.}
\label{fig:cap-intuition}
\end{figure}
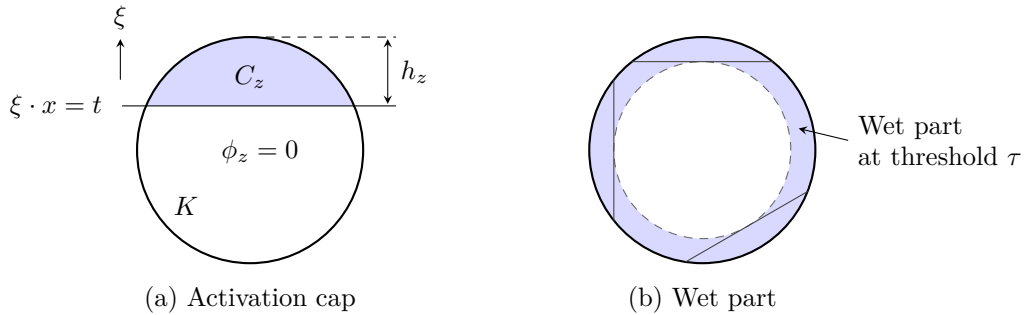

Caps in different directions reach different parts of the boundary,
so $W_K(\tau)$ can be much larger than $\tau$.
Figure~\ref{fig:cap-intuition} illustrates these geometric
objects for a disk: panel (a) shows an individual activation
cap, while panel (b) shows the wet part formed by caps in
different directions. The following lemma, proved in
Appendix~\ref{app:auxiliary-proofs}, shows that
a point belongs to the wet part at level $\tau$
if and only if its floating depth is at most $\tau$.

\begin{lemma}\label{lem:wet-part-depth}
For every $0<\tau<1$,
\[
 \bigcup_{\theta_z\le\tau}C_z=\{x\in K:v_K(x)\le\tau\},
 \qquad W_K(\tau)=\mu_K\{v_K\le\tau\}.
\]
\end{lemma}

Thus $W_K(\tau)$ is the distribution function of
$v_K$ with respect to the normalized volume measure $\mu_K$. To relate the volume of the wet part to the number of
caps at volume scale $\tau$, we introduce the ratio
\begin{equation}\label{intr:eq-profile}
 \mathcal C_K(\tau)=\frac{W_K(\tau)}{\tau},
 \qquad 1\le\mathcal C_K(\tau)\le\tau^{-1}.
\end{equation}
We call $\mathcal C_K(\tau)$ the \emph{effective cap count}.

For upper bounds on the approximation error, we use
Lemma~\ref{hil:economic-cover} to cover the wet part,
at each sufficiently small scale $\tau$,
by $N\asymp_d\mathcal C_K(\tau)$ possibly overlapping caps
of relative volume comparable to $\tau$.
For the corresponding lower bounds, cap packing estimates
relate $\mathcal C_K(\tau)$ to the number of caps of relative
volume $\tau$ with pairwise disjoint interiors
(see, e.g., Proposition~\ref{intr:planar-cover}). Moreover, an invertible affine map $T(x)=Ax+b$ maps caps
bijectively to caps and scales both cap and body volumes
by $|\det A|$.
It therefore preserves relative cap volumes, giving
$W_{T(K)}(\tau)=W_K(\tau)$ and
$\mathcal C_{T(K)}(\tau)=\mathcal C_K(\tau)$.

Covering and packing estimates relate $\mathcal C_K(\tau)$
to the number of caps at relative volume scale $\tau$.
For approximation, the key geometric information is
how rapidly this number grows as $\tau\downarrow0$.
We capture its polynomial growth by
\begin{equation}\label{intr:eq-kappa}
 \begin{aligned}
 \overline\kappa(K)
 &=\limsup_{\tau\downarrow0}
   \frac{\log\mathcal C_K(\tau)}{\log(1/\tau)},\\
 \kappa(K)
 &=\lim_{\tau\downarrow0}
   \frac{\log\mathcal C_K(\tau)}{\log(1/\tau)}
   \quad\text{when the limit exists}.
 \end{aligned}
\end{equation}

The upper growth exponent $\overline\kappa(K)$ enters
our approximation upper bounds for general convex bodies. In the classification results of Section~\ref{sec:main-results}, it determines the optimal polynomial approximation exponent together with $d$, $q$, and $\eta$. These results yield a geometric classification of
optimal approximation exponents for these classes of convex bodies.

The use of a limsup allows for growth rates that
do not converge. When the limit exists,
$\mathcal C_K(\tau)=\tau^{-\kappa(K)+o(1)}$.
Multiplicative logarithmic factors do not affect
these exponents.
By Lemma~\ref{rand:wet-bound},
\[
 0\le\overline\kappa(K)\le\frac{d-1}{d+1}.
\]

Representative values of $\kappa(K)$ and the corresponding
optimal approximation exponents are given in
Table~\ref{tab:geometric-exponents} in
Section~\ref{sec:main-results}.

\section{Main Results}
\label{sec:main-results}

We write $\alpha=(d+3)/(2d)$ for the classical ReLU
approximation exponent \cite{SiegelXu2024}.
For arbitrary convex bodies, we establish upper and lower
bounds on the approximation error, with the upper bounds
expressed in terms of the growth of the effective cap count.
We then give geometric classifications of optimal polynomial
approximation exponents for convex bodies with $C^2$
boundary, planar convex bodies, and products of planar
convex bodies with polytopes.

Under the hypotheses of Theorems~\ref{main:c2}, \ref{main:curved}, and~\ref{main:intrinsic},
the optimal exponent takes the common form
\[
r_*(K,\eta,q)
=
\min\left\{
\alpha,\,
1-\frac1q+\frac{1/q-\eta}{\overline\kappa(K)}
\right\},
\]
where a positive numerator divided by zero is
interpreted as $+\infty$.

Table~\ref{tab:geometric-exponents} summarizes the geometric
exponents $\kappa(K)$ and the optimal polynomial approximation
exponents $r_*(K,\eta,q)$ for representative domains.
All entries assume $2\le q<\infty$ and $\eta<1/q$.
Here $d=\dim K\ge2$, and $P\subset\mathbb R^m$, $m\ge1$,
is a convex polytope with nonempty interior.

\begin{table}[H]
\centering
\small
\setlength{\tabcolsep}{4pt}
\renewcommand{\arraystretch}{1.25}
\begin{tabular}{@{}lccc@{}}
\toprule
Domain $K$
& Dimension $d$
& $\kappa(K)$
& $r_*(K,\eta,q)$\\
\midrule
\begin{tabular}[c]{@{}l@{}}
Convex polytope\\[-2pt]
(including cubes)
\end{tabular}
& $d$
& $0$
& $\dfrac{d+3}{2d}$\\

Disk $B^2$
& $2$
& $1/3$
& $\min\left\{\dfrac54,\,
               1+\dfrac2q-3\eta\right\}$\\

Cylinder $B^2\times[-1,1]$
& $3$
& $1/3$
& $\min\left\{1,\,
               1+\dfrac2q-3\eta\right\}$\\

Product $B^2\times P$
& $m+2$
& $1/3$
& $\min\left\{\dfrac{m+5}{2(m+2)},\,
               1+\dfrac2q-3\eta\right\}$\\

Ball $B^d$
& $d$
& $\dfrac{d-1}{d+1}$
& $\min\left\{\dfrac{d+3}{2d},\,
               1+\dfrac{2}{(d-1)q}
                -\dfrac{d+1}{d-1}\eta\right\}$\\

\begin{tabular}[c]{@{}l@{}}
$C^2$ body with\\[-2pt]
positive curvature
\end{tabular}
& $d$
& $\dfrac{d-1}{d+1}$
& $\min\left\{\dfrac{d+3}{2d},\,
               1+\dfrac{2}{(d-1)q}
                -\dfrac{d+1}{d-1}\eta\right\}$\\
\bottomrule
\end{tabular}
\caption{Examples of geometric exponents and optimal
polynomial approximation exponents for $2\le q<\infty$
and $\eta<1/q$.}
\label{tab:geometric-exponents}
\end{table}

The geometric exponents for balls and $C^2$ bodies with
positive curvature follow from Theorem~\ref{main:c2},
and the approximation exponents follow from
Theorem~\ref{main:curved}.
Theorem~\ref{main:intrinsic} gives the entries for the
cylinder and $B^2\times P$.
For convex polytopes, \eqref{intr:eq-polytope-wet}
gives $\kappa(K)=0$, and the approximation exponent
$\alpha$ follows from Theorems~\ref{main:randomized}
and~\ref{geo:interior-lower}.

\subsection{General Bounds with Optimality for
\texorpdfstring{$C^2$}{C2} Bodies}

We begin with upper bounds for the worst-case approximation
error $\mathcal E_n^K(w_{K,\eta};q)$ on arbitrary convex
bodies $K\subset\mathbb R^d$ with $d\ge2$.
Theorem~\ref{main:randomized} gives a lower bound for
the decay exponent $r_*(K,\eta,q)$ in terms of the
weight parameter $\eta$ and the growth exponent
$\overline\kappa(K)$ of the effective cap count.

The general upper bound involves an exponent $\Gamma_d(q)$
that equals $1-1/q$ for $q\le Q_d$ and is smaller than
$1-1/q$ for $q>Q_d$.
The threshold $Q_d$ depends only on the dimension.
Specifically, set
\begin{equation}\label{main:eq-universal-parameters}
 Q_2=\infty,\qquad
 Q_d=2+\frac{4d}{(d+1)(d-2)}
 \quad(d\ge3).
\end{equation}
For $1\le q<\infty$, define
\[
 \Gamma_d(q)=
 \begin{cases}
  \displaystyle 1-\frac1q,
    & q\le Q_d,\\[2pt]
  \displaystyle\frac12+\frac{Q_d/2-1}{q},
    & q>Q_d.
 \end{cases}
\]
For example, $Q_3=5$ and $Q_4=18/5$.
In dimension two, the first branch applies to every
finite $q\ge1$.
\begin{theorem}[Upper bounds for general convex bodies]
\label{main:randomized}
Let $K\subset\mathbb R^d$ be a convex body with $d\ge2$,
and let $1\le q<\infty$. For $\eta<1/q$,
\begin{equation}\label{main:eq-universal-order}
 r_*(K,\eta,q)
 \ge
 \min\left\{
  \frac{d+3}{2d},\,
  \Gamma_d(q)+\frac{1/q-\eta}{\overline\kappa(K)}
 \right\},
\end{equation}
where a positive number divided by zero is interpreted
as $+\infty$.

If $1/q-\eta>
\overline\kappa(K)(\alpha-\Gamma_d(q))$,
the minimum in \eqref{main:eq-universal-order} equals
$\alpha$, and this exponent is attained without
logarithmic factors:
\begin{equation}\label{main:eq-universal-classical}
 \mathcal E_n^K(w_{K,\eta};q)
 \le C_{K,q,\eta}(n+1)^{-\alpha},
 \qquad \forall n\ge0.
\end{equation}

At the critical weight $\eta=1/q$, we have
$r_*(K,1/q,q)\ge\Gamma_d(q)$.
For $q<Q_d$, $\Gamma_d(q)=1-1/q$, and this exponent
is attained without logarithmic factors:
\begin{equation}\label{main:eq-universal-critical-rate}
 \mathcal E_n^K(w_{K,1/q};q)
 \le C_{d,q}(n+1)^{-(1-1/q)},
 \qquad \forall n\ge0.
\end{equation}

For a finite network with a given representation, these
upper bounds can be realized without introducing new atom
parameters or increasing the weighted coefficient sum (see Remark~\ref{rem:compression}).
\end{theorem}

The approximation upper bounds hold uniformly over the weighted variation unit balls.
The constant $C_{d,q}$ depends only on $d$ and $q$,
while $C_{K,q,\eta}$ may also depend on $K$ and $\eta$.
Both constants are independent of $n$. 
Proposition~\ref{rand:profile} gives more detailed estimates
that retain the dependence on the full function
$\mathcal C_K(\tau)$.


\begin{remark}[Compression of finite networks]
\label{rem:compression} 
The upper bounds above, as well as the upper bounds in the
classification results below, admit a stronger realization
for a finite network with a given representation.
Let
\[
 F=\sum_{i=1}^N c_i\phi_{z_i},
 \qquad
 M=\sum_{i=1}^N|c_i|w_{K,\eta}(z_i),
 \qquad z_i\in Z_K.
\]
Under the hypotheses of the corresponding result, the
approximant may be chosen as $S_n=\sum_{i=1}^N c_i'\phi_{z_i}$ with
\begin{equation}\label{main:eq-compression}
 \#\{i:c_i'\ne0\}\le n,
 \qquad
 \sum_{i=1}^N|c_i'|w_{K,\eta}(z_i)\le M.
\end{equation}
Thus no new atom parameters are introduced, and the
weighted coefficient sum does not increase.

For every explicit upper bound on
$\mathcal E_n^K(w_{K,\eta};q)$, the same bound holds for
$\|F-S_n\|_{L^q(\mu_K)}$ after multiplying its right-hand
side by $M$.
If a result gives $r_*(K,\eta,q)\ge R$ or
$r_*(K,\eta,q)=R$, then for every $0\le r<R$ one can choose
$S_n$ satisfying \eqref{main:eq-compression} such that
\[
 \|F-S_n\|_{L^q(\mu_K)}
 \le C_r M(n+1)^{-r}.
\]
The constants are independent of $n$, $N$, and the given
representation.
\end{remark}

Under the unweighted coefficient constraint
$\sum_i|c_i|\le1$, the optimal polynomial approximation
exponent in $L^q$ is $\alpha$ for $2\le q<\infty$
\cite{SiegelXu2024,Siegel2025}.
The next theorem extends the corresponding lower bound
to the classes defined by $w_{K,\eta}$ on arbitrary
convex bodies, showing that $\alpha$ remains an upper
limit on the optimal polynomial exponent.

\begin{theorem}[Lower bound for general convex bodies]
\label{geo:interior-lower}
Let $K\subset\mathbb R^d$ be a convex body with $d\ge2$,
let $2\le q<\infty$, and let $\eta\in\mathbb R$.
There exists a constant $c_{K,d,q,\eta}>0$ such that
\begin{equation}\label{geo:eq-interior-lower}
 \mathcal E_n^K(w_{K,\eta};q)
 \ge c_{K,d,q,\eta}
 \bigl((n+1)\log(n+2)\bigr)^{-\alpha},
 \qquad \forall n\ge0.
\end{equation}
In particular, for $\eta\le1/q$, we have $r_*(K,\eta,q)\le\alpha.$
\end{theorem}

For convex bodies with $C^2$ boundary, strict positivity
of curvature is not required: the geometric exponent
equals $(d-1)/(d+1)$ even when some principal curvatures
vanish.
The next theorem determines the optimal polynomial
approximation exponent for $2\le q\le Q_d$,
$q<\infty$, and $\eta\le1/q$.
\begin{theorem}[Optimal exponents for $C^2$ convex bodies]
\label{main:c2}
Let $K\subset\mathbb R^d$ be a convex body with
$C^2$ boundary and $d\ge2$.
Suppose that $2\le q\le Q_d$, $q<\infty$, and
$\eta\le1/q$.
Set $\kappa_d=(d-1)/(d+1)$.
Then $\kappa(K)=\kappa_d$, and
\begin{equation}\label{main:eq-c2-exponent}
 r_*(K,\eta,q)
 =
 \min\left\{
  \frac{d+3}{2d},\,
  1-\frac1q+\frac{1/q-\eta}{\kappa_d}
 \right\}.
\end{equation}
Moreover, the logarithmic decay rate converges:
\begin{equation}\label{main:eq-c2-log-limit}
 \lim_{n\to\infty}
 \frac{-\log\mathcal E_n^K(w_{K,\eta};q)}{\log n}
 =r_*(K,\eta,q).
\end{equation}
\end{theorem}

For fixed $d$, $q$, and $\eta$ in the stated range,
every convex body with $C^2$ boundary has the same
optimal polynomial approximation exponent as the
Euclidean ball.
The limit in \eqref{main:eq-c2-log-limit} also shows
that this exponent describes the logarithmic decay
of the error along the sequence $n\to\infty$.
\begin{remark}\label{rem:c2-extension}
Theorem~\ref{main:c2} also holds if $\partial K$
contains a nonempty relatively open $C^2$ subset
with all principal curvatures positive. No regularity is required on the rest of the boundary,
and the ranges of $q$ and $\eta$ remain unchanged. Examples include balls truncated by a hyperplane
and convex hulls of a ball and an exterior point. 
See Remark~\ref{rem:c2-local-extension} in
Subsection~\ref{hil:c2}.
\end{remark}

For convex bodies with $C^2$ boundary and positive
principal curvatures everywhere, the exponent formula
\eqref{main:eq-c2-exponent} and the logarithmic limit
\eqref{main:eq-c2-log-limit} hold for every finite
$q\ge2$ and $\eta\le1/q$.
The next theorem also gives explicit error bounds,
including at the critical weight $\eta=1/q$.
To state these bounds, set
\[
 a_q=\alpha-\left(1-\frac1q\right)
     =\frac{d+3}{2d}-1+\frac1q.
\]
When $a_q>0$, the two terms in
\eqref{main:eq-c2-exponent} coincide at
$1/q-\eta=\kappa_d a_q$.

\begin{theorem}[Optimal exponents for positively curved convex bodies]
\label{main:curved}
Let $K\subset\mathbb R^d$ be a convex body with $d\ge2$
and $C^2$ boundary, and suppose that all principal
curvatures are positive. Set $\kappa_d=(d-1)/(d+1)$. For every $2\le q<\infty$ and $\eta\le1/q$,
both \eqref{main:eq-c2-exponent} and
\eqref{main:eq-c2-log-limit} hold.

If $a_q>0$, then for any $n\ge0$,
\begin{equation}\label{main:eq-curved-rates}
 \mathcal E_n^K(w_{K,\eta};q)
 \le C_{K,q,\eta}
 \begin{cases}
  (n+1)^{-\alpha},
    & 1/q-\eta>\kappa_d a_q,\\[3pt]
  (n+1)^{-\alpha}[1+\log(n+1)]^{a_q},
    & 1/q-\eta=\kappa_d a_q,\\[3pt]
  (n+1)^{-(1-1/q)-(1/q-\eta)/\kappa_d},
    & 0\le1/q-\eta<\kappa_d a_q.
 \end{cases}
\end{equation}
If $a_q\le0$, the classical error bound holds for
every $\eta\le1/q$:
\begin{equation}\label{main:eq-curved-classical}
 \mathcal E_n^K(w_{K,\eta};q)
 \le C_{K,q,\eta}(n+1)^{-\alpha},
 \qquad \forall n\ge0.
\end{equation}

In particular, for every $2\le q<\infty$, the error
at the critical weight $\eta=1/q$ satisfies
\[
 \mathcal E_n^K(w_{K,1/q};q)
 \le C_{K,q}(n+1)^{-\min\{\alpha,1-1/q\}},
 \qquad \forall n\ge0.
\]

For a finite network with a given representation, the
upper bounds above can be realized with approximants
satisfying \eqref{main:eq-compression}.
\end{theorem}

When $a_q>0$, the optimal polynomial exponent equals
$\alpha$ for $1/q-\eta\ge\kappa_d a_q$.
For $0\le1/q-\eta<\kappa_d a_q$, the boundary term
gives a smaller exponent.
At $1/q-\eta=\kappa_d a_q$, the upper bound in
\eqref{main:eq-curved-rates} contains a logarithmic
factor.



\subsection{Optimal Exponents for Planar and Product Bodies}

We now consider planar convex bodies and their
products with polytopes, with no assumption of
boundary smoothness.
For $2\le q<\infty$ and $\eta<1/q$, the next theorem
determines the optimal polynomial approximation
exponent in terms of $\overline\kappa(K)$, without
assuming that $\kappa(K)$ exists.

\begin{theorem}[Optimal exponents for planar and product domains]
\label{main:intrinsic}
Let $K$ be a planar convex body or an invertible affine
image of $D\times P$, where $D\subset\mathbb R^2$
is a convex body and \(P\subset\mathbb R^m\), \(m\ge1\), is a polytope with nonempty interior. Set $d=\dim K$.
For $2\le q<\infty$ and $\eta<1/q$,
\begin{equation}\label{main:eq-intrinsic-order}
 r_*(K,\eta,q)
 =
 \min\left\{
  \frac{d+3}{2d},\,
  1-\frac1q+\frac{1/q-\eta}{\overline\kappa(K)}
 \right\},
\end{equation}
where the quotient is interpreted as $+\infty$
when $\overline\kappa(K)=0$.

If $1/q-\eta>
\overline\kappa(K)\max\{\alpha-1+1/q,0\}$,
the classical exponent is attained without
logarithmic factors:
\begin{equation}\label{main:eq-intrinsic-classical}
 \mathcal E_n^K(w_{K,\eta};q)
 \le C_{K,q,\eta}(n+1)^{-\alpha},
 \qquad \forall n\ge0.
\end{equation}

If $\kappa(K)$ exists, then
$\overline\kappa(K)=\kappa(K)$ and
\[
 \lim_{n\to\infty}
 \frac{-\log\mathcal E_n^K(w_{K,\eta};q)}{\log n}
 =r_*(K,\eta,q).
\]

If $K$ is an invertible affine image of $D\times P$,
then $\overline\kappa(K)=\overline\kappa(D)$.
Moreover, $\kappa(K)$ exists if and only if
$\kappa(D)$ exists, in which case
$\kappa(K)=\kappa(D)$.

For a finite network with a given representation, the
upper bounds above can be realized with approximants
satisfying \eqref{main:eq-compression}.
\end{theorem}

In the product case,
$\overline\kappa(K)=\overline\kappa(D)$,
so the geometric growth exponent is determined by
the planar factor $D$.
The classical exponent $\alpha=(d+3)/(2d)$, however,
depends on the full dimension $d=m+2$.
Compared with the general upper bound in
Theorem~\ref{main:randomized}, the boundary term in
\eqref{main:eq-intrinsic-order} has $1-1/q$ in place
of $\Gamma_d(q)$ throughout $2\le q<\infty$.
For $d\ge3$ and $q>Q_d$, we have
$1-1/q>\Gamma_d(q)$.

For planar convex bodies, Lemma~\ref{rand:wet-bound}
and the inequality $\mathcal C_K(\tau)\ge1$ give
$0\le\overline\kappa(K)\le1/3$.
The polytope estimate \eqref{intr:eq-polytope-wet}
implies $\kappa(K)=0$ for polytopes, including polygons.
Theorem~\ref{main:c2} gives
$\kappa(B^d)=(d-1)/(d+1)$, so the disk satisfies
$\kappa(B^2)=1/3$. 




To isolate the effect of domain geometry,
Table~\ref{tab:geometry-comparison} compares three
domains at fixed $d$, $q$, and $\eta$.
\begin{table}[H]
\centering
\begin{tabular}{@{}lcc@{}}
\toprule
Domain $K$ & $\kappa(K)$ & $r_*(K,2/5,2)$\\
\midrule
Convex polytope (including the cube) & $0$ & $1$\\
Circular cylinder $B^2\times[-1,1]$ & $1/3$ & $4/5$\\
Ball $B^3$ & $1/2$ & $7/10$\\
\bottomrule
\end{tabular}
\caption{Optimal exponents for $d=3$, $q=2$, and $\eta=2/5$.}
\label{tab:geometry-comparison}
\end{table}
Thus the optimal polynomial exponent can differ
even when the dimension, error norm, and weighting
rule are fixed.

\section{Discussion}
\label{sec:comparison}

The preceding results connect boundary geometry to
both optimal polynomial approximation exponents and
the extent to which weighted coefficient constraints
can be relaxed.
We discuss the role of local curvature, quantify the
improvement over previous weights on the ball, and
consider the implications for network compression
and the classification of general convex bodies.

\paragraph{\normalfont\itshape Can a curved boundary patch determine the optimal exponent?}

A positively curved portion of the boundary can determine
the optimal polynomial exponent even when other parts
of the boundary are flat or nonsmooth.
If $\partial K$ contains a $C^2$ patch
(a nonempty relatively open subset of $\partial K$)
on which all principal curvatures are positive,
Remark~\ref{rem:c2-extension} gives
$\kappa(K)=\kappa_d=(d-1)/(d+1)$.
For $2\le q\le Q_d$, $q<\infty$, and $\eta\le1/q$,
\[
 r_*(K,\eta,q)=r_*(B^d,\eta,q)
 =\min\left\{
   \alpha,\,
   1-\frac1q+\frac{1/q-\eta}{\kappa_d}
 \right\}.
\]
This applies, for example, to the half-ball
$B^d\cap\{x_d\ge0\}$, whose spherical portion
meets a flat face nonsmoothly.

For every $d$-dimensional convex polytope $P$,
\eqref{intr:eq-polytope-wet} gives $\kappa(P)=0$.
In contrast, a single positively curved $C^2$ patch suffices to ensure $\kappa(K)=\kappa_d$,
the maximal value allowed by
Lemma~\ref{rand:wet-bound}. Flat portions elsewhere on the boundary cannot
reduce this exponent.

The effect of this geometric difference on the optimal
approximation exponent depends on the weight parameter
$\eta$.
Fix $2\le q\le Q_d$ with $q<\infty$, and let
$\eta<1/q$.
Theorems~\ref{main:randomized}
and~\ref{geo:interior-lower} give
$r_*(P,\eta,q)=\alpha$ for every $d$-dimensional
convex polytope $P$.
If $\eta\le1/q-\kappa_d a_q$, then
$r_*(K,\eta,q)=r_*(P,\eta,q)=\alpha$.
If $\eta>1/q-\kappa_d a_q$, then
$r_*(K,\eta,q)<r_*(P,\eta,q)=\alpha$.
Thus a single positively curved patch suffices to
lower the optimal polynomial exponent relative to
a polytope once the weighted coefficient constraint
is relaxed beyond this threshold.

\paragraph{\normalfont\itshape How far can the coefficient constraint be relaxed?}

Increasing $\eta$ in
$w_{K,\eta}(z)=h_z\theta_z^\eta$ reduces the weights
of atoms supported on small caps, allowing larger
coefficients under the same weighted budget.
Theorems~\ref{main:c2}, \ref{main:curved}, and
\ref{main:intrinsic} determine how far this constraint
can be relaxed while retaining the classical
polynomial exponent $\alpha$.
Within their respective domain and parameter ranges,
if $a_q=\alpha-1+1/q>0$, then
\begin{equation}\label{main:eq-critical-weight}
 r_*(K,\eta,q)=\alpha
 \quad\Longleftrightarrow\quad
 \eta\le\eta_{\rm fast}(K,q)
 :=\frac1q-\overline\kappa(K)a_q.
\end{equation}
The strict condition $\eta<1/q$ in
Theorem~\ref{main:intrinsic} remains in force,
including when $\overline\kappa(K)=0$ and
$\eta_{\rm fast}(K,q)=1/q$.
If $a_q\le0$, then $r_*(K,\eta,q)=\alpha$
for all $\eta\le1/q$ under
Theorem~\ref{main:curved}, and for all
$\eta<1/q$ under Theorem~\ref{main:intrinsic}.

For fixed $d$ and $q$ with $a_q>0$, a larger
geometric exponent $\overline\kappa(K)$ gives a
smaller threshold $\eta_{\rm fast}(K,q)$.
Thus faster polynomial growth of the effective cap
count permits less relaxation of the coefficient
constraint while retaining the classical exponent.
In particular, for $q=2$,
\begin{equation}\label{discussion:eq-l2-threshold}
 \eta_{\rm fast}(K,2)
 =\frac12-\frac{3}{2d}\overline\kappa(K).
\end{equation}
Since $\overline\kappa(K)\le\kappa_d$, convex bodies
with a positively curved $C^2$ boundary patch
attain the smallest $L^2$ threshold among the
classified domains in each fixed dimension.

\paragraph{\normalfont\itshape Improved weights on Euclidean balls.}

For $K=B^d$, $d\ge2$, and $q=2$, the previous
weight $w_{\rm prev}(z)=h_z^{(d+2)/4}$ yields the
classical approximation rate in
\cite[Theorems~5.2 and~6.1]{DeVoreNowakParhiSiegel2025}.
On the ball, the relative cap volume satisfies
$\theta_z\asymp_d h_z^{(d+1)/2}$, so
$w_{\rm prev}\asymp_d w_{B^d,\eta_{\rm prev}}$. 
The corresponding parameter $\eta_{\rm prev}$ and
the threshold $\eta_{\rm fast}=\eta_{\rm fast}(B^d,2)$
from \eqref{discussion:eq-l2-threshold} satisfy
\begin{equation}\label{main:eq-weight-comparison}
 \begin{aligned}
  \eta_{\rm prev}
  &=\frac{d-2}{2(d+1)},&
  \eta_{\rm fast}
  &=\frac{d^2-2d+3}{2d(d+1)},\\
  \eta_{\rm fast}-\eta_{\rm prev}
  &=\frac{3}{2d(d+1)}>0.
 \end{aligned}
\end{equation}
For every $\eta_{\rm prev}<\eta<\eta_{\rm fast}$,
Theorem~\ref{main:curved} gives
\[
 \mathcal E_n^{B^d}(w_{B^d,\eta};2)
 \le C_{d,\eta}(n+1)^{-\alpha},
 \qquad n\ge0.
\]
Thus the classical error rate is retained with
weights that are asymptotically smaller on shallow caps.

Moreover, for every such $\eta$,
\begin{equation}\label{main:eq-cost-ratio}
 \frac{w_{B^d,\eta}(z)}{w_{\rm prev}(z)}
 \asymp_{d,\eta}
 h_z^{(d+1)(\eta-\eta_{\rm prev})/2}
 \longrightarrow0,
 \qquad\text{as }h_z\downarrow0.
\end{equation}
Hence the new weight allows arbitrarily larger
coefficients for individual atoms supported on
sufficiently shallow caps. This improvement has a sharp limit. For $\eta_{\rm fast}<\eta\le1/2$,
Theorem~\ref{main:curved} gives
\[
 r_*(B^d,\eta,2)
 =\frac12+\frac{d+1}{d-1}
   \left(\frac12-\eta\right)
 <\alpha.
\]
Thus $\eta_{\rm fast}$ is the exact threshold for
retaining the classical polynomial exponent within
the family $h_z\theta_z^\eta$ on the ball.

The upper bounds in this paper also yield algebraic approximation rates on real interpolation spaces between $L^q(\mu_K)$ and the corresponding weighted variation space. This follows from a standard $K$-functional argument
(see \cite[Section~2.2]{BarronCohenDahmenDeVore2008}).

\paragraph{\normalfont\itshape What do the bounds imply for network compression?}

The approximation error $\mathcal E_n^K$ allows
approximants with arbitrary neuron parameters.
Our upper bounds give a stronger guarantee for
a finite network with a given representation
$F=\sum_{i=1}^N c_i\phi_{z_i}$.
The approximant can be chosen as
$S_n=\sum_{i=1}^N c_i'\phi_{z_i}$ satisfying
\eqref{main:eq-compression}, with the corresponding
error bounds scaled by the original weighted budget
$M=\sum_{i=1}^N|c_i|w_{K,\eta}(z_i)$.
Thus at most $n$ original atoms are retained,
their coefficients may be adjusted, and the weighted
coefficient sum does not increase.
The error guarantees are independent of the original
width $N$ for a fixed budget $M$.
The computational cost of finding such approximants
remains to be analyzed.

A broader open question is to determine the optimal
polynomial approximation exponent for arbitrary
convex bodies.
Theorem~\ref{main:randomized} gives general bounds
in terms of $\overline\kappa(K)$, while
Proposition~\ref{geo:packing-lower} provides lower
bounds on the approximation error using disjoint caps.
It remains unclear whether $\overline\kappa(K)$
alone determines $r_*(K,\eta,q)$ for general
convex bodies, or whether finer geometric
information is needed.
Even for $C^2$ convex bodies in dimensions $d\ge3$,
the classification remains open for finite $q>Q_d$
when the principal curvatures are allowed to vanish.

\section{Proof Overview} \label{sec:proof-overview}

We focus on the proof of the general upper bounds
in Theorem~\ref{main:randomized}, and briefly
outline the remaining upper bounds and the
optimality arguments at the end of this section.
The starting point is Lemma~\ref{geom:local}.
When the activation caps of several atoms lie
in a common convex region whose volume is
comparable to that of each cap, their sum can
be approximated using at most $k$ original atoms
without increasing the weighted coefficient sum.
Writing $V$ for the relative volume of this
region and $M$ for the weighted coefficient sum,
the local $L^q$ error is bounded by
$CMV^{1/q-\eta}(k+1)^{-\alpha}$.
In particular, the exponent $\alpha$ does not
depend on $\eta$.

To approximate the entire network with at most
$n$ neurons, we first group the atoms according
to the volumes and locations of their activation
caps, so that the local approximation result
applies to each group.
Combining these local approximations under the
total width constraint requires controlling both
the number of groups and the overlap of their
enclosing regions.
Both depend on the geometry of $K$.

\paragraph{\normalfont\itshape Grouping atoms by cap volume and location.}

Caps of comparable volume may lie in different parts
of the domain. We therefore group atoms by both cap
volume and location, so that each group admits a local
approximation supported in a common convex region.

Fix $1\le q<\infty$ and $\eta\le1/q$.
By \eqref{eq:domain-error} and Lemma~\ref{lem:closure},
it suffices to consider finite networks
$F=\sum_{i=1}^N c_i\phi_{z_i}$ with
\[
 w_i=w_{K,\eta}(z_i),
 \qquad
 \sum_{i=1}^N|c_i|w_i\le1.
\]
Choose a sufficiently small $0<\tau_0<1$, depending
only on $d$, and divide the atoms into volume layers:
\begin{equation}\label{volume layers}
 \tau_j=2^{-j}\tau_0,
 \qquad
 J_j=\{i:\tau_j/2<\theta_{z_i}\le\tau_j\},
 \qquad j\ge0.
\end{equation}
Lemma~\ref{hil:economic-cover} provides covering caps
$E_{j,1},\ldots,E_{j,N_j}\subset K$ such that every
cap of relative volume at most $\tau_j$ is contained
in one of them, with
\begin{equation}\label{group}
 \mu_K(E_{j,\ell})\asymp_d\tau_j,
 \qquad
 N_j\asymp_d\mathcal C_K(\tau_j).
\end{equation}
Thus the covering caps are comparable in volume to
the activation caps in layer $j$, and the effective
cap count $\mathcal{C}_K$ controls how many covering caps are needed.

Assign each index $i\in J_j$ to one covering cap
$E_{j,\ell}$ containing $C_{z_i}$, and let
$B_{j,\ell}$ be the set of indices assigned to $E_{j,\ell}$.
We call $E_{j,\ell}$ the envelope of this group.
The remaining indices form the additional group
\[
 B_* = \{i:\theta_{z_i}>\tau_0\},
 \qquad E_*=K.
\]
After omitting empty groups, the index sets
$B_{j,\ell}$ and $B_*$ form a finite partition
of $\{1,\ldots,N\}$, with at most $N_j$ groups
in layer $j$.
Although these index sets are disjoint, the
corresponding covering caps $E_{j,\ell}$ may overlap.

For each group $B$ with envelope $E_B$, set
\[
 F_B=\sum_{i\in B}c_i\phi_{z_i},
 \qquad
 M_B=\sum_{i\in B}|c_i|w_i,
 \qquad
 V_B=\mu_K(E_B).
\]
Then $F=\sum_B F_B$ and $\sum_B M_B\le1$.
Moreover,
\[
 C_{z_i}\subset E_B,
 \qquad
 \theta_{z_i}\asymp_d V_B,
 \qquad i\in B,
\]
including for $B_*$, since
$\tau_0<\theta_{z_i}\le1=V_{B_*}$.
These are precisely the geometric conditions used
in Lemma~\ref{geom:local} and
Proposition~\ref{rand:local}.
Using only atoms from $B$ ensures that both the
approximant and its error vanish outside $E_B$.

\paragraph{\normalfont\itshape Local randomized approximation.}

Lemma~\ref{geom:local} provides a local approximation
for each group. To combine the errors from different
groups, we use a randomized version of this
construction.
For each group, we randomly select original atoms
and adjust their coefficients to construct an unbiased
approximation $G_B^{(k)}$ of $F_B$, satisfying
$\mathbb E G_B^{(k)}=F_B$.
We make these constructions independently across
groups, allowing us to estimate the combined error
using moment inequalities rather than simply summing the
individual $L^q$ error bounds.

More precisely, for each group $B$ and integer $k\ge1$,
Proposition~\ref{rand:local} constructs
\[
 G_B^{(k)}=\sum_{i\in B}c_i'\phi_{z_i},
\]
using at most $4k$ original atoms in every realization.
The random approximation satisfies
\begin{equation}\label{eq:overview-random-local}
 \begin{aligned}
  \mathbb E G_B^{(k)}&=F_B,\\
  |F_B(x)-G_B^{(k)}(x)|
  &\le C_{d,\eta}M_BV_B^{-\eta}k^{-\alpha}
       \mathbf1_{E_B}(x),
  \qquad x\in K.
 \end{aligned}
\end{equation}
The pointwise bound holds in every realization.
Moreover,
\begin{equation}\label{eq:overview-local-coefficients}
 \mathbb E c_i'=c_i,\quad\forall i\in B,
 \qquad
 \sum_{i\in B}|c_i'|w_i=M_B.
\end{equation}
Thus the approximation is unbiased, while the
weighted coefficient sum is preserved in every
realization.

The proof of Proposition~\ref{rand:local} first uses
an affine change of variables to apply
Proposition~\ref{rand:uniform} to each group. 
The latter provides unbiased compression with a
uniform error bound, but only preserves the
ordinary sum of absolute coefficients.
Lemma~\ref{rand:cost-mixing} then ensures that the
prescribed weighted coefficient sum is also preserved
in every realization, without losing unbiasedness
or the pointwise error bound.

To prepare for combining the local errors, set
\[
 \varepsilon=1/q-\eta,\qquad
 \lambda_B=V_B^\varepsilon,\qquad
 \psi_B=V_B^{-1/q}\mathbf1_{E_B}.
\]
Since $\|\psi_B\|_q=1$ and
$\lambda_B\psi_B=V_B^{-\eta}\mathbf1_{E_B}$,
\eqref{eq:overview-random-local} becomes
\begin{equation}\label{eq:overview-random-local2}
 |F_B-G_B^{(k)}|
 \le C_{d,\eta}\lambda_BM_Bk^{-\alpha}\psi_B.
\end{equation}
Here $\psi_B$ is the normalized indicator of $E_B$,
while $\lambda_B$ gives the volume factor in the
$L^q$ error bound
\[
 \|F_B-G_B^{(k)}\|_q
 \le C_{d,\eta}\lambda_BM_Bk^{-\alpha}.
\]
For a group in layer $j$,
$\lambda_B\asymp_{d,q,\eta}\tau_j^\varepsilon$,
whereas $\lambda_{B_*}=1$.
Thus, for fixed $M_B$ and $k$, the $L^q$ error
bound decreases as $V_B$ decreases when
$\eta<1/q$. At $\eta=1/q$, this volume
decay disappears.

\paragraph{\normalfont\itshape Combining local approximation with sampling.}

Approximating every group separately may require
more than $n$ neurons, since each group needs its
own width allocation.
We therefore combine local approximation with
sampling.
We approximate some groups individually and treat
the remaining groups together by randomly selecting
a fixed number of their original atoms, with
coefficients adjusted to obtain an unbiased approximation.
Local approximation exploits the rate $k_B^{-\alpha}$
but requires allocating width to each group.
Sampling avoids a separate width allocation for
each remaining group, but its error must also be
controlled.
We balance these two contributions under the
total width constraint.

Partition the groups into two collections
$\mathcal H$ and $\mathcal R$.
For each $B\in\mathcal H$, choose an integer
$k_B\ge1$ and use the local randomized
approximant $G_B^{(k_B)}$ from
Proposition~\ref{rand:local}. 
For the remaining groups, write
\[
 F_{\mathcal R}=\sum_{B\in\mathcal R}F_B,
 \qquad
 A=\sum_{B\in\mathcal R}M_B.
\]
If $A>0$, draw $L\ge1$ indices
$I_1,\ldots,I_L$ independently, with replacement,
according to
\begin{equation}\label{eq:overview-sampling-probabilities}
 \mathbb P(I_r=i)=\frac{|c_i|w_i}{A},
 \qquad
 i\in\bigcup_{B\in\mathcal R}B,
 \quad r=1,\ldots,L,
\end{equation}
and define
\[
 X_r=\frac{A}{L}
      \frac{\operatorname{sgn}(c_{I_r})}{w_{I_r}}
      \phi_{z_{I_r}}.
\]
Each sampled term has weighted coefficient sum
$A/L$, while the choice of probabilities ensures
$\mathbb E X_r=F_{\mathcal R}/L$.
Thus $\sum_{r=1}^L X_r$ is an unbiased approximation
of $F_{\mathcal R}$ using at most $L$ original atoms.

Combining repeated selections of the same index, write
\[
 \sum_{r=1}^L X_r
 =\sum_{B\in\mathcal R}\sum_{i\in B}
   \widehat c_i\phi_{z_i}.
\]
Since all contributions to $\widehat c_i$ have
the same sign as $c_i$, their absolute values add.
Consequently,
\begin{equation}\label{eq:overview-sampling-budget}
 \sum_{B\in\mathcal R}\sum_{i\in B}
      |\widehat c_i|w_i=A
\end{equation}
in every realization. Moreover,
\begin{equation}\label{eq:overview-sampling-unbiased}
 \mathbb E\widehat c_i=c_i,
 \qquad i\in\bigcup_{B\in\mathcal R}B.
\end{equation}
If $A=0$, then $F_{\mathcal R}=0$, and we set
$L=0$ and omit the sampled sum.

Choose the local approximants $G_B^{(k_B)}$
independently across $B\in\mathcal H$ and
independently of the sampled indices, and set
\[
 G=\sum_{B\in\mathcal H}G_B^{(k_B)}
      +\sum_{r=1}^L X_r.
\]
The resulting network has width at most $n$ provided
\[
 4\sum_{B\in\mathcal H}k_B+L\le n.
\]
Writing $G=\sum_{i=1}^N\widetilde c_i\phi_{z_i}$,
the properties of the local approximations and
the sampling construction give
\[
 \begin{gathered}
  \mathbb E\widetilde c_i=c_i
  \quad(1\le i\le N),
  \qquad \mathbb EG=F,\\
  \sum_{i=1}^N|\widetilde c_i|w_i
  =\sum_{B\in\mathcal H}M_B+A
  =\sum_{i=1}^N|c_i|w_i\le1.
 \end{gathered}
\]
The weighted coefficient identity holds in every
realization. This is the construction in
Lemma~\ref{rand:hybrid}.
We next estimate its approximation error, which
will determine how to choose $\mathcal H$, the
local widths $k_B$, and the number of samples $L$
under the total width constraint.

\paragraph{\normalfont\itshape Estimating the total error.}

For $B\in\mathcal H$, set
$e_B=F_B-G_B^{(k_B)}$.
Since $\sum_{r=1}^L\mathbb E X_r=F_{\mathcal R}$,
the total error can be written as
\begin{equation}\label{eq:overview-error-decomposition}
 F-G=\sum_{B\in\mathcal H}e_B
     +\sum_{r=1}^L(\mathbb EX_r-X_r).
\end{equation}
When $I_r\in B$ for $B\in\mathcal R$, the sampled
term $X_r$ is supported in $E_B$.
However, its expectation
$\mathbb EX_r=F_{\mathcal R}/L$ includes
contributions from all remaining groups, so
$\mathbb EX_r-X_r$ need not be supported in $E_B$.
Using independence and unbiasedness, we apply
symmetrization to bound the total error in terms
of the local errors $e_B$ and the original samples
$X_r$, whose supports can be controlled by their
respective envelopes.

By symmetrization and Khintchine's inequality,
as in the proof of Lemma~\ref{rand:hybrid},
we obtain
\begin{equation}\label{eq:overview-symmetrized-error}
 \left(\mathbb E\|F-G\|_q^q\right)^{1/q}
 \le C_q
 \left(
  \mathbb E
  \left\|
   \left(
    \sum_{B\in\mathcal H}|e_B|^2
    +\sum_{r=1}^L|X_r|^2
   \right)^{1/2}
  \right\|_q^q
 \right)^{1/q}.
\end{equation}
We can now use the pointwise bounds on
$e_B$ and $X_r$ within their respective envelopes
to estimate the right-hand side.

The local errors $e_B$, $B\in\mathcal H$, are
controlled by \eqref{eq:overview-random-local2}.
For the sampled terms, let $B\in\mathcal R$
and $i\in B$.
Since $C_{z_i}\subset E_B$ and
$\theta_{z_i}\asymp_d V_B$, we have
\[
 \left|\frac{\phi_{z_i}}{w_i}\right|
 \le C_{d,\eta}V_B^{-\eta}\mathbf1_{E_B}
 =C_{d,\eta}\lambda_B\psi_B.
\]
Consequently, when $A>0$,
\begin{equation}\label{eq:overview-sample-bound}
 |X_r|
 \le C_{d,\eta}\frac{A}{L}\lambda_B\psi_B
 \qquad\text{if }I_r\in B.
\end{equation}

For $B\in\mathcal R$, let
$N_B=\#\{r\in\{1,\ldots,L\}:I_r\in B\}$,
counting repeated selections.
When $A>0$, applying
\eqref{eq:overview-sample-bound} to these
$N_B$ terms gives
\[
 \sum_{r:I_r\in B}|X_r|^2
 \le C_{d,\eta}^2
      \left(\frac{A}{L}\right)^2
      N_B\lambda_B^2\psi_B^2.
\]
This explains the factor $\sqrt{N_B}$ in the
following definition:
\begin{equation} \label{eq:overview-U}
    U_B=
 \begin{cases}
  \lambda_BM_Bk_B^{-\alpha},
      &B\in\mathcal H,\\[2pt]
  (A/L)\lambda_B\sqrt{N_B},
      &B\in\mathcal R,\ A>0,\\[2pt]
  0,  &B\in\mathcal R,\ A=0.
 \end{cases}
\end{equation}
Combining the pointwise bounds for the local
errors and sampled terms gives, in every realization,
\begin{equation}\label{eq:overview-combined-squares}
 \sum_{B\in\mathcal H}|e_B|^2
 +\sum_{r=1}^L|X_r|^2
 \le C_{d,\eta}^2\sum_B U_B^2\psi_B^2.
\end{equation}
Set
\begin{equation}\label{eq:overview-square-function}
 \begin{aligned}
 \mathfrak S_q(U)
 &=\left\|\left(
   \sum_B U_B^2\psi_B^2
   \right)^{1/2}\right\|_q\\
 &=\left(
   \int_K
   \left(
     \sum_{B:\,x\in E_B}
     U_B^2V_B^{-2/q}
   \right)^{q/2}
   d\mu_K(x)
   \right)^{1/q}.
 \end{aligned}
\end{equation}
Together with \eqref{eq:overview-symmetrized-error},
this yields
\begin{equation}\label{eq:overview-square}
 \left(\mathbb E\|F-G\|_q^q\right)^{1/q}
 \le C_{d,q,\eta}
 \left(\mathbb E\mathfrak S_q(U)^q\right)^{1/q}.
\end{equation}
This is the error estimate \eqref{rand:eq-hybrid}
in Lemma~\ref{rand:hybrid}.

Once the groups, the integers $k_B$, and the sample
count $L$ are fixed, $U_B$ is deterministic for
$B\in\mathcal H$, whereas for $B\in\mathcal R$
it depends on the random sample count $N_B$.
The functions $\psi_B=V_B^{-1/q}\mathbf1_{E_B}$
encode the volumes and locations of the envelopes.

In the integral representation of
\eqref{eq:overview-square-function}, the sum at
each point $x$ runs over all groups with $x\in E_B$.
For $q=2$, we have
$\mathfrak S_2(U)^2=\sum_B U_B^2$, regardless of
overlap.
For $q>2$, however, contributions from overlapping
envelopes are combined before taking the power
$q/2>1$.

\paragraph{\normalfont\itshape Controlling overlap by floating depth.}

We next bound $\mathfrak S_q(U)$
in \eqref{eq:overview-square-function}.
For $q>2$, its definition gives
\[
 \mathfrak S_q(U)^2
 =\left\|
   \sum_B U_B^2V_B^{-2/q}\mathbf1_{E_B}
  \right\|_{q/2}.
\]
We seek a bound in terms of the coefficients $U_B$,
with a constant independent of the number of groups.
To achieve this, we use floating depth to control
the effect of overlapping envelopes.

Recall that layer $j$ consists of atoms with
$\tau_j/2<\theta_{z_i}\le\tau_j$, where
$\tau_j=2^{-j}\tau_0$.
The envelopes $E_{j,\ell}$ of the corresponding
groups $B_{j,\ell}$ satisfy
$\mu_K(E_{j,\ell})\asymp_d\tau_j$.
The number of these envelopes containing $x$ is
\[
 m_j(x)=\sum_{\ell:\,B_{j,\ell}\ne\varnothing}
             \mathbf1_{E_{j,\ell}}(x).
\]
Proposition~\ref{rand:multiplicity} gives, for
almost every $x\in K$,
\begin{equation}\label{eq:overview-multiplicity}
 m_j(x)\le C_d
 \left(\frac{\tau_j}{v_K(x)}\right)^{(d-2)/d}
 \log\left(1+\frac{\tau_j}{v_K(x)}\right).
\end{equation}
The proof uses the pairwise disjoint inner sets
from Lemma~\ref{hil:economic-cover}, each of
relative volume at least $c_d\tau_j$.
Their inclusion in the envelopes gives a lower
bound $c_dm_j(x)\tau_j$ for the relative volume
of the union of envelopes containing $x$.
Combining this lower bound with an upper bound
obtained using Proposition~\ref{rand:visibility}
yields \eqref{eq:overview-multiplicity}.

Although \eqref{eq:overview-multiplicity} allows
large overlap, it leads to a uniform bound for
the sum of indicators weighted by
$(v_K(x)/V_B)^\rho$. Indeed, for $\rho>(d-2)/d$,
\begin{equation}\label{eq:overview-weighted-overlap}
 \begin{aligned}
 &\sum_B
 \left(\frac{v_K(x)}{V_B}\right)^\rho
 \mathbf1_{E_B}(x)\le
 1+C_{d,\rho}\sum_{j\ge0}
 \left(\frac{v_K(x)}{\tau_j}\right)^\rho m_j(x)
 \le C_{d,\rho}
 \end{aligned}
\end{equation}
for almost every $x\in K$.
The additional $1$ accounts for the possible
envelope $E_B=K$.
The last bound follows by summing over the
geometrically decreasing scales $\tau_j$, since
$m_j(x)>0$ implies $v_K(x)\le C_d\tau_j$;
see the proof of Theorem~\ref{rand:positive-synthesis}.

To use this bound with H\"older's inequality,
we also need to control integrals of powers
of $V_B/v_K(x)$ over each envelope.
Lemma~\ref{rand:depth-moments} gives, for every
$E_B$, including $E_B=K$,
\begin{equation}\label{eq:overview-depth-tail}
 \mu_K\{x\in E_B:v_K(x)<tV_B\}
 \le C_dV_Bt^{2/(d+1)},
 \qquad 0<t<1.
\end{equation}
Thus points with $v_K(x)<tV_B$ occupy at most
a fraction $C_dt^{2/(d+1)}$ of $E_B$. Since $0<v_K(x)\le V_B$ for almost every
$x\in E_B$, we can use the identity
\[
 r^{-\delta}
 =1+\delta\int_r^1 t^{-\delta-1}\,dt,
 \qquad 0<r\le1.
\]
Applying it with $r=v_K(x)/V_B$ and using
\eqref{eq:overview-depth-tail}, we obtain,
for $0<\delta<2/(d+1)$,
\begin{equation}\label{eq:overview-negative-depth}
 \begin{aligned}
 \int_{E_B}
  \left(\frac{V_B}{v_K(x)}\right)^\delta
  \,d\mu_K(x)
 &=V_B+\delta\int_0^1 t^{-\delta-1}
  \mu_K\{x\in E_B:v_K(x)<tV_B\}\,dt\\
 &\le V_B+C_d\delta V_B
  \int_0^1 t^{2/(d+1)-\delta-1}\,dt\le C_{d,\delta}V_B.
 \end{aligned}
\end{equation}

For $1<s<Q_d/2$, we can choose
$\rho>(d-2)/d$ such that
$\delta=\rho(s-1)<2/(d+1)$,
where $Q_d$ is defined in
\eqref{main:eq-universal-parameters}.
For any real coefficients $(a_B)$,
H\"older's inequality and
\eqref{eq:overview-weighted-overlap} give
\[
 \begin{aligned}
 \left|\sum_B a_B\mathbf1_{E_B}(x)\right|
 &\le
 \left(
 \sum_B |a_B|^s
 \left(\frac{V_B}{v_K(x)}\right)^\delta
 \mathbf1_{E_B}(x)
 \right)^{1/s}\left(
 \sum_B
 \left(\frac{v_K(x)}{V_B}\right)^\rho
 \mathbf1_{E_B}(x)
 \right)^{(s-1)/s}\\
 &\le C_{d,\rho}^{1-1/s}
 \left(
 \sum_B |a_B|^s
 \left(\frac{V_B}{v_K(x)}\right)^\delta
 \mathbf1_{E_B}(x)
 \right)^{1/s}.
 \end{aligned}
\]
This is the step used in
Lemma~\ref{geom:depth-synthesis}.
Taking $L^s(\mu_K)$ norms and applying
\eqref{eq:overview-negative-depth} gives
\begin{equation}\label{eq:overview-positive}
 \left\|\sum_B a_B\mathbf1_{E_B}\right\|_s
 \le C_{d,s}
      \left(\sum_B|a_B|^sV_B\right)^{1/s},
 \qquad 1\le s<Q_d/2.
\end{equation}
The case $s=1$ follows from the triangle
inequality.
This is \eqref{rand:eq-operator-bound}, obtained
from Theorem~\ref{rand:positive-synthesis}.

Finally, taking $s=q/2$ and
$a_B=U_B^2V_B^{-2/q}$ yields
\[
 \mathfrak S_q(U)
 \le C_{d,q}
 \left(\sum_B U_B^q\right)^{1/q},
 \qquad 2<q<Q_d,
\]
in every realization. For $1\le q\le2$, using
$(\sum_B b_B)^{q/2}\le\sum_B b_B^{q/2}$
for $b_B\ge0$, we obtain $\mathfrak S_q(U)^q
 \le\sum_B U_B^q\|\psi_B\|_q^q
 =\sum_B U_B^q$. Equality holds when $q=2$, so no geometric
overlap estimate is needed in this range.

For $q\ge Q_d$, the choice $s=q/2$ lies outside
the range of \eqref{eq:overview-positive}.
We instead bound $\mathfrak S_q(U)$ by the
$\ell^t$ norm of $(U_B)$ for suitable $t<q$,
with a constant independent of the number of
groups.
We omit the details here and refer to the proof
of Corollary~\ref{rand:strong-square}.

To state both ranges together, write
$\gamma=1-1/t$.
Corollary~\ref{rand:strong-square} gives, in every
realization,
\begin{equation}\label{eq:overview-sequence}
 \mathfrak S_q(U)
 \le C_{d,q,\gamma}
      \left(\sum_B U_B^t\right)^{1/t},
 \qquad t=\frac1{1-\gamma},
\end{equation}
where
\[
 \begin{cases}
  \gamma=1-1/q,
     &1\le q<Q_d,\\[2pt]
  \frac12\le\gamma<\Gamma_d(q),
     &Q_d\le q<\infty.
 \end{cases}
\]
Here $\Gamma_d(q)$ is as in
Theorem~\ref{main:randomized}.
Thus $t=q$ in the first range and $t<q$ in
the second.

Substituting \eqref{eq:overview-sequence} into
\eqref{eq:overview-square} yields
\begin{equation}\label{eq:overview-sequence-error}
 \left(\mathbb E\|F-G\|_q^q\right)^{1/q}
 \le C_{d,q,\eta,\gamma}
 \left[
  \mathbb E\left(\sum_B U_B^t\right)^{q/t}
 \right]^{1/q}.
\end{equation}
We next choose the groups to approximate separately,
their local widths $k_B$, and the number of samples
$L$ to control this bound under the total width
constraint.

\paragraph{\normalfont\itshape Allocating the approximation width.}

Recall that $\varepsilon=1/q-\eta\ge0$ and set
$a=\alpha-\gamma>0$.
We first consider $\varepsilon>0$. Since $\lambda_B\asymp\tau_j^\varepsilon$
for groups in layer $j$, we divide the
volume layers at an integer $J\ge0$. The groups in layers $0,\ldots,J$, together
with the possible group having envelope $K$,
are approximated separately.
The finer layers are treated using
Lemma~\ref{rand:strong-minimum}: when there are
too many groups to approximate separately,
those with larger weighted coefficient sums
$M_B$ are approximated locally, while the
remaining groups are sampled together.

Define
\[
 D_J=1+\sum_{j=0}^J\mathcal C_K(\tau_j),
 \qquad
 Q_J=1+\sum_{j=0}^J
       \mathcal C_K(\tau_j)\tau_j^{\varepsilon/a}.
\]
There are at most $C_dD_J$ groups to approximate
separately.
If $D_J\le cn$ for a sufficiently small constant
$c>0$, the two parts can be approximated with
total width at most $n$.
Proposition~\ref{rand:profile} gives, for $n\ge1$,
\begin{equation}\label{eq:overview-profile}
 \mathcal E_n^K(w_{K,\eta};q)
 \le C\left(
  n^{-\alpha}Q_J^a
  +n^{-\gamma}\tau_{J+1}^{\varepsilon}
 \right).
\end{equation}
For the groups in layers $0,\ldots,J$ and the
possible group $B_*$, we use
$U_B=\lambda_BM_Bk_B^{-\alpha}$ in
\eqref{eq:overview-sequence-error}.
Set $\rho=(\alpha+1/t)^{-1}=1/(1+a)$ and
$Z=\sum_B(\lambda_BM_B)^\rho$.
Allocating widths according to
$k_B\asymp n(\lambda_BM_B)^\rho/Z$
before rounding gives
\[
\begin{aligned}
\left(\sum_B U_B^t\right)^{1/t}
&=\left(\sum_B(\lambda_BM_B)^t
k_B^{-\alpha t}\right)^{1/t}\\
&\le C\left(\frac Zn\right)^\alpha
\left(\sum_B(\lambda_BM_B)^{t(1-\alpha\rho)}
\right)^{1/t}\\
&=Cn^{-\alpha}Z^{\alpha+1/t}
= Cn^{-\alpha}Z^{1/\rho},
\end{aligned}
\]
where the sums are over these groups. 
By H\"older's inequality, $\sum_BM_B\le1$,
and the group count in \eqref{group},
\[
\begin{aligned}
Z^{1/\rho}
&\le \left(\sum_BM_B\right)
     \left(\sum_B\lambda_B^{1/a}\right)^a\\
&\le C\left(1+\sum_{j=0}^J
\mathcal C_K(\tau_j)\tau_j^{\varepsilon/a}
\right)^a
=CQ_J^a.
\end{aligned}
\]
For the finer layers, the bound
$\lambda_B\le C\tau_{J+1}^{\varepsilon}$
yields the remaining term in
\eqref{eq:overview-profile}. Choosing a realization whose $L^q$ error is bounded by the right-hand side of
\eqref{eq:overview-profile} gives an approximant
using at most $n$ original atoms, without
increasing the weighted coefficient sum. 
To choose $J$ for a general convex body $K$,
fix $\kappa>\overline\kappa(K)$.
By \eqref{intr:eq-kappa},
$\mathcal C_K(\tau)\le C_{K,\kappa}\tau^{-\kappa}$
for all sufficiently small $\tau>0$.
Absorbing the finitely many larger scales into
the constants gives
\[
 D_J\le C_D2^{\kappa J},
 \qquad
 Q_J\le C\left(
  1+\sum_{j=0}^J2^{j(\kappa-\varepsilon/a)}
 \right).
\]
For sufficiently large $n$, choose $J$ as the
largest nonnegative integer satisfying
$C_D2^{\kappa J}\le cn$.
Then $D_J\le cn$ and
$\tau_{J+1}\asymp n^{-1/\kappa}$.
Estimating the geometric sum for $Q_J$ and
substituting into \eqref{eq:overview-profile}
gives
\[
 r_*(K,\eta,q)\ge
 \min\left\{\alpha,\,
  \gamma+\frac{\varepsilon}{\kappa}\right\},
\]
as established in Corollary~\ref{rand:power}.

If $\overline\kappa(K)>0$, letting
$\kappa\downarrow\overline\kappa(K)$ and, when
$q\ge Q_d$, $\gamma\uparrow\Gamma_d(q)$ yields
the general exponent bound in
Theorem~\ref{main:randomized}.
If $\overline\kappa(K)=0$, choosing $\kappa>0$
sufficiently small gives
$r_*(K,\eta,q)\ge\alpha$. 
At $\varepsilon=0$, we have $\lambda_B=1$
for every group.
In this case, Lemma~\ref{rand:strong-minimum}
can be applied directly to all groups, giving
$\mathcal E_n^K(w_{K,1/q};q)\le Cn^{-\gamma}$ for any $n\ge1$. 
No cutoff $J$ is needed for this estimate.

\paragraph{\normalfont\itshape The remaining upper and lower bounds.} 
The remaining upper bounds use the same local
approximation result, Lemma~\ref{geom:local},
but combine the local errors deterministically,
without sampling, using coverings adapted to
the geometry of the domain.
For planar convex bodies and their products
with polytopes, Propositions~\ref{intr:planar-cover}
and~\ref{intr:product-cover} provide the covering
estimates used in Proposition~\ref{geo:layer-upper}.
For convex bodies with $C^2$ boundary and everywhere
positive principal curvatures,
Proposition~\ref{curvednew:atlas} provides the
geometric estimates used in
Proposition~\ref{geo:hardy-upper}.
These yield the upper bounds in
Theorems~\ref{main:intrinsic} and
\ref{main:curved}, respectively.

The lower bounds use two complementary arguments.
Proposition~\ref{geo:packing-lower} uses caps
with pairwise disjoint interiors to obtain a
lower bound depending on their number and volume. 
Theorem~\ref{geo:interior-lower} gives a separate
lower bound showing that the optimal polynomial
exponent cannot exceed $\alpha$ for $q\ge2$.
Combining these lower bounds with the general
and domain-specific upper bounds, together with
the relevant geometric estimates, establishes
the optimal exponents in
Theorems~\ref{main:c2}, \ref{main:curved},
and~\ref{main:intrinsic}.

\section{Proofs of the Upper Bounds for General Convex Bodies}
\label{sec:randomized-proofs}\label{hil:section}
We prove Theorem~\ref{main:randomized} by retaining at most $n$ terms of a finite ReLU sum and adjusting their coefficients. We group terms by the volumes and locations of their caps. Proposition~\ref{rand:local} provides a random approximation to the sum within each group. When the available width $n$ is insufficient to approximate every group separately, we use these approximations for groups with larger weighted sums of absolute coefficients. As in Lemmas~\ref{rand:hybrid} and~\ref{rand:strong-minimum}, we approximate the sum over all remaining groups using a fixed total number of randomly selected terms, with coefficients chosen so that the expected approximation equals the original sum over all remaining groups. We combine the volume estimates in Lemmas~\ref{hil:economic-cover} and~\ref{rand:depth-moments} with the overlap bound in Proposition~\ref{rand:multiplicity} to derive Corollary~\ref{rand:strong-square}, which we use to control the total error.

\subsection{Network Compression in the Uniform Norm}
\label{sec:common}

In this subsection, we construct random compressions of finite
ReLU sums that preserve each coefficient in expectation and satisfy
error bounds and coefficient constraints in every realization.
These properties will be used later to construct independent local
approximations and estimate their combined error.
We also obtain a deterministic approximation using at most
the prescribed number of original neurons.

The following lemma represents any probability vector $p$ as the expectation of random weight vectors
with at most $\operatorname{rank}M$ nonzero entries in each realization.
Without this support bound, the deterministic choice $X=p$ would
already satisfy all the conclusions.
\begin{lemma}
\label{rand:positive-moments}
Let $p=(p_1,\ldots,p_N)\in\mathbb R^N$ be a fixed vector satisfying
\[
 p_i\ge0 \quad (1\le i\le N),
 \qquad \sum_{i=1}^N p_i=1,
\]
and let $M\in\mathbb R^{m\times N}$ have first row $(1,\ldots,1)$.
For $y\in\mathbb R^N$, write
$\operatorname{supp}y=\{i:y_i\ne0\}$.

There exists a random vector $X\in\mathbb R^N$, taking only finitely
many values, such that $\mathbb EX=p$ and every possible value $x$
of $X$ satisfies
\[
 x_i\ge0 \quad (1\le i\le N),
 \qquad \sum_{i=1}^N x_i=1,
 \qquad Mx=Mp,
\]
and
\[
 \operatorname{supp}x\subseteq\operatorname{supp}p,
 \qquad
 |\operatorname{supp}x|\le\operatorname{rank}M.
\]
\end{lemma}

\begin{proof}
Consider
\[
 P=\{x\in\mathbb R^N:
       x\ge0,\ Mx=Mp,\ x_i=0
       \text{ for }i\notin\operatorname{supp}p\}.
\]
This set contains $p$. Since the first row of $M$ consists of ones,
every $x\in P$ satisfies $\sum_{i=1}^N x_i=1$.
Together with $x\ge0$, this shows that $P$ is bounded.
Since $P$ contains $p$ and is defined by finitely many linear
equalities and inequalities, it is a nonempty compact polytope.

We first show that every vertex of $P$ has at most
$\operatorname{rank}M$ nonzero coordinates.
Let $x$ be a vertex and put $S=\operatorname{supp}x$.
If the columns of $M$ indexed by $S$ were linearly dependent,
there would be a nonzero vector $v$ supported on $S$ with $Mv=0$.
Since $x_i>0$ for every $i\in S$, both $x+\epsilon v$ and
$x-\epsilon v$ would belong to $P$ for sufficiently small
$\epsilon>0$.
This would express $x$ as the midpoint of two distinct points of
$P$, contradicting that $x$ is a vertex.
Hence these columns are linearly independent, and
$|S|\le\operatorname{rank}M$.

A compact polytope is the convex hull of its vertices.
We may therefore write
\[
 p=\sum_{j=1}^L\lambda_j x^{(j)},
 \qquad
 \lambda_j>0,
 \qquad
 \sum_{j=1}^L\lambda_j=1,
\]
where $x^{(1)},\ldots,x^{(L)}$ are vertices of $P$.
Define $X$ by choosing $x^{(j)}$ with probability $\lambda_j$.
Then
\[
 \mathbb EX=\sum_{j=1}^L\lambda_jx^{(j)}=p.
\]
Every possible value $x$ of $X$ belongs to $P$, so it is nonnegative,
has coordinate sum one, satisfies $Mx=Mp$, and is supported on $\operatorname{supp}p$.
Moreover, $x$ is a vertex of $P$, so the argument above gives
$|\operatorname{supp}x|\le\operatorname{rank}M$.
\end{proof}

The following consequence of Lemma~\ref{rand:positive-moments}
ensures exact preservation of a weighted coefficient sum in every
realization, at the cost of at most doubling the support size.
\begin{lemma}
\label{rand:cost-mixing}
Let $a\in\mathbb R^N$ be a random vector taking finitely many
values, with $\mathbb Ea=c$.
Suppose every realization has at most $W$ nonzero coordinates
and satisfies $a_i c_i\ge0$ for all $i$, with $a_i=0$ whenever
$c_i=0$.

For fixed weights $\rho_i\ge0$, there exists a random vector
$\widehat a$, taking finitely many values and satisfying, in every realization,
$\widehat a_i c_i\ge0$ for all $i$ and
$\widehat a_i=0$ whenever $c_i=0$, such that $\mathbb E\widehat a=c$
and, in every realization,
\begin{align}
 |\operatorname{supp}\widehat a|
 &\le2W,
 \label{rand:eq-mixed-width}\\
 \sum_i\rho_i|\widehat a_i|
 &=\sum_i\rho_i|c_i|.
 \label{rand:eq-mixed-cost}
\end{align}
Each possible value of $\widehat a$ is a convex combination
of at most two possible values of $a$.
If $\sum_i|a_i|=\sum_i|c_i|$ in every realization, then
$\sum_i|\widehat a_i|=\sum_i|c_i|$ in every realization as well.

For fixed functions $\psi_i$, set $F=\sum_i c_i\psi_i$.
If
\[
 \left|F-\sum_i a_i\psi_i\right|\le B
\]
holds pointwise in every realization for a fixed function
$B\ge0$, then the same bound holds with $a$ replaced by
$\widehat a$.
\end{lemma}

\begin{proof}
Write $\mathbb P(a=a^{(j)})=p_j$, for any $1\le j\le J$, and set
\[
 L_j=\sum_i\rho_i|a_i^{(j)}|,
 \qquad
 L_0=\sum_i\rho_i|c_i|.
\]
Since $a_i^{(j)}c_i\ge0$, with $a_i^{(j)}=0$ if $c_i=0$,
\[
 \sum_jp_j|a_i^{(j)}|
 =\left|\sum_jp_j a_i^{(j)}\right|
 =|c_i|
 \qquad \forall 1\le i\le N.
\]
Hence $\sum_jp_jL_j=L_0$.

Apply Lemma~\ref{rand:positive-moments} to
$p=(p_1,\ldots,p_J)$ and
\[
 M=
 \begin{pmatrix}
  1   & \cdots & 1\\
  L_1 & \cdots & L_J
 \end{pmatrix}.
\]
It gives a random vector $X$ taking finitely many values,
with $\mathbb EX=p$, such that every realization satisfies
\[
 X_j\ge0,\qquad
 \sum_jX_j=1,\qquad
 \sum_jX_jL_j=L_0,\qquad
 |\operatorname{supp}X|\le2.
\]
Define $\widehat a=\sum_jX_j a^{(j)}$. Then we have
$\mathbb E\widehat a
 =\sum_jp_j a^{(j)}=c.$
 
Since $|\operatorname{supp}X|\le2$, each realization of $\widehat a$
is a convex combination of at most two vectors $a^{(j)}$.
It therefore has at most $2W$ nonzero coordinates, each with
the same sign as the corresponding coordinate of $c$.

Since corresponding coefficients have the same sign,
\[
 |\widehat a_i|=\sum_jX_j|a_i^{(j)}|,
 \qquad
 \sum_i\rho_i|\widehat a_i|
 =\sum_jX_jL_j=L_0.
\]
The first identity also implies that if $\sum_i|a_i^{(j)}|=A$
for every $j$, then $\sum_i|\widehat a_i|=A$ in every realization.

Finally, using $\sum_jX_j=1$ and the triangle inequality,
\[
 \left|F-\sum_i\widehat a_i\psi_i\right|
 \le
 \sum_jX_j\left|F-\sum_i a_i^{(j)}\psi_i\right|
 \le B.
\]
This proves the pointwise error assertion.
\end{proof}

We next establish a randomized compression result for finite
ReLU sums. The construction preserves each coefficient in expectation
and achieves the uniform approximation rate $m^{-(d+3)/(2d)}$
in every realization.
The proof adapts the coefficient update used to prove
Proposition~1 in \cite{Siegel2025}, choosing between two symmetric
updates with equal probabilities.
\begin{proposition}
\label{rand:uniform}
Let
\[
 F(z)=\sum_{i=1}^N c_i(\omega_i\cdot z+b_i)_+,
 \qquad |\omega_i|=1,\quad |b_i|\le1,\quad
 A=\sum_i|c_i|.
\]
For every integer $m\ge1$, there exists a random coefficient
vector $c'=(c_1',\ldots,c_N')$, taking finitely many values,
with $\mathbb Ec_i'=c_i$ for every $i$, such that
\[
 G(z)=\sum_{i=1}^N c_i'(\omega_i\cdot z+b_i)_+
\]
satisfies, in every realization,
\begin{align}
 \|F-G\|_{C(B^d)}
 &\le C_d A m^{-\alpha},
 \label{rand:eq-uniform}\\
 |\operatorname{supp}c'|
 &\le2m,
 \qquad
 \sum_i|c_i'|=A,
 \label{rand:eq-uniform-cost}
\end{align}
where $\alpha=(d+3)/(2d)$ and $C_d$ depends only on $d$.
In every realization, $c_i'c_i\ge0$ for all $i$, and
$c_i'=0$ whenever $c_i=0$.
\end{proposition}

\begin{proof}
Write $\varphi_i(z)=(\omega_i\cdot z+b_i)_+$.
If $A=0$, take $c'=0$.
We first assume that $c_i>0$ for all $i$ and that the parameter
pairs $(\omega_i,b_i)$ are distinct.

To describe one compression step, let $a\in[0,\infty)^N$ satisfy
$\sum_i a_i=A$ and $n=|\operatorname{supp}a|\ge6$.
The construction in the proof of Proposition~1 and equations
(3.37)--(3.45) of \cite{Siegel2025}, applied to $a/A$ and
then rescaled by $A$, gives two updates $a^+$ and $a^-$
supported on $\operatorname{supp}a$, with
\[
 a^\pm\ge0,\qquad
 \sum_i a_i^\pm=A,\qquad
 \frac{a^++a^-}{2}=a,\qquad
 |\operatorname{supp}a^\pm|\le\vartheta n,
 \quad \vartheta=\frac{47}{48},
\]
and
\[
 \left\|\sum_i(a_i^\pm-a_i)\varphi_i\right\|_{C(B^d)}
 \le C_d A n^{-\alpha}.
\]

To justify the symmetric form above, we examine the update used
in the proof of Proposition~1 in \cite{Siegel2025}.
Let $t$ be the number of triples selected in that construction.
They are pairwise disjoint subsets
$\{u_j,v_j,w_j\}\subset\operatorname{supp}a$, $1\le j\le t$,
ordered so that $a_{u_j}\le a_{v_j}\le a_{w_j}$.
They cover at least half of the indices whose coefficients are
at or below the median of the $n$ positive coefficients.
There are at least $n/2$ such indices, so $3t\ge n/4$.
The argument in the cited proof then chooses
$\chi_j\in\{-1,0,1\}$ such that
\[
 \#\{j:\chi_j\ne0\}\ge t/4\ge n/48
\]
and the update with $\varepsilon=1$ below satisfies the stated
uniform error bound.

For $\varepsilon\in\{-1,1\}$, define
\[
 (a_{u_j}^{\varepsilon},
  a_{v_j}^{\varepsilon},
  a_{w_j}^{\varepsilon})
 =
 \bigl((1-\varepsilon\chi_j)a_{u_j},\,
       (1+\varepsilon\chi_j)a_{v_j},\,
       a_{w_j}+\varepsilon\chi_j(a_{u_j}-a_{v_j})\bigr),
\]
using the same $\varepsilon$ for every triple and leaving all
other coefficients unchanged.
The choices $\varepsilon=1$ and $\varepsilon=-1$ give
$a^+$ and $a^-$, respectively. The ordering within each triple ensures $a^\pm\ge0$.
Summing the update formula gives $\sum_i a_i^\pm=A$,
and averaging over the two signs gives $(a^++a^-)/2=a$.
When $\chi_j=0$, the triple is unchanged; when $\chi_j\ne0$,
one of the coefficients at $u_j$ and $v_j$ becomes zero
for either choice of $\varepsilon$.
Thus both updates have at most $47n/48$ nonzero coordinates.
Since $a^--a=-(a^+-a)$, their function errors have the same
uniform norm. The cited error bound for $a^+$ therefore
also holds for $a^-$.

We now return to the proof of the proposition.
For $m\ge6$, start from $a^{(0)}=c$.
If $|\operatorname{supp}a^{(r)}|>m$, fix a pair of updates
for $a^{(r)}$ as above, and choose either one with probability
$1/2$ to obtain $a^{(r+1)}$.
Otherwise, set $a^{(r+1)}=a^{(r)}$.
Since the two updates average to $a^{(r)}$, in both cases
\[
 \mathbb E\bigl[a^{(r+1)}\mid a^{(r)}\bigr]=a^{(r)}.
\]
Each compression removes at least one nonzero coefficient,
so after at most $N$ steps no further changes occur.
Set $c'=a^{(N)}$. It has at most $m$ nonzero coordinates.
There are at most two choices at each step, so $c'$ takes
only finitely many values.
Taking expectations in the displayed identity repeatedly gives
\[
 \mathbb Ec'=\mathbb Ea^{(N)}
 =\mathbb Ea^{(N-1)}=\cdots=\mathbb Ea^{(0)}=c.
\]
To estimate the accumulated error, consider a realization with
$R$ updates. It holds that $a^{(R)}=a^{(N)}$. Let $n_0,\ldots,n_{R-1}$ be the support sizes
immediately before those updates.
Since $n_{r+1}\le\vartheta n_r$ and $n_{R-1}>m$,
\[
 n_r>m\vartheta^{-(R-1-r)},\qquad
 \sum_{r=0}^{R-1}n_r^{-\alpha}
 \le m^{-\alpha}\sum_{k=0}^{R-1}\vartheta^{\alpha k}
 \le\frac{m^{-\alpha}}{1-\vartheta^\alpha}.
\]
Since $a^{(0)}=c$ and $a^{(R)}=c'$, the triangle inequality gives
\[
 \begin{aligned}
 \|F-G\|_{C(B^d)}
 &\le \sum_{r=0}^{R-1}
 \left\|\sum_i(a_i^{(r)}-a_i^{(r+1)})\varphi_i\right\|_{C(B^d)}\\
 &\le C_d A\sum_{r=0}^{R-1}n_r^{-\alpha}
 \le \frac{C_d A}{1-\vartheta^\alpha}m^{-\alpha}.
 \end{aligned}
\]
Absorbing $(1-\vartheta^\alpha)^{-1}$ into $C_d$ gives the
required estimate. If $R=0$, then $G=F$.

For $1\le m<6$, choose an index $I$ with probability $c_I/A$
and set $c_I'=A$, with all other coefficients zero.
Then $\mathbb Ec_i'=c_i$, and the coefficient sum remains $A$.
Since $\|\varphi_i\|_{C(B^d)}\le2$, the error is at most $4A$.
Increasing $C_d$ covers these finitely many values of $m$.

We next allow repeated parameter pairs.
Partition the indices with positive coefficients into groups $I$
having identical parameter pairs, and replace each group by one
term with coefficient
\[
 C_I=\sum_{i\in I}c_i>0.
\]
Apply the preceding construction to the merged sum.
Let $B_I\ge0$ be the resulting random coefficient for group $I$,
so that $\mathbb EB_I=C_I$.

Conditional on the compressed coefficients, choose one index
$i\in I$ with probability $c_i/C_I$, assign $B_I$ to that index,
and set the other coefficients in the group to zero.
Since the parameters within each group are identical, this
leaves the compressed function unchanged.
Each nonzero $B_I$ is assigned to exactly one original index,
so the support bound and total coefficient sum are preserved.
Moreover,
\[
 \mathbb E[c_i'\mid B_I]=\frac{c_i}{C_I}B_I,
 \qquad
 \mathbb Ec_i'=\frac{c_i}{C_I}\mathbb EB_I=c_i.
\]
The compressed coefficients and the index choices both take
finitely many values, so the resulting coefficient vector
also takes only finitely many values.

Finally, for arbitrary signs, apply the nonnegative construction
separately to $(c_i)_+$ and $(-c_i)_+$, using budget $m$ for each,
and subtract the resulting sums to obtain $G$. 
The two constructions use only indices with $c_i>0$ and $c_i<0$,
respectively. Thus $G$ has at most $2m$ nonzero coefficients,
each with the same sign as the corresponding $c_i$, and
\[
 \sum_i|c_i'|
 =\sum_i\bigl((c_i)_++(-c_i)_+\bigr)
 =A.
\]
Subtracting the coefficient expectations gives
$\mathbb Ec_i'=(c_i)_+-(-c_i)_+=c_i$.
Finally, the triangle inequality gives
\[
 \|F-G\|_{C(B^d)}
 \le C_d A m^{-\alpha}.
\]
This completes the proof.
\end{proof}

The next lemma gives a deterministic approximation using at most
$m$ original neurons. Both the total absolute coefficient sum and the prescribed
weighted sum remain bounded by their original values.

\begin{lemma}
\label{lem:sparsify}
Let
\[
 F(z)=\sum_{i=1}^N c_i(\omega_i\cdot z+b_i)_+,
 \qquad |\omega_i|=1,\quad |b_i|\le1,\quad
 A=\sum_i|c_i|.
\]
For fixed weights $k_i\ge0$ and any integer $m\ge0$, there exists
a deterministic sum
\[
 G_m(z)=\sum_{i=1}^N c_i'(\omega_i\cdot z+b_i)_+
\]
using at most $m$ original indices, such that
\begin{equation}\label{core:eq-sparse}
 \begin{aligned}
 \|F-G_m\|_{C(B^d)}
 &\le C_d A(m+1)^{-\alpha},\\
 \sum_i|c_i'|
 &\le A,\qquad
 \sum_i k_i|c_i'|\le\sum_i k_i|c_i|,
 \end{aligned}
\end{equation}
where $\alpha=(d+3)/(2d)$ and $C_d$ depends only on $d$.
\end{lemma}

\begin{proof}
For $m\ge4$, set $r=\lfloor m/4\rfloor$.
Apply Proposition~\ref{rand:uniform} with $r$ in place of $m$,
and then apply Lemma~\ref{rand:cost-mixing} with weights $k_i$. The first construction uses at most $2r$ original indices;
the second increases this number to at most $4r\le m$.
The common uniform error bound $C_d A r^{-\alpha}$ is preserved,
and every resulting realization satisfies
\[
 \sum_i|c_i'|=A,\qquad
 \sum_i k_i|c_i'|=\sum_i k_i|c_i|.
\]
Choose any such realization as $G_m$.
Since $m+1\le4r+4\le8r$, its error satisfies
\[
 \|F-G_m\|_{C(B^d)}
 \le C_d A r^{-\alpha}
 \le 8^\alpha C_d A(m+1)^{-\alpha}.
\]
Absorbing $8^\alpha$ into $C_d$ gives the required estimate.

For $0\le m<4$, take $G_m=0$.
The coefficient inequalities are immediate, and
\[
 \|F\|_{C(B^d)}
 \le2A
 \le2\cdot4^\alpha A(m+1)^{-\alpha}.
\]
Thus the same estimate holds after enlarging $C_d$.
\end{proof}

\subsection{Local Approximation}
\label{geom:section}

We now derive local approximation estimates for groups of atoms whose caps lie in a common convex set with volume comparable to that of each cap. 

Recall from \eqref{eq:domain-atoms} that, for
$z=(\xi,t)\in Z_K$, the atom and its associated geometric
quantities are
\[
 \begin{aligned}
 \phi_z(x)&=(\xi\cdot x-t)_+,
 & h_z&=H_K(\xi)-t=\max_{x\in K}\xi\cdot x - t,\\
 C_z&=K\cap\{x:\xi\cdot x\ge t\},\quad
 & \theta_z&=\mu_K(C_z)=\frac{|C_z|}{|K|}.
 \end{aligned}
\]
Here $C_z$ is the cap, $h_z$ its height, and $\theta_z$ its
relative volume. For a family indexed by $z_i=(\xi_i,t_i)$,
we write $\phi_i=\phi_{z_i}$, $h_i=h_{z_i}$,
$C_i=C_{z_i}$, and $\theta_i=\theta_{z_i}$.
When comparing different domains, we write
$h_K(z)=h_z$ and $\theta_K(z)=\theta_z$ to make the
dependence on $K$ explicit.
The weight from \eqref{eq:intrinsic-weight} is
\begin{equation*}
 w_{K,\eta}(z)=h_z\theta_z^\eta.
\end{equation*}
The following proposition proves the atom norm estimate stated
earlier and shows that invertible affine changes of variables
preserve the weighted approximation errors.

\begin{proposition}
\label{geom:atoms}
Let $K\subset\mathbb R^d$ be a convex body.
For $z\in Z_K$ and $1\le q<\infty$,
\begin{equation}\label{geom:eq-atomnorm}
 2^{-1-d/q}h_z\theta_z^{1/q}
 \le \|\phi_z\|_{L^q(\mu_K)}
 \le h_z\theta_z^{1/q}.
\end{equation}
Also, $\|\phi_z\|_{C(K)}=h_z$.
Consequently, for $\eta\in\mathbb R$ and $1\le q\le\infty$,
the normalized dictionary
\[
 \{\phi_z/w_{K,\eta}(z):z\in Z_K\}
\]
is bounded in $X_q(K)$ if and only if $\eta\le1/q$,
with the convention $1/\infty=0$.

Let $Q\subset\mathbb R^d$ be another convex body and let
$T(y)=Ay+b$ be an invertible affine map with $T(Q)=K$.
The map $f\mapsto f\circ T$ sends $\Sigma_n^K$ bijectively
onto $\Sigma_n^Q$ for every integer $n\ge0$.
For every $\eta\in\mathbb R$, $1\le q\le\infty$,
and integer $n\ge0$,
\[
 \mathcal E_n^K(w_{K,\eta};q)
 =\mathcal E_n^Q(w_{Q,\eta};q).
\]
If $\eta\le1/q$, the same map is an isometric bijection from
$\mathcal V_{w_{K,\eta}}^{(q)}(K)$ onto
$\mathcal V_{w_{Q,\eta}}^{(q)}(Q)$.
\end{proposition}

\begin{proof}
The inequality $0\le\phi_z\le h_z\mathbf1_{C_z}$ on $K$
gives the upper bound.
Choose $v\in K$ with $\xi\cdot v=H_K(\xi)$ and set
\[
 D=\{(v+x)/2:x\in C_z\}.
\]
Convexity gives $D\subset K$, and
$\mu_K(D)=2^{-d}\theta_z$.
For $x\in C_z$,
\[
 \phi_z\!\left(\frac{v+x}{2}\right)
 =\frac{h_z+\xi\cdot x-t}{2}
 \ge\frac{h_z}{2}.
\]
Hence
\[
 \|\phi_z\|_{L^q(\mu_K)}^q
 \ge 2^{-d}\theta_z(h_z/2)^q,
\]
which proves the lower bound.
Since $\phi_z(v)=h_z$, the uniform norm equals $h_z$.

Dividing by $w_{K,\eta}(z)=h_z\theta_z^\eta$ shows that
the normalized atom norm is comparable to
$\theta_z^{1/q-\eta}$ for finite $q$, and equals
$\theta_z^{-\eta}$ for $q=\infty$.
For fixed $\xi$, as $t\uparrow H_K(\xi)$, the caps $C_z$
decrease to $K\cap\{x:\xi\cdot x=H_K(\xi)\}$. So $\theta_z\to0$.
Thus, the preceding norm estimates show that
the normalized dictionary is bounded in $X_q(K)$ if and only if
$\eta\le1/q$.

We now consider the affine map $T(y)=Ay+b$, with $T(Q)=K$. For $z=(\xi,t)\in Z_K$, set
\[
 \lambda=|A^T\xi|>0,\qquad
 \xi'=\frac{A^T\xi}{\lambda},\qquad
 t'=\frac{t-\xi\cdot b}{\lambda},\qquad
 z'=(\xi',t').
\]
Then
\[
 \xi\cdot T(y)-t=\lambda(\xi'\cdot y-t').
\]
Since $T(Q)=K$, taking minima and maxima over $Q$ gives $-H_Q(-\xi')\le t'<H_Q(\xi')$. Together with $|\xi'|=1$, this proves $z'\in Z_Q$. The same identity shows that $T$ maps the cap $Q\cap\{y:\xi'\cdot y\ge t'\}$ onto $C_z$. Since both cap and body volumes are multiplied by $|\det A|$, we obtain
\begin{equation}\label{geom:eq-transport}
 \begin{aligned}
 \phi_z\circ T&=\lambda\phi_{z'},&
 h_K(z)&=\lambda h_Q(z'),\\
 \theta_K(z)&=\theta_Q(z'),&
 w_{K,\eta}(z)&=\lambda w_{Q,\eta}(z').
 \end{aligned}
\end{equation}
In particular,
\[
 \left(\frac{\phi_z}{w_{K,\eta}(z)}\right)\circ T
 =\frac{\phi_{z'}}{w_{Q,\eta}(z')}.
\]
Applying the same construction to $T^{-1}$ gives the reverse correspondence between normalized atoms. By linearity, $f\mapsto f\circ T$ therefore maps $\mathcal A_{w_{K,\eta}}(K)$ bijectively onto $\mathcal A_{w_{Q,\eta}}(Q)$. Also, composing a ReLU sum with either $T$ or $T^{-1}$ preserves its number of terms, so the same map sends $\Sigma_n^K$ bijectively onto $\Sigma_n^Q$.

For every $f\in X_q(K)$,
\[
 \|f\circ T\|_{X_q(Q)}=\|f\|_{X_q(K)},
 \quad\forall1\le q\le\infty.
\]
Applying this equality to $f-S$, then taking the infimum over $S\in\Sigma_n^K$ and the supremum over $f\in\mathcal A_{w_{K,\eta}}(K)$, proves $\mathcal E_n^K(w_{K,\eta};q)
 =\mathcal E_n^Q(w_{Q,\eta};q)$.

For $\eta\le1/q$, taking $X_q$ closures shows that $f\mapsto f\circ T$ maps the variation unit ball on $K$ onto that on $Q$. Linearity and \eqref{eq:variation-norm} then give equality of the variation norms.
\end{proof}

The next lemma adapts Lemma~\ref{lem:sparsify} to sums of atoms whose caps lie in a common convex set with volume comparable to that of each cap. It gives a local approximation using original atoms without increasing the prescribed weighted coefficient sum.

\begin{lemma}\label{geom:local}
Let $\phi_i=\phi_{z_i}$, with $z_i=(\xi_i,t_i)\in Z_K$, be finitely many atoms on $K$, with heights $h_i$ and caps $C_i$. Suppose that a convex body $E\subset K$ satisfies
\begin{equation}\label{geom:eq-envelope}
 C_i\subset E,\qquad |E|\le A|C_i|\quad\text{for every }i,
 \qquad A\ge1.
\end{equation}
For $F=\sum_i c_i\phi_i$, fixed weights $w_i>0$, and any integer $m\ge0$, there is a sum $S_m=\sum_i c_i'\phi_i$ using at most $m$ original indices such that
\begin{align}
 |F-S_m|&\le C_{d,A}(m+1)^{-\alpha}
       \Bigl(\sum_i|c_i|h_i\Bigr)\mathbf1_E
       \qquad\text{on }K,\label{geom:eq-localpoint}\\
 \sum_i|c_i'|w_i&\le\sum_i|c_i|w_i.
       \label{geom:eq-localcost}
\end{align}
In particular, for $\eta\in\mathbb R$ and $1\le q<\infty$, suppose that the prescribed weights satisfy $w_i=h_i\theta_i^\eta$, and set $\Theta=\mu_K(E)$ and $M=\sum_i|c_i|w_i$. Then
\begin{equation}\label{geom:eq-localq}
 \|F-S_m\|_{L^q(\mu_K)}
 \le C_{d,q,A,\eta}\Theta^{1/q-\eta}M(m+1)^{-\alpha}.
\end{equation}
\end{lemma}

\begin{proof}
By John's ellipsoid theorem~\cite{John1948}, $E$ contains an ellipsoid whose dilation by a factor of $d$ about its center contains $E$. Choose an invertible affine map $T(y)=x_0+Ry$ that maps $B^d$ onto the dilated ellipsoid. Then $T^{-1}$ maps the original ellipsoid onto $d^{-1}B^d$, so
\[
 d^{-1}B^d\subset E':=T^{-1}(E)\subset B^d.
\]
Set $C_i'=T^{-1}(C_i)$, $u_i=R^T\xi_i/h_i$, and $b_i=(\xi_i\cdot x_0-t_i)/h_i$. Then
\[
 \frac{\phi_i(T(y))}{h_i}=(u_i\cdot y+b_i)_+.
\]
For each $i$, choose $p_i,r_i\in C_i$ with $\xi_i\cdot p_i=t_i$ and $\xi_i\cdot r_i=H_K(\xi_i)$, and set $y_0=T^{-1}(p_i)$, $y_1=T^{-1}(r_i)$. Since $C_i\subset E\subset K$, these points lie in $E'$, where $u_i\cdot y+b_i$ vanishes at $y_0$ and attains its maximum value one at $y_1$. Both points lie in $B^d$, so $|y_1-y_0|\le2$ and $|y_0|\le1$. Thus $1=u_i\cdot(y_1-y_0)\le2|u_i|$ and $|b_i|=|u_i\cdot y_0|\le|u_i|$, yielding
\begin{equation}\label{geom:eq-slope-lower}
 |u_i|\ge\tfrac12,\qquad |b_i|\le|u_i|.
\end{equation}
The set $C_i'$ lies in $B^d$ between the parallel planes $u_i\cdot y+b_i=0$ and $u_i\cdot y+b_i=1$, whose distance is $1/|u_i|$. Its volume is therefore at most $C_d/|u_i|$. Combining this with \eqref{geom:eq-envelope} gives
\[
 \frac{|B^d|}{A d^d}
 \le\frac{|E'|}{A}
 \le |C_i'|
 \le\frac{C_d}{|u_i|},
\]
so $|u_i|\le C_{d,A}$. Set $\lambda_i=|u_i|$, $\nu_i=u_i/\lambda_i$, and $\beta_i=b_i/\lambda_i$. Then
\[
 \phi_i(T(y))=h_i\lambda_i(\nu_i\cdot y+\beta_i)_+,
 \qquad
 \tfrac12\le\lambda_i\le C_{d,A},\quad
 |\nu_i|=1,\quad |\beta_i|\le1.
\]

Apply Lemma~\ref{lem:sparsify} with the same integer $m$ to the sum $\sum_i a_i(\nu_i\cdot y+\beta_i)_+$ obtained by substituting $x=T(y)$ into $F(x)$, with coefficients $a_i=c_i h_i\lambda_i$ and weights $k_i=w_i/(h_i\lambda_i)$. If $a_i'$ are the resulting coefficients, set $c_i'=a_i'/(h_i\lambda_i)$ and $S_m=\sum_i c_i'\phi_i$. At most $m$ coefficients are nonzero, and
\[
 \sum_i|c_i'|w_i
 =\sum_i|a_i'|k_i
 \le\sum_i|a_i|k_i
 =\sum_i|c_i|w_i.
\]
Since $E'\subset B^d$ and $\sum_i|a_i|\le C_{d,A}\sum_i|c_i|h_i$, the uniform error estimate gives \eqref{geom:eq-localpoint} on $E$. Both $F$ and $S_m$ vanish on $K\setminus E$, because they use only atoms whose caps are contained in $E$. Thus \eqref{geom:eq-localpoint} holds throughout $K$.

If the prescribed weights satisfy $w_i=h_i\theta_i^\eta$, then $\Theta/A\le\theta_i\le\Theta$ gives
\[
 \sum_i|c_i|h_i
 =\sum_i|c_i|w_i\theta_i^{-\eta}
 \le C_{A,\eta}\Theta^{-\eta}M.
\]
Taking the $L^q(\mu_K)$ norm in \eqref{geom:eq-localpoint} and using $\|\mathbf1_E\|_{L^q(\mu_K)}=\Theta^{1/q}$ proves \eqref{geom:eq-localq}.
\end{proof}

The next proposition gives a random version of Lemma~\ref{geom:local} that preserves each coefficient in expectation and the prescribed weighted coefficient sum in every realization.
\begin{proposition}
\label{rand:local}
Let $\phi_i=\phi_{z_i}$, $z_i\in Z_K$, be finitely many atoms on $K$, with heights $h_i$ and caps $C_i$. Suppose that a convex body $E\subset K$ and a constant $A\ge1$ satisfy
\[
 C_i\subset E,\qquad |E|\le A|C_i|
 \quad\text{for every }i.
\]
Let $F=\sum_i c_i\phi_i$ and fix weights $w_i>0$. For every integer $m\ge1$, there is a random coefficient vector $c'=(c_i')$ taking only finitely many values, with $\mathbb Ec_i'=c_i$ for every $i$. In every realization, the sum $G_m=\sum_i c_i'\phi_i$ satisfies
\begin{align}
 |F-G_m|&\le C_{d,A}m^{-\alpha}
       \Bigl(\sum_i|c_i|h_i\Bigr)\mathbf1_E
       \qquad\text{on }K,
 \label{rand:eq-local-point}\\
 |\operatorname{supp}c'|&\le4m,\qquad
 \sum_i|c_i'|w_i=\sum_i|c_i|w_i.
 \label{rand:eq-local-mass}
\end{align}
Each realization also satisfies $c_i'=0$ whenever $c_i=0$, and $c_i'c_i\ge0$ for every $i$.

For $\eta\in\mathbb R$ and $1\le q<\infty$, suppose that the
prescribed weights satisfy $w_i=h_i\theta_i^\eta$, and set
\[
 \Theta=\mu_K(E),\qquad M=\sum_i|c_i|w_i.
\]
Then
\begin{equation}\label{rand:eq-local-envelope}
 \begin{aligned}
 \mathbb E G_m&=F,\qquad
 \|\Theta^{-1/q}\mathbf1_E\|_{L^q(\mu_K)}=1,\\
 |F-G_m|&\le C_{d,A,\eta}\Theta^{-\eta}M m^{-\alpha}\mathbf1_E
 \qquad\text{on }K,
 \end{aligned}
\end{equation}
where the error bound holds in every realization.
\end{proposition}

\begin{proof}
As in the proof of Lemma~\ref{geom:local}, choose an invertible affine map $T$ with $E'=T^{-1}(E)\subset B^d$ such that
\[
 \phi_i(T(y))=h_i\lambda_i(\nu_i\cdot y+\beta_i)_+,
 \qquad
 \tfrac12\le\lambda_i\le C_{d,A},\quad
 |\nu_i|=1,\quad |\beta_i|\le1.
\]
Apply Proposition~\ref{rand:uniform} with the given integer $m$ to the sum $\sum_i a_i(\nu_i\cdot y+\beta_i)_+$ obtained by substituting $x=T(y)$ into $F(x)$, where $a_i=c_i h_i\lambda_i$. The resulting random coefficients have expectations $a_i$, retain their signs, and have at most $2m$ nonzero entries in every realization.

Apply Lemma~\ref{rand:cost-mixing} to this random coefficient vector with $W=2m$ and weights $k_i=w_i/(h_i\lambda_i)$. Denote its output coefficients by $a_i'$. They take only finitely many joint values, satisfy $\mathbb Ea_i'=a_i$, and have at most $4m$ nonzero entries. Lemma~\ref{rand:cost-mixing} also preserves the coefficient signs and the uniform error bound from Proposition~\ref{rand:uniform}. Thus, in every realization,
\[
 \left\|\sum_i(a_i-a_i')(\nu_i\cdot y+\beta_i)_+\right\|_{C(B^d)}
 \le C_d m^{-\alpha}\sum_i|a_i|.
\]

Set $c_i'=a_i'/(h_i\lambda_i)$ and $G_m=\sum_i c_i'\phi_i$. Since $h_i\lambda_i>0$, at most $4m$ coefficients $c_i'$ are nonzero, and each nonzero $c_i'$ has the same sign as $c_i$. Moreover, $\mathbb Ec_i'=c_i$. The weighted identity from Lemma~\ref{rand:cost-mixing} gives
\[
 \sum_i|c_i'|w_i
 =\sum_i|a_i'|k_i
 =\sum_i|a_i|k_i
 =\sum_i|c_i|w_i.
\]
For $x=T(y)\in E$, we have $y\in E'\subset B^d$. The atom identity above and the definitions of $a_i$ and $c_i'$ give
\[
 \begin{aligned}
 |F(x)-G_m(x)|
 &=\left|\sum_i(a_i-a_i')(\nu_i\cdot y+\beta_i)_+\right|\\
 &\le C_d m^{-\alpha}\sum_i|a_i|
 \le C_{d,A}m^{-\alpha}\sum_i|c_i|h_i.
 \end{aligned}
\]
Both sums vanish on $K\setminus E$, because they use only atoms whose caps are contained in $E$. This proves \eqref{rand:eq-local-point}.

If the prescribed weights satisfy $w_i=h_i\theta_i^\eta$, then $\Theta/A\le\theta_i\le\Theta$ gives
\[
 \sum_i|c_i|h_i
 =\sum_i|c_i|w_i\theta_i^{-\eta}
 \le C_{A,\eta}\Theta^{-\eta}M.
\]
Combining the preceding estimate with
\eqref{rand:eq-local-point} gives the pointwise bound in
\eqref{rand:eq-local-envelope}. Finally, $\mathbb EG_m=F$ follows from $\mathbb Ec_i'=c_i$.
\end{proof}

\subsection{Cap Coverings and Floating Depth Estimates}
\label{rand:geometry}

We choose covering caps at each volume scale and establish depth estimates for combining local errors. Throughout this subsection, $K\subset\mathbb R^d$ is an arbitrary convex body and $\mu_K$ is its normalized volume measure. For the definitions of floating depth $v_K$ and wet-part volume $W_K$, see Subsection~\ref{sec:geometric-guide}.

\begin{lemma}
\label{hil:economic-cover}
There exist constants $c_d,C_d>0$ and $\tau_0\in(0,1)$, depending only on $d$, such that for every convex body $K\subset\mathbb R^d$ and every $0<\tau<\tau_0$, there are caps $E_1,\ldots,E_N$ of $K$ and pairwise disjoint convex sets $D_1,\ldots,D_N$ satisfying
\begin{equation}\label{rand:eq-economic-cover}
 \begin{gathered}
 D_i\subset E_i,\qquad
 c_d\tau\le\mu_K(D_i)\le\mu_K(E_i)\le C_d\tau,
 \qquad\forall 1\le i\le N,\\
 \bigcup_{i=1}^N D_i
 \subset\{x\in K:v_K(x)\le\tau\}
 \subset\bigcup_{i=1}^N E_i.
 \end{gathered}
\end{equation}
Every cap $C$ of $K$ with $\mu_K(C)\le\tau$ is contained in some $E_i$. Moreover,
\[
 c_d\mathcal C_K(\tau)\le N\le C_d\mathcal C_K(\tau).
\]
\end{lemma}

\begin{proof}
Rescale $K$ by the factor $|K|^{-1/d}$ to obtain a convex body of volume one. Scaling preserves relative volumes and maps caps to caps, so any sets $E_i,D_i$ satisfying the conclusions for the rescaled body give the required sets in $K$ after scaling back. We may therefore assume $|K|=1$. Apply the economic cap-covering theorem of B\'ar\'any and Larman~\cite{BaranyLarman1988}, as stated in \cite[Theorem~4]{BaranyFodorVigh2010}, with $t=\tau$, taking $\tau_0$ to be the dimensional threshold in that theorem. Denote its covering caps by $E_i$ and its pairwise disjoint convex sets by $D_i$. Then $D_i\subset E_i$, and parts (i) and (ii) of the theorem give \eqref{rand:eq-economic-cover}.

If $C$ is a cap with $\mu_K(C)\le\tau$, then the definition of $v_K$ gives $v_K(x)\le\mu_K(C)\le\tau$ for every $x\in C$. Thus $C\subset\{v_K\le\tau\}$, and part (iii) of Theorem~4 in \cite{BaranyFodorVigh2010} gives $C\subset E_i$ for some $i$.

Finally, the disjointness of the sets $D_i$ and the covering by the caps $E_i$, together with their volume bounds, give
\[
 \begin{aligned}
 c_dN\tau
 &\le\sum_i\mu_K(D_i)
 =\mu_K\!\left(\bigcup_iD_i\right)
 \le W_K(\tau),\\
 W_K(\tau)
 &\le\mu_K\!\left(\bigcup_iE_i\right)
 \le\sum_i\mu_K(E_i)
 \le C_dN\tau.
 \end{aligned}
\]
Since $\mathcal C_K(\tau)=W_K(\tau)/\tau$, these inequalities give
\[
 C_d^{-1}\mathcal C_K(\tau)\le N\le c_d^{-1}\mathcal C_K(\tau).
\]
This completes the proof.
\end{proof}
The pairwise disjoint sets $D_i$ will be used to bound the number of covering caps containing a given point. These inner sets need not be caps.


The next lemma gives an upper bound for the relative volume of the wet part, with a constant depending only on the dimension.

\begin{lemma}
\label{rand:wet-bound}
\label{rand:wet-volume}
For every integer $m\ge1$, there is a constant $C_m>0$, depending only on $m$, such that every convex body $D\subset\mathbb R^m$ satisfies
\begin{equation}\label{rand:eq-uniform-wet}
 W_D(t)\le C_m t^{\beta_m},
 \qquad \beta_m=\frac{2}{m+1},
 \qquad 0<t\le1.
\end{equation}
Consequently, every convex body $K\subset\mathbb R^d$ with $d\ge2$ satisfies
\[
 \mathcal C_K(t)\le C_dt^{-(d-1)/(d+1)},
 \qquad 0<t\le1.
\]
\end{lemma}

\begin{proof}
For $m=1$, the body $D$ is an interval. Its wet part is the union of its two endpoint subintervals of relative length $t$, so
\[
 W_D(t)=\min\{2t,1\}\le2t.
\]
This proves \eqref{rand:eq-uniform-wet} since $\beta_1=1$.

Now suppose $m\ge2$. Rescaling preserves relative cap volumes and $W_D(t)$, so we may assume $|D|=1$. Choose $n$ points $X_1,\ldots,X_n$ independently and uniformly from $D$. Their convex hull $\operatorname{conv}\{X_1,\ldots,X_n\}$ is the smallest convex set containing these points and is contained in $D$. Define
\[
 \Delta_D(n)
 =\mathbb E\left[
   \left|D\setminus\operatorname{conv}\{X_1,\ldots,X_n\}\right|
 \right].
\]
Thus $\Delta_D(n)$ is the average volume of the part of $D$ not covered by the random convex hull, where the expectation is taken over the sampled points.

For $n\ge2$, each point $x$ in the wet part at threshold $\tau=1/n$ belongs to a cap of volume at most $1/n$. The probability that all $n$ sampled points lie outside this cap is at least $(1-1/n)^n\ge1/4$. On this event, the sampled points lie in the opposite open halfspace, so their convex hull cannot contain $x$. Taking expectations and applying Fubini's theorem, we obtain
\[
 \begin{aligned}
 \Delta_D(n)
 &=\int_D
 \mathbb P\!\left(x\notin\operatorname{conv}\{X_1,\ldots,X_n\}\right)\,dx\\
 &\ge\int_{\{x\in D:v_D(x)\le1/n\}}\frac14\,dx
 =\frac14 W_D(1/n).
 \end{aligned}
\]
The inequality uses the preceding probability bound on the wet part, and the last equality uses $|D|=1$.

Groemer's comparison \cite[(1.5)]{BaranyLarman1988} shows that, among convex bodies $D$ of volume one, $\Delta_D(n)$ is maximized by a Euclidean ball. Combining the estimate for the unit ball in \cite[(1.4)]{BaranyLarman1988}, rescaled to volume one, with the preceding probability bound gives
\[
 W_D(1/n)\le4\Delta_D(n)\le C_m n^{-\beta_m}
\]
for all sufficiently large $n$, where both $C_m$ and the required lower bound on $n$ depend only on $m$. By increasing $C_m$ if necessary, this bound holds for every integer $n\ge1$.

For any $0<t\le1$, take $n=\lfloor1/t\rfloor$. Then $t\le1/n$ and $n\ge1/(2t)$. Since $W_D$ is nondecreasing,
\[
 W_D(t)\le W_D(1/n)
 \le C_m n^{-\beta_m}
 \le 2^{\beta_m}C_m t^{\beta_m}.
\]
This proves \eqref{rand:eq-uniform-wet}. Finally, applying this bound with $D=K$ and $m=d$ gives
\[
 \mathcal C_K(t)=\frac{W_K(t)}{t}
 \le C_dt^{\beta_d-1}
 =C_dt^{-(d-1)/(d+1)}.
\]
This completes the proof.
\end{proof}

The next lemma derives local estimates for $v_K$ from the uniform wet-volume bound.

\begin{lemma}
\label{rand:depth-moments}
Let $E\subset K$ be a convex body with relative volume $V=\mu_K(E)>0$, and let $\beta_d=2/(d+1)$. Then, for every $R\ge1$,
\begin{equation}
\label{rand:eq-local-depth-tail}
 \mu_K\{x\in E:v_K(x)<V/R\}
 \le C_d V R^{-\beta_d}.
\end{equation}
For every $0<a<\beta_d$,
\begin{equation}
\label{rand:eq-negative-depth}
 \int_E v_K(x)^{-a}\,d\mu_K(x)
 \le C_d\left(1+\frac{a}{\beta_d-a}\right)V^{1-a}.
\end{equation}
Here $C_d$ depends only on $d$.
\end{lemma}

\begin{proof}
Let $\mu_E=|E|^{-1}dx$ on $E$, and let $v_E$ be the floating depth defined using caps of $E$ and their relative volumes in $E$. For every $x\in E$ and every halfspace $H$ containing $x$,
\[
 \mu_K(K\cap H)
 \ge \mu_K(E\cap H)
 =V\mu_E(E\cap H).
\]
Taking the infimum over all such halfspaces gives $v_K(x)\ge Vv_E(x)$. Since $d\mu_K=V\,d\mu_E$ on $E$, Lemma \ref{lem:wet-part-depth} and Lemma~\ref{rand:wet-bound} yield
\[
 \begin{aligned}
 \mu_K\{x\in E:v_K(x)<V/R\}
 &\le V\mu_E\{x\in E:v_E(x)<1/R\}\\
 &\le V W_E(1/R)
 \le C_d V R^{-\beta_d},
 \end{aligned}
\]
which proves \eqref{rand:eq-local-depth-tail}.

Since $W_E(t)\to0$ as $t\downarrow0$, we have $\mu_E\{v_E=0\}=0$. For $0<a<\beta_d$, apply the identity $s^{-a}=1+a\int_s^1t^{-a-1}\,dt$ to $s=v_E(x)$, which lies in $(0,1]$ almost everywhere. Integrating over $E$ and interchanging the nonnegative integrals gives
\[
 \begin{aligned}
 \int_E v_E(x)^{-a}\,d\mu_E(x)
 &=1+a\int_0^1 t^{-a-1}
       \mu_E\{x\in E:v_E(x)<t\}\,dt\\
 &\le1+a\int_0^1 t^{-a-1}W_E(t)\,dt\\
 &\le1+C_d a\int_0^1 t^{\beta_d-a-1}\,dt\\
 &=1+\frac{C_d a}{\beta_d-a}.
 \end{aligned}
\]
Here Lemma~\ref{rand:wet-bound} gives $W_E(t)\le C_dt^{\beta_d}$ with $C_d$ depending only on $d$, independently of $E$ and $K$.
Using $v_K\ge Vv_E$ and $d\mu_K=V\,d\mu_E$ once more, we obtain
\[
 \begin{aligned}
 \int_E v_K(x)^{-a}\,d\mu_K(x)
 &\le V^{1-a}\int_E v_E(x)^{-a}\,d\mu_E(x)\\
 &\le C_d\left(1+\frac{a}{\beta_d-a}\right)V^{1-a},
 \end{aligned}
\]
after enlarging $C_d$ if necessary. This proves \eqref{rand:eq-negative-depth}.
\end{proof}

\begin{lemma}
\label{geom:depth-synthesis}
Let $(E_B)$ be a finite or countable family of convex bodies contained in $K$, and set $V_B=\mu_K(E_B)>0$. For $\gamma>0$, suppose that
\begin{equation}\label{geom:eq-weighted-overlap}
 H_\gamma
 :=\mathop{\mathrm{ess\,sup}}_{x\in K}
 \sum_B\left(\frac{v_K(x)}{V_B}\right)^\gamma
 \mathbf1_{E_B}(x)
 <\infty,
\end{equation}
where the essential supremum is taken with respect to $\mu_K$. If $1<s<\infty$ and $a=\gamma(s-1)<\beta_d$, then every family $(c_B)\subset\mathbb{R}$ with only finitely many nonzero entries satisfies
\[
 \left\|\sum_B c_BV_B^{-1/s}\mathbf1_{E_B}\right\|_{L^s(\mu_K)}
 \le C_{d,a}^{1/s}H_\gamma^{1-1/s}
      \left(\sum_B|c_B|^s\right)^{1/s},
\]
where $C_{d,a}$ depends only on $d$ and $a$.

For $s=1$, the bound
\[
 \left\|\sum_B c_BV_B^{-1}\mathbf1_{E_B}\right\|_{L^1(\mu_K)}
 \le \sum_B|c_B|
\]
holds without the assumption $H_\gamma<\infty$.
\end{lemma}

\begin{proof}
It suffices to prove the estimate with $c_B$ replaced by $|c_B|$. Since $v_K>0$ almost everywhere, consider a point $x$ with positive floating depth $v_K(x)$. Using $a=\gamma(s-1)$, write
\[
 |c_B|V_B^{-1/s}
 =
 \left(|c_B|^sV_B^{a-1}v_K(x)^{-a}\right)^{1/s}
 \left(\frac{v_K(x)}{V_B}\right)^{\gamma(1-1/s)}.
\]
Apply H\"older's inequality with exponents $s$ and $s/(s-1)$ to the sum over indices for which $x\in E_B$. The definition of $H_\gamma$ then gives, almost everywhere,
\[
 \left(\sum_B|c_B|V_B^{-1/s}\mathbf1_{E_B}(x)\right)^s
 \le H_\gamma^{s-1}
      \sum_B|c_B|^sV_B^{a-1}v_K(x)^{-a}\mathbf1_{E_B}(x).
\]

Lemma~\ref{rand:depth-moments}, applied to each $E_B$, gives
\[
 \int_{E_B}v_K(x)^{-a}\,d\mu_K(x)
 \le C_{d,a}V_B^{1-a},
 \qquad
 C_{d,a}=C_d\left(1+\frac{a}{\beta_d-a}\right).
\]
Integrating the pointwise estimate therefore yields
\[
 \begin{aligned}
 \left\|\sum_B c_BV_B^{-1/s}\mathbf1_{E_B}\right\|_{L^s(\mu_K)}^s
 &\le H_\gamma^{s-1}
       \sum_B|c_B|^sV_B^{a-1}
       \int_{E_B}v_K(x)^{-a}\,d\mu_K(x)\\
 &\le C_{d,a}H_\gamma^{s-1}\sum_B|c_B|^s.
 \end{aligned}
\]
Taking the $s$-th root proves the assertion for $s>1$.

For $s=1$, the triangle inequality and
$\|V_B^{-1}\mathbf1_{E_B}\|_{L^1(\mu_K)}=1$
give the stated bound.
\end{proof}

\subsection{A Technical Lemma on Cap Coverings}

A common upper bound on cap volumes does not prevent individual caps from becoming arbitrarily thin. The next lemma shows that, for a finite family of caps of a convex body $Q$, requiring each cap to contain a ball of a fixed radius gives an upper bound on the volume of the union of all caps in the family. 
For $d\ge3$, the proof of Proposition~\ref{rand:visibility} applies this lemma to caps in a section of dimension $d-1$. This gives \eqref{rand:eq-slice-cover}, the bound used in the second integral in Step 4 of that proof.

\begin{lemma}
\label{rand:cap-cover}
Fix $m\ge2$, $c>0$, and $S>0$. Let $Q\subset\mathbb R^m$ be a convex body, and let $\mathcal A$ be a finite family of caps of $Q$. Suppose that each $C\in\mathcal A$ contains a Euclidean ball of radius $c$ and satisfies $|C|\le S$. Then there are caps $F_1,\ldots,F_N$ of $Q$ such that every $C\in\mathcal A$ is contained in some $F_i$, each $|F_i|\le2^mS$, and
\begin{equation}
\label{rand:eq-thick-cover-count}
 N\le C_{m,c}S\log(1+S),
\end{equation}
where $C_{m,c}$ depends only on $m$ and $c$. In particular,
\begin{equation}
\label{rand:eq-thick-union}
 \left|\bigcup_{C\in\mathcal A}C\right|
 \le C_{m,c}S^2\log(1+S).
\end{equation}
\end{lemma}

\begin{proof}
For each $C\in\mathcal A$, we will construct a set $G_C$ of unit normal directions such that every direction in $G_C$ admits a cap of $Q$ containing $C$ and having volume at most $2^mS$. We will then select finitely many directions so that, for every $C\in\mathcal A$, at least one selected direction belongs to $G_C$. Whenever a selected direction $\theta$ belongs to $G_C$ for one or more caps $C\in\mathcal A$, choose the largest cap of $Q$ in direction $\theta$ with volume at most $2^mS$, such that it contains all the chosen containing caps and hence every original cap $C$ with $\theta\in G_C$. Thus at most one covering cap is needed for each selected direction.

To bound the number of directions needed, we will prove that the set $G_C$ constructed for each cap $C\in\mathcal A$ has normalized spherical measure at least $a_{m,c}/S$, where $a_{m,c}>0$ depends only on $m,c$. We will also verify the hypothesis of Lemma~\ref{rand:epsilon-net}. That lemma will then give a set of at most $C_{m,c}S\log(1+S)$ directions meeting every $G_C$, independently of the number of original caps, namely, $|\mathcal{A}|$.

\medskip
\noindent\textit{Step 1. Constructing the direction sets.}
If $\mathcal A$ is empty, take $N=0$. Otherwise, $S\ge|cB^m|$, where $B^m$ denotes the Euclidean unit ball in $\mathbb R^m$. Fix $C\in\mathcal A$ and write
\[
 C=Q\cap\{x:u\cdot x\ge b\},
 \qquad |u|=1.
\]
Here $b\in\mathbb R$ determines the position of the cutting plane $\{x:u\cdot x=b\}$. We will perturb the normal $u$ and adjust the cutting plane to obtain a cap containing $C$ with volume at most $2^mS$. For $\delta>0$ and a vector $z$ perpendicular to $u$, replacing $u$ by $u+\delta z$ changes the left-hand side of the defining inequality to
\[
 (u+\delta z)\cdot x=u\cdot x+\delta z\cdot x.
\]
To determine how to adjust the cutting plane, fix a point $y\in C$ and write $\delta z\cdot x=\delta z\cdot y+\delta z\cdot(x-y)$. The term $\delta z\cdot y$ is independent of $x$ and can be absorbed into the constant on the right-hand side of the new cutting inequality. We next seek vectors $z$ satisfying $|z\cdot(x-y)|\le1$ for all $x,y\in C$. This condition bounds the variation of the added term by
\[
 |\delta z\cdot x-\delta z\cdot y|
 =\delta|z\cdot(x-y)|
 \le\delta.
\]
Consequently, for every $x\in C$,
\[
 (u+\delta z)\cdot x
 =u\cdot x+\delta z\cdot y+\delta z\cdot(x-y)
 \ge b+\delta z\cdot y-\delta.
\]
Thus, for every $y\in C$,
\begin{equation}
\label{rand:eq-cap-from-perturbation}
 C\subset Q\cap\{x:(u+\delta z)\cdot x
                    \ge b+\delta z\cdot y-\delta\}.
\end{equation}
The bound $|z\cdot(x-y)|\le1$ will also be used below, together with convexity and the ball contained in $C$, to control the volume of the enlarged cap. We will choose $\delta>0$ later in terms of $c$. Since the condition on $z$ involves the difference vectors in $C-C=\{x-y:x,y\in C\}$, we first enclose this difference set in an ellipsoid. This will allow us to construct a set of suitable perturbation vectors and estimate its $(m-1)$-dimensional volume.

We now begin the construction by obtaining an ellipsoidal bound for $C-C$. The cap $C$ is compact and convex, with nonempty interior because it contains a ball of radius $c$. John's ellipsoid theorem~\cite{John1948} therefore gives a point $z_0\in\mathbb R^m$ and a symmetric positive definite matrix $A$ such that
\[
 z_0+AB^m\subset C\subset z_0+mAB^m.
\]
Since $C$ contains a ball of radius $c$, its width in every unit direction $\omega$ is at least $2c$. The inclusion $C\subset z_0+mAB^m$ bounds this width above by the width of the ellipsoid in the same direction, namely $2m|A\omega|$, and hence $2c\le2m|A\omega|$. The inclusion $z_0+AB^m\subset C$ gives $\det A\,|B^m|\le|C|\le S$, so $\det A\le S/|B^m|$. Moreover, $C\subset z_0+mAB^m$ gives $C-C\subset2mAB^m$. Setting $M=2mA$, we obtain
\begin{equation}
\label{rand:eq-cap-polar}
 C-C\subset MB^m,\qquad
 \lambda_{\min}(M)\ge2c,\qquad
 \det M\le C_mS,
\end{equation}
where $\lambda_{\min}(M)$ denotes the smallest eigenvalue of $M$ and $C_m$ depends only on $m$.

Let $u^\perp=\{x\in\mathbb R^m:u\cdot x=0\}$ denote the hyperplane through the origin orthogonal to $u$. We choose $z\in u^\perp$ to describe changes of the normal in directions perpendicular to $u$, and will then normalize $u+\delta z$. To ensure that $|z\cdot(x-y)|\le1$ for all $x,y\in C$, we also require $z\in M^{-1}B^m$. For such a vector $z$ and any $w\in C-C$, we have $|Mz|\le1$ and $|M^{-1}w|\le1$. Symmetry of $M$ and the Cauchy--Schwarz inequality then give $|z\cdot w|\le1$. We therefore use the section
\[
 Z=M^{-1}B^m\cap u^\perp.
\]
Since $\lambda_{\min}(M)\ge2c$, we have $2c|z|\le|Mz|\le1$, and hence $|z|\le(2c)^{-1}$.  Thus $Z\subset(2c)^{-1}B^m$, and every $z\in Z$ satisfies $|z\cdot(x-y)|\le1$ for all $x,y\in C$.

We now use the vectors in $Z$ to construct directions for caps containing $C$. Set $\delta=c/4$ and define
\[
 G_C=\left\{\frac{u+\delta z}{|u+\delta z|}:z\in Z\right\}
 \subset S^{m-1}.
\]
Here $S^{m-1}=\{\theta\in\mathbb R^m:|\theta|=1\}$ denotes the unit sphere in $\mathbb R^m$. We first prove that every direction in $G_C$ admits a containing cap of the required volume. We will then estimate the spherical measure of $G_C$.

\medskip
\noindent\textit{Step 2. Bounding the volume of the enlarged cap.}
Choose $q\in Q$ such that $u\cdot q=\max_{x\in Q}u\cdot x$. Since $C$ is nonempty, $u\cdot q\ge b$, so $q\in C$. We will dilate $C$ about $q$ to bound the volume of the cap constructed in the perturbed direction. Put
\[
 H=u\cdot q-b.
\]
The cap $C$ lies between the planes $u\cdot x=b$ and $u\cdot x=u\cdot q$, whose distance is $H$. Since $C$ contains a ball of radius $c$, we have $H\ge2c$.

For $z\in Z$, put $v=u+\delta z$, so that $v/|v|\in G_C$. Applying \eqref{rand:eq-cap-from-perturbation} with $y=q$ and using $b+\delta z\cdot q=v\cdot q-H$, we obtain
\[
 C\subset D_{C,v}
 :=Q\cap\{x:v\cdot(x-q)\ge-H-\delta\}.
\]
The choice $\delta=c/4$ gives $\delta/H\le1/8$ and $\delta|z|\le1/8$ for $z\in Z$, which we will use, respectively, to bound $|D_{C,v}|$ by $2^m|C|$ and to obtain a lower bound for the normalized spherical measure of $G_C$.

To bound the volume of $D_{C,v}$, we will show that it is contained in $q+2(C-q)$, the dilation of $C$ by a factor of two about $q$. For a point $y\in D_{C,v}\setminus C$, we compare its distance from $q$ with the distance to the point where the segment from $q$ to $y$ meets the cutting plane of $C$.

For $y\in D_{C,v}\setminus C$, let $w$ be the intersection of the segment joining $q$ to $y$ with the cutting plane $\{x:u\cdot x=b\}$. Since $u\cdot q>b>u\cdot y$, this point lies strictly between $q$ and $y$. The ratio of the segment lengths is
\[
 a=\frac{|y-q|}{|w-q|}
  =\frac{u\cdot q-u\cdot y}{u\cdot q-u\cdot w}
  =\frac{u\cdot q-u\cdot y}{H}>1,
\]
and hence $w=q+\frac{y-q}{a}$. Since $y-q=a(w-q)$, proving $a\le2$ will place $y$ in $q+2(C-q)$. Since $u\cdot(w-q)=b-u\cdot q=-H$, the bound $|z\cdot(w-q)|\le1$ gives $v\cdot(w-q)=-H+\delta z\cdot(w-q)\le-(H-\delta)$. Therefore,
\[
 -H-\delta
 \le v\cdot(y-q)
 =a\,v\cdot(w-q)
 \le-a(H-\delta).
\]
Since $H\ge2c$ and $\delta=c/4$, we have $H-\delta>0$, and this gives
\[
 a\le\frac{H+\delta}{H-\delta}\le\frac97<2.
\]
The set $C-q$ is convex and contains zero. Thus $y-q=a(w-q)\in2(C-q)$ for every $y\in D_{C,v}\setminus C$. Points already in $C$ also belong to $q+2(C-q)$. Consequently,
\begin{equation}
\label{rand:eq-perturbed-cap}
 C\subset D_{C,v}\subset q+2(C-q),
 \qquad
 |D_{C,v}|\le2^m|C|\le2^mS.
\end{equation}

\medskip
\noindent\textit{Step 3. A lower bound for the measure of the direction sets.} 
Since $M$ is symmetric, $(Mx)\cdot(M^{-1}u)=x\cdot u$ for every $x\in\mathbb R^m$. Together with invertibility, this gives $M(u^\perp)=(M^{-1}u)^\perp$ and
\[
 MZ=B^m\cap(M^{-1}u)^\perp.
\]
Thus $MZ$ is the $(m-1)$-dimensional unit ball in $(M^{-1}u)^\perp$.

To compare the $(m-1)$-dimensional volumes of $Z$ and $MZ$, choose an orthonormal basis $\omega_1,\ldots,\omega_{m-1}$ of $u^\perp$. Together with $u$, these vectors form an orthonormal basis of $\mathbb R^m$, so the parallelepiped spanned by $M\omega_1,\ldots,M\omega_{m-1},Mu$ has volume $\det M$. Its base is spanned by $M\omega_1,\ldots,M\omega_{m-1}$ and lies in the hyperplane $(M^{-1}u)^\perp$, whose unit normal can be chosen as $M^{-1}u/|M^{-1}u|$. The height above this base is therefore
\[
 (Mu)\cdot\frac{M^{-1}u}{|M^{-1}u|}
 =\frac{1}{|M^{-1}u|}.
\]
Dividing the volume by this height gives the base volume $\det M\,|M^{-1}u|$. Since the parallelepiped spanned by $\omega_1,\ldots,\omega_{m-1}$ has unit $(m-1)$-dimensional volume, the linear change of variables on $u^\perp$ gives
\[
 |MZ|_{m-1}
 =\det M\,|M^{-1}u|\,|Z|_{m-1}.
\]
Hence
\begin{equation}
\label{rand:eq-normal-section}
 |Z|_{m-1}
 =\frac{|B^{m-1}|}{\det M\,|M^{-1}u|}
 \ge\frac{c_{m,c}}{S},
\end{equation}
where $|Z|_{m-1}$ denotes volume in $u^\perp$. The inequality follows from $\det M\le C_mS$ in \eqref{rand:eq-cap-polar} and
\[
 |M^{-1}u|
 \le\lambda_{\min}(M)^{-1}|u|
 \le(2c)^{-1}.
\]

To convert the volume lower bound \eqref{rand:eq-normal-section} for $Z$ into a surface area lower bound for $G_C$, define $F(z)=(u+\delta z)/|u+\delta z|$ for all $z\in u^\perp$. Since $|u+\delta z|^2=1+\delta^2|z|^2\ge1$, this map is smooth on $u^\perp$. Since $u\cdot z=0$ and $|u|=1$, we have
\[
 F(z)\cdot u=\frac1{|u+\delta z|}>0,
 \qquad
 z=\delta^{-1}\left(\frac{F(z)}{F(z)\cdot u}-u\right).
\]
Thus $F(z)$ uniquely determines $z$, so $F$ is injective.

The surface Jacobian $J_F(z)$ equals the $(m-1)$-dimensional volume of the parallelepiped spanned by the vectors obtained by applying the derivative $DF(z)$ of $F$ at $z$ to an orthonormal basis of $u^\perp$. Set $r(z)=\sqrt{1+\delta^2|z|^2}$. For a fixed $z\ne0$, write $r=r(z)$ and choose a new orthonormal basis of $u^\perp$ with its first vector parallel to $z$. Write this basis as $\omega_1=z/|z|,\omega_2,\ldots,\omega_{m-1}$. Since $Dr(z)[\omega_j]=\delta^2(z\cdot\omega_j)/r$, differentiating $F(z)=(u+\delta z)/r$ gives
\[
 DF(z)[\omega_j]
 =\lim_{\epsilon\to0}\frac{F(z+\epsilon\omega_j)-F(z)}{\epsilon}
 =\frac{\delta}{r}\omega_j
 -\frac{\delta^2(z\cdot\omega_j)}{r^3}(u+\delta z).
\]
Using $z=|z|\omega_1$ and $r^2=1+\delta^2|z|^2$, we obtain
\[
 DF(z)[\omega_1]
 =\frac{\delta}{r^3}(\omega_1-\delta|z|u),
 \qquad
 DF(z)[\omega_j]
 =\frac{\delta}{r}\omega_j
 \quad(2\le j\le m-1).
\]
For $j\ge2$, the vector $\omega_j$ is orthogonal to both $\omega_1$ and $u$, so $\left\{DF(z)[\omega_j]\right\}_{j=1,\cdots m-1}$ are mutually orthogonal. Their lengths satisfy
\[
 |DF(z)[\omega_1]|
 =\frac{\delta}{r^3}\sqrt{1+\delta^2|z|^2}
 =\frac{\delta}{r^2},
 \qquad
 |DF(z)[\omega_j]|=\frac{\delta}{r}
 \quad(2\le j\le m-1).
\]
Since the vectors $DF(z)[\omega_j]$ are mutually orthogonal, multiplying their lengths gives
\[
 J_F(z)
 =\frac{\delta}{r^2}\left(\frac{\delta}{r}\right)^{m-2}
 =\delta^{m-1}(1+\delta^2|z|^2)^{-m/2}.
\]
The formula also holds at $z=0$ by continuity. Directly, $DF(0)[h]=\delta h$ for $h\in u^\perp$, so $J_F(0)=\delta^{m-1}$. Since $F$ is smooth and injective and $J_F(z)>0$, the area formula \cite[Chapter~2, \S3, (3.2)]{simon2014introduction} gives the surface area of $G_C$ as $\int_Z J_F(z)\,dz$, where $dz$ denotes $(m-1)$-dimensional volume in $u^\perp$.

The earlier bound $|z|\le(2c)^{-1}$ and the choice $\delta=c/4$ give $\delta|z|\le1/8$. Hence $J_F(z)$ is bounded below on $Z$ by a positive constant depending only on $m,c$. Let $\sigma$ denote normalized surface measure on $S^{m-1}$, so that $\sigma(S^{m-1})=1$. Combining the area formula with \eqref{rand:eq-normal-section} gives
\begin{equation}
\label{rand:eq-normal-patch}
 \sigma(G_C)\ge a_{m,c}/S.
\end{equation}
We will use $a_{m,c}/S$ as the parameter $\epsilon$ in Lemma~\ref{rand:epsilon-net}, which requires this parameter to be at most $1/2$. Since $S\ge|cB^m|$, we may decrease $a_{m,c}>0$ so that $a_{m,c}/S\le1/2$ for every admissible $S$.

\medskip
\noindent\textit{Step 4. Selecting a common set of directions.}
We apply Lemma~\ref{rand:epsilon-net} to the finite family
$\{G_C:C\in\mathcal A\}$ on $S^{m-1}$, equipped with its
Borel $\sigma$-algebra and normalized surface measure $\sigma$.
Each $G_C$ is the continuous image of the compact set $Z$,
so it is compact and hence Borel measurable.
To verify the hypothesis in Lemma~\ref{rand:epsilon-net},
fix any integer $n\ge1$ and any list
$\theta_1,\ldots,\theta_n\in S^{m-1}$, allowing repeated directions.
We must show that the index sets
\[
 \{i\in\{1,\ldots,n\}:\theta_i\in G_C\},
 \qquad C\in\mathcal A,
\]
take at most $A(n+1)^D$ distinct values as $C$ varies,
where $A$ and $D$ are the parameters in
Lemma~\ref{rand:epsilon-net} and can be chosen to depend
only on $m$.

Fix $C\in\mathcal A$ and use its associated $u$, $M$, and $Z$. Let $G$ be the symmetric matrix representing the quadratic form $z\mapsto|Mz|^2$ on $u^\perp$, extended by zero in the direction $u$. Thus $Gu=0$ and $z^{\mathsf T}Gz=|Mz|^2$ for $z\in u^\perp$. Since $M$ is invertible, $G$ is positive definite on $u^\perp$, and
\[
 Z=\{z\in u^\perp:|Mz|\le1\}
  =\{z\in u^\perp:z^{\mathsf T}Gz\le1\}.
\]
For a unit vector $\theta$, membership in $G_C$ is equivalent to
\begin{equation}
\label{rand:eq-patch-conditions}
 \theta\cdot u>0,
 \qquad
 \theta^{\mathsf T}G\theta
 \le\delta^2(\theta\cdot u)^2.
\end{equation}
Every $\theta\in G_C$ satisfies $\theta\cdot u>0$ because $Z\subset u^\perp$. When $\theta\cdot u>0$, the only possible preimage of $\theta$ under the parametrization of $G_C$ is
\[
 z=\frac{\theta-(\theta\cdot u)u}{\delta(\theta\cdot u)}.
\]
This vector lies in $u^\perp$ and satisfies $u+\delta z=\theta/(\theta\cdot u)$, so normalizing $u+\delta z$ gives $\theta$. Symmetry of $G$ and the identity $Gu=0$ yield
\[
 z^{\mathsf T}Gz
 =\frac{\theta^{\mathsf T}G\theta}
        {\delta^2(\theta\cdot u)^2}.
\]
Hence $z\in Z$ exactly when the second inequality holds, proving the equivalence.

We introduce new coordinates to express the two inequalities in \eqref{rand:eq-patch-conditions} as linear inequalities. For $x=(x_1,\ldots,x_m)\in\mathbb R^m$, define
\[
 \Phi(x)=\bigl((x_r)_{1\le r\le m},
              (x_rx_s)_{1\le r\le s\le m}\bigr)
 \in\mathbb R^{M_m},
 \qquad
 M_m=m+\frac{m(m+1)}2.
\]
Thus $\Phi(x)$ records the original coordinates and all products $x_rx_s$ with $r\le s$. Both $\theta\cdot u$ and $\theta^{\mathsf T}G\theta-\delta^2(\theta\cdot u)^2$ are linear combinations of the coordinates of $\Phi(\theta)$. Therefore, the first inequality in \eqref{rand:eq-patch-conditions} requires $\Phi(\theta)$ to lie in an open halfspace of $\mathbb R^{M_m}$, and the second requires it to lie in a closed halfspace.

Apply Lemma~\ref{rand:halfspace-count} with $d=M_m$ to the fixed points $\Phi(\theta_1),\ldots,\Phi(\theta_n)$. Each of the two halfspace conditions defining $G_C$ gives at most $C_{M_m}(n+1)^{M_m+1}$ different index sets as $C$ varies. The set $\{i:\theta_i\in G_C\}$ is the intersection of the index sets from the two conditions. Counting the possible pairs therefore bounds the number of distinct intersections by
\[
 C_{M_m}^2(n+1)^{2(M_m+1)}.
\]
This verifies the hypothesis of Lemma~\ref{rand:epsilon-net}
with its parameters $A=C_{M_m}^2$ and $D=2(M_m+1)$,
both depending only on $m$.

Applying Lemma~\ref{rand:epsilon-net} to the family $\{G_C:C\in\mathcal A\}$ on $S^{m-1}$, with normalized surface measure and $\epsilon=a_{m,c}/S$, gives a finite set of unit directions such that every $G_C$ contains at least one of them. Discard any direction that belongs to no $G_C$, and denote the remaining directions by $\theta_1,\ldots,\theta_N$. They meet every $G_C$, and
\[
 N
 \le C_m\frac{S}{a_{m,c}}
          \log\!\left(\frac{2S}{a_{m,c}}\right)
 \le C_{m,c}S\log(1+S).
\]
Our choice of $a_{m,c}$ ensures that $0<\epsilon\le1/2$. The last inequality uses $S\ge|cB^m|$ and the fact that $a_{m,c}>0$ depends only on $m$ and $c$.

\medskip
\noindent\textit{Step 5. Constructing the covering caps.}
Fix $i\in\{1,\ldots,N\}$. For each $C\in\mathcal A$ with $\theta_i\in G_C$, use the associated $u$ and $Z$ to choose $z\in Z$ such that
\[
 \theta_i=\frac{u+\delta z}{|u+\delta z|},
 \qquad v=u+\delta z.
\]
By \eqref{rand:eq-perturbed-cap}, the cap $D_{C,v}$ contains $C$, has normal direction $\theta_i$, and satisfies $|D_{C,v}|\le2^mS$. For this fixed $i$, the chosen caps $D_{C,v}$ form a nonempty finite family. They all have the form $Q\cap\{x:\theta_i\cdot x\ge t\}$ and are therefore nested. Choose $F_i$ from this family such that it contains all the others and $|F_i|\le2^mS$. Moreover, $F_i$ contains every $C\in\mathcal A$ for which $\theta_i\in G_C$.

Every $C\in\mathcal A$ has $\theta_i\in G_C$ for some $i$, and hence $C\subset F_i$. Thus $F_1,\ldots,F_N$ give the required cover and satisfy \eqref{rand:eq-thick-cover-count}. Finally,
\[
 \left|\bigcup_{C\in\mathcal A}C\right|
 \le\sum_{i=1}^N|F_i|
 \le2^mSN
 \le C_{m,c}S^2\log(1+S),
\]
after enlarging $C_{m,c}$. This proves \eqref{rand:eq-thick-union}.
\end{proof}

\subsection{Overlap Bounds for Covering Caps}
\label{Volumes of Cap Unions and Pointwise Multiplicity}

We first prove a volume bound for unions of caps. We then use this bound to estimate how many covering caps at a fixed volume scale can contain a given point.

\begin{proposition}
\label{rand:visibility}
Let $d\ge2$ and $T\ge1$. Let $L\subset\mathbb R^d$ be a convex body containing the Euclidean unit ball $B^d$, and let $U$ be the union of finitely many caps of $L$, each containing the origin and having volume at most $T$. Then
\begin{equation}
\label{rand:eq-visibility}
 |U|\le C_dT^{2-2/d}\log(1+T),
\end{equation}
where $C_d$ depends only on $d$, independently of $L$ and the number of caps.
\end{proposition}

\begin{proof}
We first show that every point of $U$ lies within distance $C_dT$ of the origin. We then split $U$ into a part near the origin and a part farther away, and estimate their volumes separately.

\medskip
\noindent\textit{Step 1. Bounding the distance from the origin.} 
Fix a cap $E$ in the union and write
\[
 E=L\cap\{x:\nu\cdot x\ge t\},
 \qquad |\nu|=1.
\]
Since $0\in E$, we have $t\le0$. Thus $E$ contains the half unit ball
\[
 B^d\cap\{x:\nu\cdot x\ge0\},
\]
and hence the ball $\nu/2+\tfrac12B^d$.

For $x\in E$ with $x\ne\nu/2$, consider the $(d-1)$-dimensional disk of radius $1/2$ centered at $\nu/2$ in the hyperplane through $\nu/2$ perpendicular to $x-\nu/2$. By convexity, the cone with this disk as its base and $x$ as its vertex is contained in $E$. Its height is $|x-\nu/2|$, so the cone volume formula gives
\[
 \frac{|B^{d-1}|}{d\,2^{d-1}}\,|x-\nu/2|
 \le |E|
 \le T.
\]
Since $|\nu/2|=1/2$ and $T\ge1$, this gives $|x|\le C_dT$. The same bound holds for $x=\nu/2$. Consequently,
\begin{equation}
\label{rand:eq-visibility-radius}
 E\subset C_dT B^d.
\end{equation}
This estimate holds for every cap in the union and therefore also for $U$.

\medskip
\noindent\textit{Step 2. Reduction to section volume estimates.} 
The part of $U$ inside $[-2,2]^d$ has volume at most $4^d$. To estimate the part of $U$ outside $[-2,2]^d$, we group points according to which coordinate has maximal absolute value and its sign. This gives the $2d$ sectors
\[
 \left\{x\in\mathbb R^d:
 \pm x_j=\max_{1\le k\le d}|x_k|\right\},
 \qquad j=1,\ldots,d.
\]
These sectors cover $\mathbb R^d$, with overlaps where different coordinates have the same maximal absolute value. We bound the volume of $U\setminus[-2,2]^d$ by summing the volumes of its intersections with these sectors.

Fix one sector. Without loss of generality, assume that the chosen sector is
\[
 \{(y,z)\in\mathbb R^{d-1}\times[0,\infty):
   |y_j|\le z\ \text{for }1\le j\le d-1\}.
\]
Since $z\ge|y_j|$ for all $j$ in the chosen sector, points outside $[-2,2]^d$ have $z>2$. For $r\ge1$, define $V_r$ as the set of $y$-coordinates of points of $U$ in this sector at height $2r$:
\[
 V_r=\{y\in[-2r,2r]^{d-1}:(y,2r)\in U\}.
\]
By \eqref{rand:eq-visibility-radius}, the volume of the part of $U$ in the chosen sector outside $[-2,2]^d$ is $2\int_1^{C_dT/2} |V_r|_{d-1}\,dr$, where $|V_r|_{d-1}$ denotes $(d-1)$-dimensional volume. We will therefore bound $|V_r|_{d-1}$ and integrate the resulting estimates.

For this fixed $r$, consider only those original caps $E\subset U$ that contain a point $(y,2r)$ with $y\in[-2r,2r]^{d-1}$. If there are no such caps, then $V_r$ is empty and there is nothing to estimate at this height. Otherwise, define the sections of $L$ and the retained caps at height $r$, restricted to the box $[-r-1,r+1]^{d-1}$, by
\[
 Q_r=\{y:(y,r)\in L\}\cap[-r-1,r+1]^{d-1},
 \qquad
 C_E=\{y:(y,r)\in E\}\cap Q_r.
\]

Fix a retained cap $E$ and choose $(y_*,2r)\in E$ with $y_*\in[-2r,2r]^{d-1}$. Since $B^d\subset L$ and the cutting halfspace of $E$ contains the origin, the section of $E$ at height zero contains a $(d-1)$-dimensional half-ball of radius one. Hence there is $b\in\mathbb R^{d-1}$ with $|b|\le\tfrac12$ such that
\[
 b+\tfrac12B^{d-1}\subset\{y:(y,0)\in E\}.
\]
Taking midpoints between $(y_*,2r)$ and the points of this ball at height zero gives
\[
 \frac{y_*+b}{2}+\tfrac14B^{d-1}
 \subset \{y:(y,r)\in E\}.
\]
Since $|(y_*)_j|\le2r$ and $|b_j|\le1/2$, every point of this midpoint ball has each coordinate bounded in absolute value by $r+1/2$. Thus this ball is contained in $C_E$.

Therefore every $C_E$ contains a $(d-1)$-dimensional ball of radius $1/4$. Since $Q_r$ is compact and convex and contains such a ball, it is a convex body in $\mathbb R^{d-1}$. Write $E=L\cap\{x:\nu\cdot x\ge t\}$. At height $z=r$, we obtain
\[
 C_E=Q_r\cap\left\{y\in\mathbb R^{d-1}:
 \sum_{j=1}^{d-1}\nu_jy_j\ge t-r\nu_d\right\}.
\]
Thus $C_E$ is a cap of $Q_r$, possibly equal to $Q_r$. Moreover, the definition of $Q_r$ and $r\ge1$ give
\[
 |Q_r|_{d-1}\le(2r+2)^{d-1}\le C_dr^{d-1}.
\]
Since $0\in E$, the cone with vertex zero and base $C_E\times\{r\}$ lies in $E$, whose volume is $r|C_E|_{d-1}/d$, so
\begin{equation}
\label{rand:eq-slice-cap-volume}
 |C_E|_{d-1}
 \le \frac{d|E|}{r}
 \le \frac{dT}{r}.
\end{equation}
The ball of radius $1/4$ contained in each $C_E$ also gives $dT/r\ge|(1/4)B^{d-1}|_{d-1}$.

Finally, if $y\in V_r$, then $(y,2r)$ belongs to a retained cap $E$. Convexity and $0\in E$ give $(y/2,r)\in E$, and $y/2\in[-r,r]^{d-1}\subset[-r-1,r+1]^{d-1}$. Therefore,
\[
 \tfrac12V_r\subset\bigcup_E C_E,
 \qquad
 |V_r|_{d-1}\le2^{d-1}\left|\bigcup_E C_E\right|_{d-1}.
\]
It remains to estimate $\left|\bigcup_E C_E\right|_{d-1}$, where $E$ ranges over the original caps containing a point $(y,2r)$ with $y\in[-2r,2r]^{d-1}$.

\medskip
\noindent\textit{Step 3. Bounding $\left|\bigcup_E C_E\right|_{d-1}$.} 
We first use Lemma~\ref{rand:wet-bound} in dimension $d-1$. By \eqref{rand:eq-slice-cap-volume}, if $dT/r\le|Q_r|_{d-1}$, then $\bigcup_E C_E$ is contained in the wet part of $Q_r$ with relative volume parameter $dT/(r|Q_r|_{d-1})$. Therefore, 
\[
 \left|\bigcup_E C_E\right|_{d-1}
 \le|Q_r|_{d-1} W_{Q_r}\!\left(\frac{dT}{r|Q_r|_{d-1}}\right)
 \le C_d|Q_r|_{d-1}^{1-2/d}
       \left(\frac{dT}{r}\right)^{2/d}.
\]
If $dT/r>|Q_r|_{d-1}$, the same upper bound follows from
\[
 \left|\bigcup_E C_E\right|_{d-1}
 \le |Q_r|_{d-1}
 \le |Q_r|_{d-1}^{1-2/d}
       \left(\frac{dT}{r}\right)^{2/d}.
\]
Using $|Q_r|_{d-1}\le C_dr^{d-1}$, we therefore obtain
\begin{equation}
\label{rand:eq-slice-wet}
 \left|\bigcup_E C_E\right|_{d-1}
 \le C_d|Q_r|_{d-1}^{1-2/d}
       \left(\frac{dT}{r}\right)^{2/d}
 \le C_dT^{2/d}r^{d-3}.
\end{equation}

For $d\ge3$, we apply Lemma~\ref{rand:cap-cover} to the finite family $\{C_E\}$ with $m=d-1\ge2$, $Q=Q_r$, $c=1/4$, and $S=dT/r$. As shown in Step~2, $Q_r$ is a convex body, and each $C_E$ is a cap of $Q_r$ containing a ball of radius $1/4$ and having volume at most $dT/r$. Thus \eqref{rand:eq-thick-union} gives
\begin{equation}
\label{rand:eq-slice-cover}
 \left|\bigcup_E C_E\right|_{d-1}
 \le C_d\left(\frac{dT}{r}\right)^2
       \log\!\left(1+\frac{dT}{r}\right)
 \le C_d(T/r)^2\log(1+T/r).
\end{equation}
The last inequality uses $\log(1+ds)\le d\log(1+s)$ with $s=T/r$ and absorbs the factors depending only on $d$ into $C_d$.

Thus \eqref{rand:eq-slice-wet} applies in every dimension $d\ge2$, while \eqref{rand:eq-slice-cover} provides an additional bound when $d\ge3$. 

\medskip
\noindent\textit{Step 4. Bounding the volume of $U$.}
By \eqref{rand:eq-visibility-radius}, $V_r=\varnothing$ whenever $r>C_dT$. When $d=2$, \eqref{rand:eq-slice-wet} gives the bound $CT/r$. For each of the four coordinate sectors, the area of the part of $U$ in that sector outside $[-2,2]^2$ is $2\int_1^{CT}|V_r|_1\,dr$. Using $|V_r|_1\le CT/r$, summing over the four sectors, and adding the area inside $[-2,2]^2$, we obtain
\[
 \begin{aligned}
 |U|
 &\le16+8CT\int_1^{CT}\frac{dr}{r}\\
 &=16+8CT\log(CT)\\
 &\le CT\log(1+T).
 \end{aligned}
\]
This is \eqref{rand:eq-visibility} in dimension two.

Now suppose $d\ge3$.
We use \eqref{rand:eq-slice-wet} for $1\le r\le T^{2/d}$ and \eqref{rand:eq-slice-cover} for $r\ge T^{2/d}$. The choice of $T^{2/d}$ balances the powers of $T$ in the two integrated estimates. Moreover, $\log(1+T/r)\le\log(1+T)$ for $r\ge1$. We have
\[
 \begin{aligned}
 |U|
 &\le C_d\left(
 1+T^{2/d}\int_1^{T^{2/d}} r^{d-3}\,dr
   +T^2\log(1+T)\int_{T^{2/d}}^\infty r^{-2}\,dr
 \right)\\
 &\le C_d\left(
 T^{2/d}\bigl(T^{2/d}\bigr)^{d-2}
 +\frac{T^2\log(1+T)}{T^{2/d}}
 \right)\\
 &\le C_dT^{2-2/d}\log(1+T).
 \end{aligned}
\]

This completes the proof.
\end{proof}

We next bound the number of covering caps at a fixed volume scale that contain a given point. Let $K\subset\mathbb R^d$ be a convex body with $d\ge2$, and let $\mu_K$ denote its normalized volume measure. Choose $\tau_0>0$ sufficiently small, depending only on $d$, so that Lemma~\ref{hil:economic-cover} applies whenever $0<\tau\le\tau_0$. Set
\[
 \tau_j=2^{-j}\tau_0,\qquad j\ge0.
\]
For each $j\ge0$, Lemma~\ref{hil:economic-cover} provides $N_j$ covering caps $E_{j,\ell}$ and pairwise disjoint convex sets $D_{j,\ell}\subset E_{j,\ell}$ satisfying
\[
 \begin{gathered}
 c_d\tau_j
 \le\mu_K(D_{j,\ell})
 \le\mu_K(E_{j,\ell})
 \le C_d\tau_j,
 \quad\forall1\le\ell\le N_j,\\
 c_d\mathcal C_K(\tau_j)
 \le N_j
 \le C_d\mathcal C_K(\tau_j),
 \end{gathered}
\]
where $c_d,C_d>0$ depend only on $d$. Every cap $C$ of $K$ with $\mu_K(C)\le\tau_j$ is contained in at least one $E_{j,\ell}$. The sets $D_{j,\ell}$ are pairwise disjoint for each fixed $j$, but sets corresponding to different values of $j$ may overlap.

\begin{proposition}
\label{rand:multiplicity}
Fix $j\ge0$, and consider the covering caps $E_{j,1},\ldots,E_{j,N_j}$ constructed above at scale $\tau_j$. Choose any index set $I_j\subset\{1,\ldots,N_j\}$, and define the number of selected caps containing $x$ by
\[
 m_j(x)=\sum_{\ell\in I_j}\mathbf1_{E_{j,\ell}}(x).
\]
Recall that $v_K$ denotes the floating depth defined in Definition~\ref{def:wet-part-depth}. For $\mu_K$-almost every $x\in K$ with $m_j(x)>0$, set $T=\tau_j/v_K(x)$. Then
\begin{equation}
\label{rand:eq-multiplicity}
 m_j(x)\le C_dT^{(d-2)/d}\log(1+T),
\end{equation}
where $C_d$ depends only on $d$ and is independent of $K$, $j$, and the choice of $I_j$.
\end{proposition}
\begin{proof}
Fix an interior point $x$ of $K$ with $m_j(x)>0$, and choose $\ell\in I_j$ such that $x\in E_{j,\ell}$. The definition of floating depth and the volume bound for $E_{j,\ell}$ give
\[
 0<v:=v_K(x)\le\mu_K(E_{j,\ell})\le C_d\tau_j,
 \qquad
 T=\frac{\tau_j}{v}\ge\frac1{C_d}.
\]
We use the \emph{Macbeath region}~\cite[p.~278]{BaranyLarman1988}
\[
 M_K(x)
 =x+\big((K-x)\cap(x-K)\big)
 =K\cap(2x-K),
 \qquad
 u=\mu_K(M_K(x)).
\]
It is a convex body because $x$ lies in the interior of $K$. We also require $\tau_0$ to satisfy
\[
 C_d\tau_0\le(2d)^{-2d}.
\]
Since $\tau_j\le\tau_0$, this ensures that $v\le(2d)^{-2d}$. Lemmas~1 and~2 of \cite[p.~279]{BaranyLarman1988}, applied after rescaling $K$ to volume one, therefore give $u\le2v$ and $v\le(3d)^d u$.

Since $M_K(x)$ is centrally symmetric about $x$, John's ellipsoid theorem~\cite{John1948} gives a symmetric positive definite matrix $P$ such that
\[
 x+PB^d\subset M_K(x)\subset x+\sqrt d\,PB^d.
\]
Define the invertible affine map
\[
 A(y)=P^{-1}(y-x),
 \qquad y\in\mathbb R^d.
\]
Then $A(x)=0$ and
\[
 B^d\subset A(M_K(x))\subset\sqrt d\,B^d.
\]
In particular, $B^d\subset A(K)$. Since $A$ multiplies the $d$-dimensional volume of every measurable set by the same positive constant, we have
\begin{equation}
\label{rand:eq-affine-volume}
 |A(F)|
 =\frac{|A(M_K(x))|}{u}\,\mu_K(F)
 \asymp_d\frac{\mu_K(F)}{v},
 \qquad F\subset K\text{ measurable}.
\end{equation}
Here we used the previously established inequalities $u\le2v$ and $v\le(3d)^du$.

Now consider the indices $\ell\in I_j$ for which $x\in E_{j,\ell}$. There are exactly $m_j(x)$ such indices. For each such index $\ell$, $A(E_{j,\ell})$ is a cap of $A(K)$ containing zero. Applying \eqref{rand:eq-affine-volume} to $F=E_{j,\ell}$ and $F=D_{j,\ell}$, and using $\mu_K(E_{j,\ell})\le C_d\tau_j$ and $\mu_K(D_{j,\ell})\ge c_d\tau_j$, we obtain
\[
 |A(E_{j,\ell})|
 \le C_d\frac{\tau_j}{v}=C_dT,
 \qquad
 |A(D_{j,\ell})|
 \ge c_d\frac{\tau_j}{v}=c_dT.
\]
Moreover, the sets $A(D_{j,\ell})$ remain pairwise disjoint because $A$ is injective, and each lies in the corresponding cap $A(E_{j,\ell})$.

Since $T\ge1/C_d$, we may enlarge the constant in the bound $|A(E_{j,\ell})|\le C_dT$ so that $C_dT\ge1$. We apply Proposition~\ref{rand:visibility} to the convex body $A(K)$ and the caps $A(E_{j,\ell})$ with $\ell\in I_j$ and $x\in E_{j,\ell}$, using $C_dT$ as the volume parameter. Therefore,
\[
 c_dT\,m_j(x)
 \le
 \left|
 \bigcup_{\substack{\ell\in I_j\\x\in E_{j,\ell}}}
 A(E_{j,\ell})
 \right|
 \le C_dT^{2-2/d}\log(1+T),
\]
which yields \eqref{rand:eq-multiplicity}. Since the boundary of $K$ has zero volume, this proves the asserted estimate for $\mu_K$-almost every $x$ with $m_j(x)>0$. 
\end{proof}

\subsection{\texorpdfstring{$L^s$}{L\textasciicircum s} Bounds for Sums of Cap Indicators}

We bound the $L^s(\mu_K)$ norm of linear combinations of normalized indicator functions of the covering caps by the $\ell^s$ norm of their coefficients. 

Recall from Subsection~\ref{Volumes of Cap Unions and Pointwise Multiplicity} that, at each scale $\tau_j=2^{-j}\tau_0$ with $j\ge0$, we constructed covering caps $E_{j,1},\ldots,E_{j,N_j}$. By Lemma~\ref{hil:economic-cover}, every cap of $K$ with relative volume at most $\tau_j$ is contained in some $E_{j,\ell}$, and $\mu_K(E_{j,\ell})\le C_d\tau_j$.

Given a finite sum $F=\sum_i c_i\phi_{z_i}$, for each $j\ge0$ assign every index $i$ satisfying $\tau_j/2<\theta_{z_i}\le\tau_j$ to exactly one covering cap $E_{j,\ell}$ containing $C_{z_i}$. For each $E_{j,\ell}$ that receives at least one index, let $B$ be the set of indices assigned to it. Each nonempty group $B$ consists of the indices assigned to one covering cap $E_{j,\ell}$, so there are at most $N_j$ such groups at scale $\tau_j$. For every index $i$ in a group assigned to $E_{j,\ell}$, we have
\[
 C_{z_i}\subset E_{j,\ell},\qquad
 \theta_{z_i}\le\mu_K(E_{j,\ell})
 \le C_d\tau_j<2C_d\theta_{z_i}.
\]
The indices with $\theta_{z_i}>\tau_0$, if any, form one additional group, for which we use $K$ as the covering set. For every index in this group,
\[
 C_{z_i}\subset K,\qquad
 \theta_{z_i}\le\mu_K(K)=1
 \le\tau_0^{-1}\theta_{z_i}.
\]
Here $\tau_0^{-1}$ depends only on $d$. Since $0<\theta_{z_i}\le1$ and the intervals $(\tau_j/2,\tau_j]$, $j\ge0$, partition $(0,\tau_0]$, the resulting nonempty groups form a finite partition of the indices of $F$.

Put $h_d=(d-2)/d$, so that \eqref{rand:eq-multiplicity} reads
\[
 m_j(x)\le C_dT^{h_d}\log(1+T).
\]
The next theorem bounds the $L^s(\mu_K)$ norm of linear combinations of normalized indicator functions of the covering sets for $1\le s<r_d$. For $d\ge3$, define
\begin{equation}
\label{rand:eq-geometry-parameters}
 r_d=\frac{Q_d}{2}
     =1+\frac{\beta_d}{h_d}
     =1+\frac{2d}{(d+1)(d-2)},
\end{equation}
where $\beta_d=2/(d+1)$ is the exponent in the wet part volume bound in Lemma~\ref{rand:wet-bound}. For $d=2$, we have $h_2=0$ and set $r_2=\infty$.

\begin{theorem}
\label{rand:positive-synthesis}
For each group $B$ constructed above, let $E_B$ be its assigned covering set. Thus $E_B=E_{j,\ell}$ for a group assigned to a covering cap at scale $\tau_j$, and $E_B=K$ for the group of indices $i$ with $\theta_{z_i}>\tau_0$, if this group is nonempty. Set $V_B=\mu_K(E_B)>0$. For every $1\le s<r_d$ and every choice of coefficients $(a_B)$ indexed by the groups $B$, with only finitely many nonzero entries,
\begin{equation}
\label{rand:eq-positive-synthesis}
 \left\|\sum_B a_BV_B^{-1/s}\mathbf1_{E_B}\right\|_{L^s(\mu_K)}
 \le C_{d,s}\left(\sum_B|a_B|^s\right)^{1/s}.
\end{equation}
The constant $C_{d,s}$ depends only on $d$ and $s$, uniformly over $K$, the network representation, and the number of groups.
\end{theorem}

\begin{proof}
For $s=1$, the triangle inequality gives
\[
 \left\|\sum_B a_BV_B^{-1}\mathbf1_{E_B}\right\|_{L^1(\mu_K)}
 \le\sum_B|a_B|V_B^{-1}\mu_K(E_B)
 =\sum_B|a_B|.
\]

Now let $1<s<r_d$. The definition of $r_d$ gives $h_d(s-1)<\beta_d$, so we can choose $\gamma$, depending only on $d$ and $s$, such that
\[
 h_d<\gamma<\frac{\beta_d}{s-1}.
\]
By Lemma~\ref{geom:depth-synthesis}, it remains to bound
\[
 \mathcal H_\gamma(x)
 =\sum_B\left(\frac{v_K(x)}{V_B}\right)^\gamma
              \mathbf1_{E_B}(x)
\]
by a constant depending only on $d$ and $\gamma$, for $\mu_K$-almost every $x$. We first estimate the contribution from each volume scale $\tau_j$.

For each $j\ge0$, let $I_j$ consist of the indices $\ell$ for which a nonempty group is assigned to $E_{j,\ell}$, and set $m_j(x)=\sum_{\ell\in I_j}\mathbf1_{E_{j,\ell}}(x)$. Since the scales are countable, Proposition~\ref{rand:multiplicity} gives \eqref{rand:eq-multiplicity} simultaneously for all $j$ at $\mu_K$-almost every $x$. Fix such an $x$ with $v_K(x)>0$.

The construction gives $c_d\tau_j\le\mu_K(E_{j,\ell})\le C_d\tau_j$. Thus the contribution from scale $\tau_j$ to $\mathcal H_\gamma(x)$ satisfies
\[
 \begin{aligned}
 &\sum_{\ell\in I_j}
 \left(\frac{v_K(x)}{\mu_K(E_{j,\ell})}\right)^\gamma
 \mathbf1_{E_{j,\ell}}(x)\\
 &\qquad\le
 \left(\frac{v_K(x)}{c_d\tau_j}\right)^\gamma m_j(x)\\
 &\qquad\le C_{d,\gamma}
 \left(\frac{\tau_j}{v_K(x)}\right)^{h_d-\gamma}
 \log\!\left(1+\frac{\tau_j}{v_K(x)}\right).
 \end{aligned}
\]

If a group at scale $\tau_j$ contributes to $\mathcal H_\gamma(x)$, then $v_K(x)\le V_B\le C_d\tau_j$. Hence there is a unique integer $k\ge0$ such that
\[
 \frac{2^k}{C_d}
 \le\frac{\tau_j}{v_K(x)}
 <\frac{2^{k+1}}{C_d}.
\]
Since $\tau_{j+1}=\tau_j/2$, each $k$ corresponds to at most one scale. The preceding estimate for that scale is at most $C_{d,\gamma}\,2^{-k(\gamma-h_d)}(k+1)$, because $\log(1+\tau_j/v_K(x))\le C_d(k+1)$. The group of indices $i$ with $\theta_{z_i}>\tau_0$, if nonempty, has $E_B=K$ and $V_B=1$, so its contribution to $\mathcal H_\gamma(x)$ is $v_K(x)^\gamma\le1$. Using the bounds above for each scale $\tau_j$, we obtain
\begin{equation}
\label{rand:eq-weighted-multiplicity}
 \mathcal H_\gamma(x)
 \le 1+C_{d,\gamma}
       \sum_{k=0}^{\infty}2^{-k(\gamma-h_d)}(k+1)
 \le C_{d,\gamma}.
\end{equation}
The series converges because $\gamma>h_d$.

By \eqref{rand:eq-weighted-multiplicity}, applying Lemma~\ref{geom:depth-synthesis} with $a=\gamma(s-1)<\beta_d$ gives \eqref{rand:eq-positive-synthesis}. Since $\gamma$ depends only on $d$ and $s$, the constant in this estimate also depends only on $d$ and $s$.
\end{proof}

As an immediate consequence of Theorem~\ref{rand:positive-synthesis}, we obtain the following estimate for sums of indicator functions. For a finite collection $\mathcal F$ of the groups constructed above and coefficients $a=(a_B)_{B\in\mathcal F}$, define
\[
 T_{\mathcal F}a=\sum_{B\in\mathcal F}a_B\mathbf1_{E_B}.
\]
Applying \eqref{rand:eq-positive-synthesis} with coefficients $a_BV_B^{1/s}$ gives
\begin{equation}
\label{rand:eq-operator-bound}
 \|T_{\mathcal F}a\|_{L^s(\mu_K)}
 \le C_{d,s}
       \left(\sum_{B\in\mathcal F}|a_B|^sV_B\right)^{1/s},
 \qquad 1\le s<r_d.
\end{equation}
The constant depends only on $d$ and $s$ and is independent of the choice of $\mathcal F$.

\subsection{Square Function Estimates for Cap Indicators}

Recall that $V_B=\mu_K(E_B)$. For $1\le q<\infty$ and coefficients $u=(u_B)$ indexed by the groups $B$, with only finitely many nonzero entries, define
\begin{equation}\label{rand:eq-square-function}
 \mathfrak S_q(u)
 =\left\|\left(
   \sum_B|u_B|^2V_B^{-2/q}\mathbf1_{E_B}
 \right)^{1/2}\right\|_{L^q(\mu_K)}.
\end{equation}

The following corollary bounds $\mathfrak S_q(u)$ in terms of the coefficient norm $\|u\|_{\ell^t}$. These estimates will be used in randomized compression.

\begin{corollary}
\label{rand:strong-square}
Let $1\le q<\infty$, and let $Q_d$ and $\Gamma_d(q)$ be as in Theorem~\ref{main:randomized}. For every coefficient family $u=(u_B)$ with only finitely many nonzero entries, the estimate
\begin{equation}
\label{rand:eq-strong-square}
 \mathfrak S_q(u)\le C\|u\|_{\ell^t},
 \qquad t=\frac1{1-\gamma},
\end{equation}
holds in either of the following cases:
\begin{enumerate}
\item[(1)] $q<Q_d$ and $\gamma=1-1/q$, so that $t=q$.
\item[(2)] $q\ge Q_d$ and $1/2\le\gamma<\Gamma_d(q)$.
\end{enumerate}
The constant depends only on $d,q$ and, in the second case, on $\gamma$.
\end{corollary}

\begin{proof}
For $1\le q\le2$, the $\ell^2$ norm of a sequence is at most its $\ell^q$ norm. Thus we have
\[
 \left(
   \sum_B|u_B|^2V_B^{-2/q}\mathbf1_{E_B}(x)
 \right)^{q/2}
 \le
 \sum_B|u_B|^qV_B^{-1}\mathbf1_{E_B}(x).
\]
Integrating and using $V_B=\mu_K(E_B)$, we obtain
\[
 \mathfrak S_q(u)^q
 \le\sum_B|u_B|^qV_B^{-1}\mu_K(E_B)
 =\sum_B|u_B|^q.
\]
This proves the claim with $t=q$.

For $2<q<Q_d$, we have $1<q/2<r_d$ by \eqref{rand:eq-geometry-parameters} and the convention $r_2=\infty$. Applying \eqref{rand:eq-positive-synthesis} with $s=q/2$ and coefficients $|u_B|^2$ gives
\[
 \begin{aligned}
 \mathfrak S_q(u)^2
 &=\left\|
   \sum_B|u_B|^2V_B^{-2/q}\mathbf1_{E_B}
   \right\|_{L^{q/2}(\mu_K)}\\
 &\le C_{d,q}
       \left(\sum_B|u_B|^q\right)^{2/q}.
 \end{aligned}
\]
Taking square roots proves the remaining part of the first case.

Now suppose $q\ge Q_d$, so that $d\ge3$. Recall that each $B$ consists of indices $i$ assigned to the same covering set $E_B$. We first bound $\mathfrak S_q(u)$ when at most $N$ coefficients $u_B$ are nonzero. We will then apply this bound to each collection $\{B:2^{-k-1}<|u_B|\le2^{-k}\}$ and sum over $k$ to obtain an estimate independent of $N$.

Recall from \eqref{rand:eq-geometry-parameters} that $r_d=Q_d/2$. Since $\gamma<\Gamma_d(q)=\frac12+\frac{r_d-1}{q}$, we can choose $1<s<r_d$, depending only on $d,q,\gamma$, such that
\[
 \gamma<\gamma_1:=\frac12+\frac{s-1}{q}
 <\Gamma_d(q)<1.
\]
The condition $q\ge Q_d$ then gives $q/2\ge r_d>s$. Let $\mathcal F$ be a collection of $N\ge1$ groups. For the operator
\[
 T_{\mathcal F}a
 =\sum_{B\in\mathcal F}a_B\mathbf1_{E_B},
\]
H\"older's inequality with exponents $q/(2s)$ and $q/(q-2s)$ gives
\[
 \begin{aligned}
 |T_{\mathcal F}a(x)|^{q/2}
 &\le N^{q/2-s}
 \left(\sum_{B\in\mathcal F}
       |a_B|^{q/(2s)}\mathbf1_{E_B}(x)\right)^s.
 \end{aligned}
\]
Integrating and applying \eqref{rand:eq-operator-bound} with exponent $s$ and coefficients $|a_B|^{q/(2s)}$, we obtain
\begin{equation}
\label{rand:eq-operator-q-half}
 \begin{aligned}
 \|T_{\mathcal F}a\|_{L^{q/2}(\mu_K)}
 &\le N^{1-2s/q}
 \left\|
   \sum_{B\in\mathcal F}|a_B|^{q/(2s)}\mathbf1_{E_B}
 \right\|_{L^s(\mu_K)}^{2s/q}\\
 &\le C_{d,q,s}N^{1-2s/q}
       \left(
         \sum_{B\in\mathcal F}|a_B|^{q/2}V_B
       \right)^{2/q}.
 \end{aligned}
\end{equation}

For coefficients $u_B$ that vanish outside $\mathcal F$, take
$a_B=|u_B|^2V_B^{-2/q}$. Then
\[
 \mathfrak S_q(u)^2
 =\|T_{\mathcal F}a\|_{L^{q/2}(\mu_K)},
 \qquad
 \sum_{B\in\mathcal F}|a_B|^{q/2}V_B
 =\sum_{B\in\mathcal F}|u_B|^q.
\]
By \eqref{rand:eq-operator-q-half} and using
$\|u\|_{\ell^q}\le N^{1/q}\max_{B\in\mathcal F}|u_B|$, we obtain
\begin{equation}
\label{rand:eq-square-finite-family}
 \mathfrak S_q(u)
 \le C_{d,q,s}N^{1/2-s/q}\|u\|_{\ell^q}
 \le C_{d,q,s}N^{1-\gamma_1}
       \max_{B\in\mathcal F}|u_B|.
\end{equation}

We now derive a bound independent of the number of nonzero coefficients $N$. The case $u=0$ is immediate. Otherwise, by homogeneity, assume that $\sum_B|u_B|^t=1$. Then $|u_B|\le1$ for every $B$. Partition the indices with nonzero coefficients into
\[
 \mathcal F_k
 =\{B:2^{-k-1}<|u_B|\le2^{-k}\},
 \qquad k\ge0.
\]
The normalization gives
\[
 \#\mathcal F_k
 \le 2^{(k+1)t}
       \sum_{B\in\mathcal F_k}|u_B|^t
 \le 2^{(k+1)t}.
\]
Applying the triangle inequality in $L^q(\mu_K)$ and then \eqref{rand:eq-square-finite-family} to each nonempty $\mathcal F_k$, we obtain
\[
 \begin{aligned}
 \mathfrak S_q(u)
 &\le\sum_{k\ge0}
 \left\|\left(
   \sum_{B\in\mathcal F_k}
     |u_B|^2V_B^{-2/q}\mathbf1_{E_B}
 \right)^{1/2}\right\|_{L^q(\mu_K)}\\
 &\le C_{d,q,s}
       \sum_{k=0}^{\infty}
       2^{-k}2^{(k+1)t(1-\gamma_1)}.
 \end{aligned}
\]
This geometric series converges because $t(1-\gamma_1)
 =\frac{1-\gamma_1}{1-\gamma}<1$. Rescaling $u$ proves \eqref{rand:eq-strong-square}.
\end{proof}

\subsection{Local Compression and Sampling}
\label{rand:construction}

Consider a finite ReLU sum with weighted coefficient sum at most one. Using the groups of indices $B$ constructed above, write
\begin{equation} \label{rand:eq-group-data}
    \begin{aligned}
 F&=\sum_i c_i\phi_{z_i}=\sum_B F_B,
 & F_B&=\sum_{i\in B}c_i\phi_{z_i},\\
 M_B&=\sum_{i\in B}|c_i|w_{K,\eta}(z_i),
 & \sum_B M_B&\le1.
    \end{aligned}
\end{equation}
For $1\le q<\infty$ and $\varepsilon=1/q-\eta\ge0$, recall that $V_B=\mu_K(E_B)$ and set
\[
 \lambda_B=V_B^\varepsilon,
 \qquad
 \psi_B=V_B^{-1/q}\mathbf1_{E_B}.
\]
By the construction preceding Theorem~\ref{rand:positive-synthesis}, each $E_B$ is a convex body and satisfies
\[
 C_{z_i}\subset E_B,\qquad
 |E_B|\le A|C_{z_i}|,
 \qquad
 A=\max\{2C_d,\tau_0^{-1}\},
 \qquad i\in B,
\]
where $A$ depends only on $d$.
For every integer $k_B\ge1$, apply Proposition~\ref{rand:local} with $E=E_B$, $m=k_B$, and $w_i=w_{K,\eta}(z_i)$. The quantities $\Theta$ and $M$ in that proposition are then
$V_B$ and $M_B$, respectively, and
$V_B^{-\eta}\mathbf1_{E_B}=\lambda_B\psi_B$.
We obtain a random sum
\begin{equation}\label{rand:eq-local-sum}
 G_B^{(k_B)}=\sum_{i\in B}\widetilde c_i\phi_{z_i}.
\end{equation}
The coefficient vector of $G_B^{(k_B)}$ takes finitely many values and has at most $4k_B$ nonzero entries. Its coefficients are unbiased, and its pointwise error satisfies
\begin{equation}\label{rand:eq-local-input}
 \mathbb E\widetilde c_i=c_i\quad\forall i\in B,
 \qquad
 |F_B-G_B^{(k_B)}|
 \le C_{d,\eta}\lambda_BM_Bk_B^{-\alpha}\psi_B
 \quad\text{in every realization}.
\end{equation}
In every realization, $\widetilde c_i c_i\ge0$, with $\widetilde c_i=0$ whenever $c_i=0$, and
\[
 \sum_{i\in B}|\widetilde c_i|w_{K,\eta}(z_i)=M_B.
\]

The same volume bounds give $V_B/A\le\theta_{z_i}\le V_B$. Together with $0\le\phi_{z_i}\le h_{z_i}\mathbf1_{C_{z_i}}$ and $C_{z_i}\subset E_B$, this yields
\begin{equation}\label{rand:eq-atom-envelope}
 \left|\frac{\phi_{z_i}}{w_{K,\eta}(z_i)}\right|
 \le \theta_{z_i}^{-\eta}\mathbf1_{E_B}
 \le C_{d,\eta}V_B^{-\eta}\mathbf1_{E_B}
 =C_{d,\eta}\lambda_B\psi_B,
 \qquad i\in B.
\end{equation}

We now combine the local approximations provided by Proposition~\ref{rand:local} with random sampling to approximate $F$.
\begin{lemma}
\label{rand:hybrid}
Let $1\le q<\infty$. Let $F=\sum_i c_i\phi_{z_i}$ be the finite ReLU sum considered above, with coefficients $c_i\in\mathbb R$ and parameters $z_i\in Z_K$. Partition its groups of indices $B$ into two collections $H$ and $R$. Choose integers $k_B\ge1$ for $B\in H$, and set
\[
 A=\sum_{B\in R}M_B.
\]
Suppose $A>0$, and let $L\ge1$ be an integer. Draw $L$ indices independently from $\bigcup_{B\in R}B$, choosing index $i$ with probability $|c_i|w_{K,\eta}(z_i)/A$. For each $B\in R$, let $N_B$ be the number of draws that select an index in $B$, counting repetitions.

There is a finite-valued random coefficient vector $(c_i')$, with $\mathbb Ec_i'=c_i$ for every $i$, such that
\[
 G=\sum_i c_i'\phi_{z_i}
\]
has width at most $4\sum_{B\in H}k_B+L$ in every realization. The coefficients satisfy, in every realization,
\[
 \begin{gathered}
 c_i'c_i\ge0,\qquad
 c_i'=0\quad\text{whenever }c_i=0,\\
 \sum_i|c_i'|w_{K,\eta}(z_i)
 =\sum_i|c_i|w_{K,\eta}(z_i).
 \end{gathered}
\]
Set
\[
 U_B=
 \begin{cases}
 \lambda_BM_Bk_B^{-\alpha},&B\in H,\\
 (A/L)\lambda_B\sqrt{N_B},&B\in R.
 \end{cases}
\]
Then
\begin{equation}\label{rand:eq-hybrid}
 \left(\mathbb E\|F-G\|_q^q\right)^{1/q}
 \le C_{d,q,\eta}
      \left(\mathbb E\mathfrak S_q(U)^q\right)^{1/q},
\end{equation}
where $\mathfrak S_q$ is defined in \eqref{rand:eq-square-function}.

If $A=0$, there is a finite-valued random coefficient vector satisfying the same coefficient conditions for which $G$ has width at most $4\sum_{B\in H}k_B$ and satisfies \eqref{rand:eq-hybrid}, with $U_B=\lambda_BM_Bk_B^{-\alpha}$ for $B\in H$ and $U_B=0$ for $B\in R$.
\end{lemma}

\begin{proof}
For each $B\in H$, take $G_B^{(k_B)}$ as in \eqref{rand:eq-local-sum}, making these choices independently for different groups. Suppose first that $A>0$. Independently of these replacements, draw $L$ indices $I_1,\ldots,I_L$ independently with
\[
 \mathbb P(I_r=i)
 =\frac{|c_i|w_{K,\eta}(z_i)}{A},
 \qquad i\in\bigcup_{B\in R}B.
\]
Define
\[
 X_r=\frac{A}{L}\operatorname{sgn}(c_{I_r})
          \frac{\phi_{z_{I_r}}}{w_{K,\eta}(z_{I_r})},
 \qquad
 G=\sum_{B\in H}G_B^{(k_B)}+\sum_{r=1}^L X_r.
\]
Each draw contributes a coefficient with the sign of the selected $c_i$ and weighted absolute value $A/L$. Contributions at the same index have the same sign, so the $L$ draws have total weighted coefficient sum $A$. For every $i\in\bigcup_{B\in R}B$,
\[
 \mathbb Ec_i'
 =\frac{A\,\operatorname{sgn}(c_i)}{Lw_{K,\eta}(z_i)}
   \sum_{r=1}^L\mathbb P(I_r=i)
 =\frac{A\,\operatorname{sgn}(c_i)}{w_{K,\eta}(z_i)}
   \frac{|c_i|w_{K,\eta}(z_i)}{A}
 =c_i.
\]
Thus $G$ has width at most $4\sum_{B\in H}k_B+L$ and weighted coefficient sum $\sum_{B\in H}M_B+A=\sum_BM_B$. Each coefficient $c_i'$ has the sign of $c_i$ or is zero, and satisfies $\mathbb Ec_i'=c_i$.

Set $e_B=F_B-G_B^{(k_B)}$ for $B\in H$. By construction, the random functions $e_B$ and $X_r$ are mutually independent, and $\mathbb Ee_B=0$ by \eqref{rand:eq-local-input}. Since $\sum_{r=1}^L\mathbb EX_r=\sum_{B\in R}F_B$, we have
\[
 F-G=\sum_{B\in H}e_B+\sum_{r=1}^L(\mathbb EX_r-X_r).
\]
If $I_r\in B$, then $X_r$ vanishes outside $E_B$. However,
\[
 \mathbb E X_r=\frac1L\sum_{D\in R}F_D,
\]
so $\mathbb E X_r-X_r$ may be nonzero outside $E_B$. We introduce independent copies and random signs below to obtain \eqref{rand:eq-error-bound}. The right hand side of that bound contains $\sum_{r=1}^L|X_r|^2$, and each $X_r$ can be estimated by \eqref{rand:eq-atom-envelope} with the group $B$ containing $I_r$.

Let $e_B'$ and $X_r'$ be independent copies of all the random functions $e_B$ and $X_r$, independent of the original functions, and let $\mathbb E'$ denote expectation over these copies. Also introduce mutually independent Rademacher random variables $\epsilon_B$ and $\epsilon_r$, each taking the values $\pm1$ with equal probability and independent of all these functions. Write $\mathbb E_\epsilon$ for expectation over the signs.

By Jensen's inequality and the symmetry of the independent differences, we have
\[
 \begin{aligned}
 \mathbb E\|F-G\|_q^q
 &\le \mathbb E\mathbb E'
       \left\|
       \sum_{B\in H}(e_B-e_B')
       +\sum_{r=1}^L(X_r'-X_r)
       \right\|_q^q\\
 &=\mathbb E\mathbb E'\mathbb E_\epsilon
       \left\|
       \sum_{B\in H}\epsilon_B(e_B-e_B')
       +\sum_{r=1}^L\epsilon_r(X_r'-X_r)
       \right\|_q^q\\
 &\le 2^q\mathbb E\mathbb E_\epsilon
       \left\|
       \sum_{B\in H}\epsilon_Be_B
       -\sum_{r=1}^L\epsilon_rX_r
       \right\|_q^q.
 \end{aligned}
\]
The last inequality follows from
$\|u-v\|_q^q\le2^{q-1}(\|u\|_q^q+\|v\|_q^q)$. Now fix a realization of the functions $e_B$ and $X_r$. At each $x\in K$, the Khintchine inequality \cite{Haagerup1981} gives
\[
 \mathbb E_\epsilon
 \left|
 \sum_{B\in H}\epsilon_Be_B(x)
 -\sum_{r=1}^L\epsilon_rX_r(x)
 \right|^q
 \le C_q
 \left(
 \sum_{B\in H}|e_B(x)|^2
 +\sum_{r=1}^L|X_r(x)|^2
 \right)^{q/2}.
\]
Integrating over $K$ with respect to $\mu_K$ and absorbing $2^q$ into $C_q$, we obtain
\begin{equation} \label{rand:eq-error-bound}
    \mathbb E\|F-G\|_q^q
 \le C_q\mathbb E
       \left\|
       \left(
       \sum_{B\in H}|e_B|^2+\sum_{r=1}^L|X_r|^2
       \right)^{1/2}
       \right\|_q^q.
\end{equation}

By \eqref{rand:eq-local-input}, for any $B\in H$, we have $|e_B|\le C_{d,\eta}U_B\psi_B$. If $I_r\in B\in R$, then \eqref{rand:eq-atom-envelope} gives
\[
 |X_r|\le C_{d,\eta}(A/L)\lambda_B\psi_B.
\]
Therefore, we have
\[
 \sum_{B\in H}|e_B|^2+\sum_{r=1}^L|X_r|^2
 \le C_{d,\eta}\sum_B U_B^2\psi_B^2.
\]
Substituting this estimate into \eqref{rand:eq-error-bound} and using the definition of $\mathfrak S_q$ proves \eqref{rand:eq-hybrid}. If $A=0$, take $G=\sum_{B\in H}G_B^{(k_B)}$ and apply the same argument with all sampling terms omitted.
\end{proof}

We now apply Lemma~\ref{rand:hybrid} with an $\ell^t$ bound for $\mathfrak S_q$ and suitable choices of $k_B$ and $L$.
\begin{lemma}\label{rand:strong-minimum}
Let $1\le t\le q<\infty$ and set $\gamma=1-1/t$. Let $F=\sum_i c_i\phi_{z_i}$, with only finitely many nonzero coefficients $c_i$, and let $M_B$ be as in \eqref{rand:eq-group-data}, with $\sum_BM_B\le1$.  Assume that
\begin{equation}\label{rand:eq-strong-assumption}
 \mathfrak S_q(u)\le A_0\|u\|_{\ell^t}
\end{equation}
for every finitely supported family $u=(u_B)$, where $\mathfrak S_q$ is defined in \eqref{rand:eq-square-function}. Assume also that
\begin{equation}\label{rand:eq-lambda-bound}
 \lambda_B=V_B^{1/q-\eta}\le\Lambda
 \qquad\text{for every group }B.
\end{equation}

Let $N$ be a positive integer such that $M_B>0$ for at most $N$ groups. For every integer $m\ge1$, there is a finite-valued random coefficient vector $(c_i')$ satisfying the coefficient conditions in Lemma~\ref{rand:hybrid}, such that $G=\sum_i c_i'\phi_{z_i}$ has width at most $Cm$ in every realization and
\begin{equation}\label{rand:eq-strong-minimum}
 \left(\mathbb E\|F-G\|_q^q\right)^{1/q}
 \le C\Lambda m^{-\alpha}
       \min\{N,m\}^{(\alpha-\gamma)_+}.
\end{equation}
In particular, the left-hand side is at most $C\Lambda m^{-\min\{\alpha,\gamma\}}$, independently of $N$. The constant $C$ depends only on $d,q,\eta,t,\alpha,A_0$.
\end{lemma}

\begin{proof}
We consider only groups with $M_B>0$. If there are none, take $G=0$.

First suppose $N\le m$. We apply local compression to every group. To choose the integers $k_B$, first allow them to take any positive real values with $\sum_B k_B=m$. The sum $\sum_B(\lambda_BM_B)^t k_B^{-\alpha t}$ in the first line of \eqref{rand:eq-finite-allocation} is then minimized when $k_B$ is proportional to $(\lambda_BM_B)^{1/(\alpha+1/t)}$. Rounding these values up gives
\[
 \rho=\frac1{\alpha+1/t},
 \qquad
 Z=\sum_B(\lambda_BM_B)^\rho,
 \qquad
 k_B=\left\lceil
       \frac{m(\lambda_BM_B)^\rho}{Z}
      \right\rceil.
\]
Then $\sum_B k_B\le m+N\le2m$. Lemma~\ref{rand:hybrid}, with $R=\varnothing$, therefore gives an approximation of width at most $8m$. Since $t(1-\alpha\rho)=\rho$, equations \eqref{rand:eq-hybrid} and \eqref{rand:eq-strong-assumption} give
\begin{equation}\label{rand:eq-finite-allocation}
 \begin{aligned}
 \left(\mathbb E\|F-G\|_q^q\right)^{1/q}
 &\le C
       \left(\sum_B
       (\lambda_BM_Bk_B^{-\alpha})^t
       \right)^{1/t}\\
 &\le C\left(\frac Zm\right)^\alpha Z^{1/t}
  =Cm^{-\alpha}Z^{1/\rho}.
 \end{aligned}
\end{equation}
Using $\lambda_B\le\Lambda$ and $\sum_BM_B\le1$, we obtain
\[
 Z^{1/\rho}
 \le\Lambda\left(\sum_BM_B^\rho\right)^{1/\rho}
 \le\Lambda N^{(1/\rho-1)_+}
 =\Lambda N^{(\alpha-\gamma)_+}.
\]
This proves \eqref{rand:eq-strong-minimum} for $N\le m$.

Now suppose $N>m$. If we chose $k_B=\lceil mM_B\rceil$ for every group, each group with $0<M_B<1/m$ would have $k_B=1$. We apply local compression to the groups with $M_B\ge1/m$ and sample from the remaining groups together. Set
\[
 H=\{B:M_B\ge1/m\},
 \qquad
 R=\{B:0<M_B<1/m\},
\]
and choose $k_B=\lceil mM_B\rceil$ for $B\in H$. Since $\sum_BM_B\le1$, we have
\[
 \#H\le m,
 \qquad
 \sum_{B\in H}k_B
 \le m\sum_{B\in H}M_B+\#H
 \le2m.
\]
Let $\delta=\min\{\alpha,\gamma\}$. Since $0\le\delta\le\alpha$ and $k_B\ge mM_B\ge1$, the definition of $U_B$ in Lemma~\ref{rand:hybrid} and \eqref{rand:eq-lambda-bound} give
\[
 \|(U_B)_{B\in H}\|_{\ell^t}
 \le\Lambda m^{-\delta}
       \left(\sum_{B\in H}M_B^{t(1-\delta)}\right)^{1/t}
 \le\Lambda m^{-\delta},
\]
where the last inequality uses $t(1-\delta)\ge1$ and $\sum_BM_B\le1$.

Set $A=\sum_{B\in R}M_B$. If $A=0$, Lemma~\ref{rand:hybrid} and \eqref{rand:eq-strong-assumption} give the desired error bound, with width at most $8m$. Suppose henceforth that $A>0$. Choose $L=\lceil mA\rceil$ draws as in Lemma~\ref{rand:hybrid}. Then $1\le L\le m$ and $A/L\le1/m$. For each $B\in R$, the number $N_B$ of draws selecting an index in $B$ has binomial distribution with mean $LM_B/A$. Since $L\le mA+1$, $M_B<1/m$, and $M_B\le A$, we obtain
\[
 \mathbb E N_B=\frac{LM_B}{A}
 \le mM_B+\frac{M_B}{A}
 \le2.
\]
For $1\le q\le2$, the count $N_B$ satisfies $N_B^{q/2}\le N_B$, so $\mathbb E N_B^{q/2}\le\mathbb E N_B$. For $q>2$, set $s=\lceil q/2\rceil$ and $p_B=M_B/A$. Since $N_B\sim\operatorname{Bin}(L,p_B)$, $\binom{L}{k}\le L^k/k!$, and $Lp_B\le2$, we have
\[
 \begin{aligned}
 \mathbb E N_B^{q/2}
 &\le \mathbb E N_B^s
 =\sum_{k=1}^L k^s\binom{L}{k}p_B^k(1-p_B)^{L-k}\\
 &\le\sum_{k=1}^{\infty}\frac{k^s(Lp_B)^k}{k!}\\
 &\le Lp_B\sum_{k=1}^{\infty}\frac{k^s2^{k-1}}{k!}
 =C_q\mathbb EN_B.
 \end{aligned}
\]
Thus, for every $q\ge1$, we have 
$\sum_{B\in R}\mathbb E N_B^{q/2}
 \le C_q\sum_{B\in R}\mathbb EN_B
 =C_qL$.

In every realization, at most $L$ of the counts $N_B$ are nonzero. Since $t\le q$, H\"older's inequality on the at most $L$ nonzero coordinates gives $\|(U_B)_{B\in R}\|_{\ell^t}
 \le L^{1/t-1/q}\|(U_B)_{B\in R}\|_{\ell^q}$. Taking $q$-th moments and using $U_B=(A/L)\lambda_B\sqrt{N_B}$ and \eqref{rand:eq-lambda-bound}, we obtain
\[
 \begin{aligned}
 \left(
 \mathbb E\|(U_B)_{B\in R}\|_{\ell^t}^q
 \right)^{1/q}
 &\le \frac{\Lambda A}{L}\,
       L^{1/t-1/q}
       \left(
       \sum_{B\in R}\mathbb E N_B^{q/2}
       \right)^{1/q}\\
 &\le C_q\Lambda A L^{-\gamma}\\
 &\le C_q\Lambda m^{-\gamma}A^{1-\gamma}
 \le C_q\Lambda m^{-\gamma},
 \end{aligned}
\]
where we used $L\ge mA$ and $A\le1$ in the last line.

Combining the bounds for $H$ and $R$ by Minkowski's inequality and applying \eqref{rand:eq-hybrid} and \eqref{rand:eq-strong-assumption}, we obtain
\[
 \left(\mathbb E\|F-G\|_q^q\right)^{1/q}
 \le C\Lambda\bigl(m^{-\delta}+m^{-\gamma}\bigr)
 \le C\Lambda m^{-\delta}.
\]
Since $m^{-\delta}=m^{-\alpha}m^{(\alpha-\gamma)_+}$, this is \eqref{rand:eq-strong-minimum} for $N>m$. The width is at most $4\sum_{B\in H}k_B+L\le9m$. 
This completes the proof.
\end{proof}

\subsection{Approximation Rates from Effective Cap Counts}
\label{rand:profiles-section}

We use the groups and notation from \eqref{rand:eq-group-data}. For a fixed integer $J\ge0$, we use \eqref{rand:eq-finite-allocation} for the groups at scales $\tau_0,\ldots,\tau_J$ and for the additional group with $E_B=K$, if present. We apply Lemma~\ref{rand:strong-minimum} to all remaining groups.


By \eqref{eq:domain-error} and Lemma~\ref{lem:closure}, it suffices to prove the approximation bounds uniformly for finite sums $F$ as in \eqref{rand:eq-group-data}.
\begin{proposition}
\label{rand:profile}
Let $1\le t\le q<\infty$ and assume \eqref{rand:eq-strong-assumption}. Recall that $\alpha=(d+3)/(2d)$, and set
\[
 \gamma=1-\frac1t,
 \qquad
 a=\alpha-\gamma,
 \qquad
 \varepsilon=\frac1q-\eta.
\]
Assume also that $a>0$ and $\varepsilon\ge0$. For each integer $J\ge0$, define
\[
 D_J=1+\sum_{j=0}^J\mathcal C_K(\tau_j),
 \qquad
 Q_J=1+\sum_{j=0}^J
             \mathcal C_K(\tau_j)\tau_j^{\varepsilon/a}.
\]
Here $\mathcal C_K$ is defined in \eqref{intr:eq-profile}. 

There are constants $c,C>0$, depending only on $d,q,\eta,t,A_0$, such that, for all integers $n\ge1$ and $J\ge0$ satisfying $D_J\le cn$,
\begin{equation}\label{rand:eq-profile}
 \mathcal E_n^K(w_{K,\eta};q)
 \le C\left(n^{-\alpha}Q_J^a
             +n^{-\gamma}\tau_{J+1}^\varepsilon\right).
\end{equation}
For every finite sum $F=\sum_i c_i\phi_{z_i}$ as in \eqref{rand:eq-group-data}, there is an approximant $G=\sum_i c_i'\phi_{z_i}$ of width at most $n$, with $\|F-G\|_q$ bounded by the right-hand side of \eqref{rand:eq-profile} and
\[
 \sum_i|c_i'|w_{K,\eta}(z_i)
 \le \sum_i|c_i|w_{K,\eta}(z_i).
\]
\end{proposition}

\begin{proof}
Fix a finite sum $F$ as in \eqref{rand:eq-group-data} and omit groups with $M_B=0$. Let $\mathcal B_J$ denote the collection of all groups at scales $\tau_0,\ldots,\tau_J$, together with the additional group with covering set $K$, if present.  Write
\[
 F_{\le J}=\sum_{B\in\mathcal B_J}F_B,
 \qquad
 F_{>J}=F-F_{\le J}.
\]
When either collection of groups is empty, use the zero approximation for that part.

By Lemma~\ref{hil:economic-cover}, we have $\#\mathcal B_J\le C_dD_J$, where we take $C_d\ge1$. Choose $c\le(40C_d)^{-1}$ and set $m=\lfloor n/20\rfloor$. Since $D_J\ge1$, the condition $D_J\le cn$ implies $n\ge40$ and
\[
 \#\mathcal B_J
 \le C_dD_J
 \le \frac n{40}
 \le m.
\]

For the approximation of $F_{\le J}$, set
\[
 \rho=(\alpha+1/t)^{-1}=\frac1{1+a},
 \qquad
 Z=\sum_{B\in\mathcal B_J}(\lambda_BM_B)^\rho.
\]
Since $0<\rho<1$ and $1-\rho=a\rho$, H\"older's inequality with exponents $1/\rho$ and $1/(1-\rho)$ gives
\[
 \begin{aligned}
 Z^{1/\rho}
 =\left(
   \sum_{B\in\mathcal B_J}
   M_B^\rho(\lambda_B^{1/a})^{1-\rho}
   \right)^{1/\rho}\le
 \left(\sum_{B\in\mathcal B_J}M_B\right)
 \left(\sum_{B\in\mathcal B_J}\lambda_B^{1/a}\right)^a.
 \end{aligned}
\]
At scale $j$, Lemma~\ref{hil:economic-cover} gives at most $C_d\mathcal C_K(\tau_j)$ groups, each with a covering cap of normalized volume  $V_B\le C_d\tau_j$. Since $\lambda_B=V_B^\varepsilon$, it follows that
\[
 \sum_{B\in\mathcal B_J}\lambda_B^{1/a}
 \le 1+C\sum_{j=0}^J
              \mathcal C_K(\tau_j)\tau_j^{\varepsilon/a}
 \le C Q_J.
\]
Together with $\sum_BM_B\le1$, this yields $Z^{1/\rho}\le C Q_J^a$. The construction giving \eqref{rand:eq-finite-allocation}, applied with parameter $m$, therefore produces a random approximant $G_{\le J}$ of width at most $8m$ such that
\[
 \left(\mathbb E\|F_{\le J}-G_{\le J}\|_q^q\right)^{1/q}
 \le C m^{-\alpha}Z^{1/\rho}
 \le C n^{-\alpha}Q_J^a.
\]

Every group contributing to $F_{>J}$ has scale $j\ge J+1$. We have $\lambda_B=V_B^\varepsilon
 \le C\tau_j^\varepsilon
 \le C\tau_{J+1}^\varepsilon$. Apply Lemma~\ref{rand:strong-minimum} to this finite collection with parameter $m$ and $\Lambda=C\tau_{J+1}^\varepsilon$. Since $\alpha>\gamma$, the construction in its proof gives a random approximant $G_{>J}$ of width at most $9m$ satisfying
\[
 \left(\mathbb E\|F_{>J}-G_{>J}\|_q^q\right)^{1/q}
 \le C m^{-\gamma}\tau_{J+1}^\varepsilon
 \le C n^{-\gamma}\tau_{J+1}^\varepsilon.
\]

Set $G=G_{\le J}+G_{>J}$. Its width is at most $17m\le n$, and Minkowski's inequality gives
\begin{equation}\label{rand:eq-total-error-moment}
 \left(\mathbb E\|F-G\|_q^q\right)^{1/q}
 \le C\left(n^{-\alpha}Q_J^a
             +n^{-\gamma}\tau_{J+1}^\varepsilon\right).
\end{equation}
Both constructions use original indices and preserve the weighted sum of absolute coefficients on their respective index sets. Since these index sets are disjoint, $G$ satisfies the coefficient assertion in every realization. Estimate~\eqref{rand:eq-total-error-moment} guarantees a realization with $L^q$ error bounded by the right-hand side of \eqref{rand:eq-profile}, proving the result for finite sums.
\end{proof}

We next apply Proposition~\ref{rand:profile} when $\mathcal C_K(\tau)$ satisfies a power bound.

\begin{corollary}\label{rand:power}
Set $a=\alpha-\gamma$. Under the hypotheses of Proposition~\ref{rand:profile}, suppose that $\mathcal C_K(\tau)\le C_{\mathrm{cap}}\tau^{-\kappa}$ for some $\kappa>0$, $C_{\mathrm{cap}}>0$, and all $0<\tau\le\tau_*$, where $\tau_*\in(0,1)$ is fixed. Then, for every integer $n\ge1$,
\begin{equation}\label{rand:eq-strong-power}
 \mathcal E_n^K(w_{K,\eta};q)\le C
 \begin{cases}
 n^{-\alpha},&\varepsilon>\kappa a,\\
 n^{-\alpha}(1+\log n)^a,&\varepsilon=\kappa a,\\
 n^{-\gamma-\varepsilon/\kappa},
       &0\le\varepsilon<\kappa a.
 \end{cases}
\end{equation}
The constant $C$ depends only on $d,q,\eta,t,A_0,\kappa,C_{\mathrm{cap}}$, and $\tau_*$.
\end{corollary}

\begin{proof}
Enlarging $C$ to cover the finitely many scales with $\tau_j>\tau_*$ gives $\mathcal C_K(\tau_j)\le C2^{\kappa j}$ for all $j\ge0$. Consequently, for some $C_D>0$,
\[
 D_J\le C_D2^{\kappa J},
 \qquad
 Q_J\le C\left(
 1+\sum_{j=0}^J2^{j(\kappa-\varepsilon/a)}
 \right).
\]

Let $c$ be the constant in Proposition~\ref{rand:profile}. For $n\ge C_D/c$, choose $J\ge0$ to be the largest integer satisfying $C_D2^{\kappa J}\le cn$. Then $D_J\le cn$, $2^{\kappa J}\asymp n$, and $J+1\le C(1+\log n)$. Summing the geometric series gives
\[
 Q_J\le C
 \begin{cases}
 1,&\varepsilon>\kappa a,\\
 J+1,&\varepsilon=\kappa a,\\
 2^{J(\kappa-\varepsilon/a)},
       &0\le\varepsilon<\kappa a.
 \end{cases}
\]
By the maximality of $J$, $\tau_{J+1}\le Cn^{-1/\kappa}$. Since $\varepsilon\ge0$, the second term in \eqref{rand:eq-profile} satisfies $n^{-\gamma}\tau_{J+1}^{\varepsilon}\le Cn^{-\gamma-\varepsilon/\kappa}$. When $\varepsilon\ge\kappa a$, this is at most $Cn^{-\alpha}$ because $\alpha=\gamma+a$. When $0\le\varepsilon<\kappa a$, the first term satisfies
\[
 n^{-\alpha}Q_J^a
 \le Cn^{-\alpha}2^{J(\kappa a-\varepsilon)}
 \le Cn^{-\gamma-\varepsilon/\kappa}.
\]
These estimates and $J+1\le C(1+\log n)$ give the three cases of \eqref{rand:eq-strong-power} upon substitution into \eqref{rand:eq-profile}.

For $1\le n<C_D/c$, the zero approximation gives $\mathcal E_n^K(w_{K,\eta};q)\le1$ by \eqref{geom:eq-atomnorm} and \eqref{eq:intrinsic-weight}. Enlarging $C$ proves the result for every integer $n\ge1$.
\end{proof}

\begin{proof}[Proof of Theorem~\ref{main:randomized}]
Recall that $\alpha=(d+3)/(2d)$. For $d=2$, we have $\Gamma_2(q)<1<\alpha$. For $d\ge3$, the definition of $Q_d$ in \eqref{main:eq-universal-parameters} gives $\Gamma_d(q)\le 1-\frac1{Q_d}
 <\frac12+\frac1d<\alpha$. Choose
\[
 \gamma=\Gamma_d(q)=1-\frac1q \quad\text{if }q<Q_d,
 \qquad
 \frac12\le\gamma<\Gamma_d(q) \quad\text{if }q\ge Q_d,
\]
and set $t=(1-\gamma)^{-1}$. Then $1\le t\le q$, and Corollary~\ref{rand:strong-square} gives \eqref{rand:eq-strong-assumption}. Moreover, $a=\alpha-\gamma>0$, as required in Proposition \ref{rand:profile} and Corollary~\ref{rand:power}. By \eqref{eq:domain-error} and Lemma~\ref{lem:closure}, it suffices to prove the error bounds uniformly for finite sums $F$ as in \eqref{rand:eq-group-data}.

First suppose $\eta=1/q$. Then $\lambda_B=V_B^{1/q-\eta}=1$ for every group. With $\Lambda=1$, Lemma~\ref{rand:strong-minimum} gives an approximation of width at most $C_{\mathrm w}m$, where $C_{\mathrm w}\ge1$ is independent of $m$ and $F$. Choose $m=\lfloor n/C_{\mathrm w}\rfloor$. For all sufficiently large $n$, we have $m\asymp n$ and $C_{\mathrm w}m\le n$, so the approximation $G$ has width at most $n$. Since $\alpha>\gamma$, the lemma gives
\[
 \left(\mathbb E\|F-G\|_q^q\right)^{1/q}
 \le Cn^{-\gamma}
\]
uniformly over the finite sums in \eqref{rand:eq-group-data}. Choosing a realization with error no greater than this bound yields
\[
 \mathcal E_n^K(w_{K,1/q};q)\le Cn^{-\gamma}.
\]
For $q<Q_d$, this proves the rate in \eqref{main:eq-universal-critical-rate} for all sufficiently large $n$. For $q\ge Q_d$, the estimate holds for every $\gamma$ in the stated range, so the definition of $r_*$ in \eqref{intr:eq-order} gives $r_*(K,1/q,q)\ge\Gamma_d(q)$.

Now suppose $\varepsilon=1/q-\eta>0$. By \eqref{intr:eq-kappa}, for every $\kappa>\overline\kappa(K)$ there is a constant $C_{K,\kappa}$ such that
\[
 \mathcal C_K(\tau)\le C_{K,\kappa}\tau^{-\kappa}
\]
for all sufficiently small $\tau>0$. Corollary~\ref{rand:power} and the definition of $r_*$ therefore give
\[
 r_*(K,\eta,q)
 \ge\min\{\alpha,\gamma+\varepsilon/\kappa\}.
\]
In the case $\varepsilon=\kappa a$, the logarithmic factor in \eqref{rand:eq-strong-power} permits every exponent strictly below $\alpha$, which suffices for this conclusion by \eqref{intr:eq-order}.

If $\overline\kappa(K)>0$, let $\kappa\downarrow\overline\kappa(K)$ and, when $q\ge Q_d$, let $\gamma\uparrow\Gamma_d(q)$. This proves \eqref{main:eq-universal-order}. If $\overline\kappa(K)=0$, fix $\gamma$ and take $\kappa>0$ sufficiently small that $\gamma+\varepsilon/\kappa\ge\alpha$. Thus $r_*(K,\eta,q)\ge\alpha$.

To prove the attained rate in \eqref{main:eq-universal-classical}, suppose that
\[
 \varepsilon>\overline\kappa(K)\bigl(\alpha-\Gamma_d(q)\bigr).
\]
Choose $\gamma=\Gamma_d(q)$ when $q<Q_d$, or choose $\gamma<\Gamma_d(q)$ sufficiently close to $\Gamma_d(q)$ when $q\ge Q_d$, so that
\[
 \varepsilon>\overline\kappa(K)(\alpha-\gamma).
\]
We can then choose $\kappa>\overline\kappa(K)$ sufficiently close to $\overline\kappa(K)$ that $\varepsilon>\kappa(\alpha-\gamma)$. With these parameters fixed, the first case of \eqref{rand:eq-strong-power} gives
\[
 \mathcal E_n^K(w_{K,\eta};q)\le C_{K,q,\eta}n^{-\alpha}.
\]

The preceding argument applies to all sufficiently large $n$. For smaller $n$, including $n=0$, the zero approximation gives $\mathcal E_n^K(w_{K,\eta};q)\le1$ by \eqref{geom:eq-atomnorm} and \eqref{eq:intrinsic-weight}. Enlarging the constants gives the stated bounds for all $n\ge0$.

The constructions in Lemma~\ref{rand:strong-minimum}
and Proposition~\ref{rand:profile} use only original
atoms and do not increase the weighted coefficient sum
in any realization.
With the preceding choices of width and auxiliary
exponents, they give \eqref{main:eq-compression}
for the explicit error bounds and for every exponent
strictly below the stated lower bound for $r_*$.
The zero approximants for small $n$ satisfy these
conditions as well.
For $M>0$, applying the construction to $F/M$ and
scaling back gives the factor $M$ in the error bound.
For $M=0$, take $S_n=0$.
\end{proof}

\section{Lower Bounds and Optimal Approximation Exponents}
\label{sec:lower-classification}

In Proposition~\ref{geo:packing-lower}, we use caps
of a fixed relative volume $\tau$ with pairwise
disjoint interiors to prove a lower bound in terms
of their number and $\tau$. We then prove Theorem~\ref{geo:interior-lower},
establishing $\alpha$ as a universal upper bound
on the optimal approximation exponent for $q\ge2$.

For convex bodies with $C^2$ boundary, we prove
Lemma~\ref{geo:c2-packing}, which gives two-sided
estimates for the maximum number of caps of relative
volume $\tau$ with pairwise disjoint interiors,
for all sufficiently small $\tau>0$.
We apply the lower estimate in
Proposition~\ref{geo:packing-lower} to obtain a lower
bound for the approximation error.
Combining this bound with the lower bound in
Theorem~\ref{geo:interior-lower} and the upper bounds
in Theorem~\ref{main:randomized} proves
Theorem~\ref{main:c2}.

\subsection{General Lower Bounds}

We first prove a lower bound from disjoint caps, which
will be used in the proofs of Theorems~\ref{main:c2},
\ref{main:curved}, and~\ref{main:intrinsic}.
We then give a separate proof of
Theorem~\ref{geo:interior-lower}.
For $0<\tau\le1$, define
\begin{equation}\label{geo:packing}
 P_K(\tau)=\max\left\{m\in\mathbb{N}:
 \begin{array}{l}
 \text{there are $m$ caps of relative volume $\tau$}\\
 \text{whose interiors are pairwise disjoint}
 \end{array}
 \right\}.
\end{equation}
Since caps with pairwise disjoint interiors have total relative volume at most one, $P_K(\tau)\le1/\tau$. Moreover, it is clear that  $P_K$ is nonincreasing.

\begin{proposition} \label{geo:packing-lower}
Let $1\le q<\infty$ and $\eta\in\mathbb R$, and write $b_q=1-1/q$. There exist constants $C_d,c_{d,q}>0$, independent of $K$ and $\eta$, such that, for every integer $n\ge0$ and every $0<\tau\le1$ satisfying
\[
 P_K(\tau)\ge m\quad\text{and}
 \quad m=\left\lceil C_d(n+1)\log(n+2)\right\rceil,
\]
we have
\begin{equation}\label{geo:eq-packing-lower}
 \mathcal E_n^K(w_{K,\eta};q)
 \ge c_{d,q}m^{-b_q}\tau^{1/q-\eta}.
\end{equation}
\end{proposition}

\begin{proof}
Since $P_K(\tau)\ge m$, we can choose caps $C_i=C_{z_i}$, $1\le i\le m$, with pairwise disjoint interiors and $\theta_{z_i}=\tau$. Write $h_i=h_{z_i}$ and $\phi_i=\phi_{z_i}$, and set
\[
 D_i=\{x\in K:\phi_i(x)\ge h_i/2\}.
\]
Since $K$ is compact, $\phi_i$ attains its maximum $h_i$ at some $v_i\in K$. The map $x\mapsto(v_i+x)/2$ sends $C_i$ into $D_i\subset C_i$ and multiplies volume by $2^{-d}$. Hence, we have 
\[
 2^{-d}\tau\le\mu_K(D_i)\le\tau.
\]
The sets $D_i$ also have pairwise disjoint interiors.

For each sign vector $\sigma\in\{-1,1\}^m$, define the finite ReLU sum
\[
 f_\sigma=\frac1m\sum_{i=1}^m
       \frac{\sigma_i}{h_i\tau^\eta}\phi_i.
\]
Since $w_{K,\eta}(z_i)=h_i\tau^\eta$, the coefficients satisfy $\sum_{i=1}^mw_{K,\eta}(z_i)
 \left|\sigma_i/(mh_i\tau^\eta)\right|
 =1$. Since the caps $C_i$ have pairwise disjoint interiors, $\phi_j=0$ almost everywhere on $D_i$ for $j\ne i$. Thus
\begin{equation}\label{geo:eq-target-amplitude}
 \sigma_i f_\sigma(x)\ge A_0/2
 \quad\text{for almost every }x\in D_i,
 \qquad A_0=m^{-1}\tau^{-\eta}.
\end{equation}

Next, we prove that, for some $\sigma\in\{-1,1\}^m$,
$\inf_{S\in\Sigma_n^K}\|f_\sigma-S\|_q
\ge c_0A_0(m\tau)^{1/q}$,
where $c_0=c_0(d,q)>0$ will be chosen below.
Suppose, for a contradiction, that for every
$\sigma\in\{-1,1\}^m$ there exists a network
$S_\sigma\in\Sigma_n^K$ such that
\begin{equation}\label{geo:eq-assumed-error}
 \|f_\sigma-S_\sigma\|_q<c_0A_0(m\tau)^{1/q}.
\end{equation}
Fix one such network $S_\sigma$ for each $\sigma$.

We will find one common set of points $x_i\in D_i$
with the following property: for at least $2^{m-1}$
vectors $\sigma$, $\operatorname{sign}S_\sigma(x_i)$
differs from $\sigma_i$ for at most $m/8$ indices $i$.
The VC dimension bound limits the number of sign
patterns that the networks can produce on these points.
For each pattern, we also bound the number of vectors
$\sigma$ that differ from it in at most $m/8$ coordinates.
Taking the constant $C_d$ in the definition of $m$
large enough makes the product of these two upper
bounds smaller than $2^{m-1}$, a contradiction.

Consider the class of sign functions
\[
 \operatorname{sign}(\Sigma_n^K)
 =\{x\mapsto\operatorname{sign}(S(x)):S\in\Sigma_n^K\},
\]
where $\operatorname{sign}(u)=1$ for $u\ge0$ and $\operatorname{sign}(u)=-1$ for $u<0$. 

By the definition of Vapnik--Chervonenkis (VC) dimension in \cite[Section~4.3.2]{Wainwright2019}, the VC dimension of $\operatorname{sign}(\Sigma_n^K)$ is the largest integer $r$ for which there exist distinct points $x_1,\ldots,x_r\in K$ satisfying
\[
 \left\{
 \bigl(\operatorname{sign}(S(x_1)),\ldots,
       \operatorname{sign}(S(x_r))\bigr):
 S\in\Sigma_n^K
 \right\}
 =\{-1,1\}^r.
\]
Varying $S\in\Sigma_n^K$ produces all $2^r$ sign patterns on these fixed points. Since these networks have fixed depth and at most $(d+2)n$ parameters, \cite[Theorem~7]{Bartlett2019} gives
\[
 \operatorname{VCdim}\bigl(\operatorname{sign}(\Sigma_n^K)\bigr)
 \le v_n,
 \qquad
 v_n=\left\lceil C'_d(n+1)\log(n+2)\right\rceil.
\]
Choose the constant $C_d$ in the definition of $m$
large enough that $m\ge K_0v_n$, where $K_0\ge8$
is an absolute constant to be fixed below.

Let $X_1,\ldots,X_m$ be independent random points,
with $X_i$ uniformly distributed on $D_i$. For $u,v\in\{-1,1\}^m$, define their Hamming distance by
\[
 d_H(u,v)=\#\{i\in\{1,\ldots,m\}:u_i\ne v_i\}.
\]
By \eqref{geo:eq-target-amplitude}, for almost every $x\in D_i$ we have
\begin{equation}\label{geo:eq-sign-error}
 \operatorname{sign}(S_\sigma(x))\ne\sigma_i
 \quad\Longrightarrow\quad
 |f_\sigma(x)-S_\sigma(x)|
 \ge |\sigma_i f_\sigma(x)|\ge A_0/2.
\end{equation}
Since $X_i$ is uniform on $D_i$, we use \eqref{geo:eq-sign-error} and $\mu_K(D_i)\ge2^{-d}\tau$ to bound the probability that $\operatorname{sign}(S_\sigma(X_i))\ne\sigma_i$. Summing over $i$ and applying \eqref{geo:eq-assumed-error}, we obtain
\[
 \begin{aligned}
 \mathbb E\,d_H\bigl(\sigma,
       (\operatorname{sign}(S_\sigma(X_i)))_{i=1}^m\bigr)
 &=\sum_{i=1}^m
   \mathbb P\bigl(\operatorname{sign}(S_\sigma(X_i))\ne\sigma_i\bigr)\\
 &\le \frac{2^{d+q}}{\tau A_0^q}
       \sum_{i=1}^m\int_{D_i}|f_\sigma-S_\sigma|^q\,d\mu_K\\
 &\le \frac{2^{d+q}}{\tau A_0^q}
       \|f_\sigma-S_\sigma\|_q^q\\
 &<2^{d+q}c_0^qm.
 \end{aligned}
\]
The second inequality holds because the sets $D_i$ overlap only on sets of measure zero. Choose $c_0=c_0(d,q)>0$ so that $2^{d+q}c_0^q\le1/16$. For each $\sigma$, Markov's inequality gives $\mathbb P\left(
 d_H\bigl(\sigma,(\operatorname{sign}(S_\sigma(X_i)))_{i=1}^m\bigr)
 \le \frac m8
 \right)\ge\frac12
$. Define the random variable
\[
 Z=\#\left\{\sigma\in\{-1,1\}^m:
 d_H\bigl(\sigma,(\operatorname{sign}(S_\sigma(X_i)))_{i=1}^m\bigr)
 \le \frac m8\right\}.
\]
We have $\mathbb EZ\ge \frac12\,2^m=2^{m-1}$. Consequently, there exist fixed points $x_i\in D_i$, $1\le i\le m$, such that
\begin{equation}\label{geo:eq-fixed-sample}
 d_H\bigl(\sigma,(\operatorname{sign}(S_\sigma(x_i)))_{i=1}^m\bigr)
 \le \frac m8
\end{equation}
for at least $2^{m-1}$ vectors $\sigma\in\{-1,1\}^m$.

Since $\operatorname{VCdim}(\operatorname{sign}(\Sigma_n^K))\le v_n$ and $1\le v_n\le m$, the Sauer--Shelah bound in \cite[Proposition~4.18 and Exercise~4.11(b)]{Wainwright2019} gives
\begin{equation}\label{geo:eq-sign-count}
 \#\left\{
 \bigl(\operatorname{sign}(S(x_i))\bigr)_{i=1}^m:
 S\in\Sigma_n^K
 \right\}
 \le \sum_{j=0}^{v_n}\binom mj
 \le \left(\frac{em}{v_n}\right)^{v_n}.
\end{equation}
For each $S\in\Sigma_n^K$, choosing the $j$ coordinates where $\sigma$ differs from $(\operatorname{sign}(S(x_i)))_{i=1}^m$ uniquely determines $\sigma$. Thus there are exactly $\binom mj$ vectors at Hamming distance $j$, and the number at distance at most $m/8$ is
\begin{equation}\label{geo:eq-hamming-count}
 \sum_{j=0}^{\lfloor m/8\rfloor}\binom mj
 \le 2^{mH_2(1/8)},
 \qquad
 H_2(s)=-s\log_2s-(1-s)\log_2(1-s).
\end{equation}
To prove the inequality, set $p=1/8$. Since $p/(1-p)<1$, for $0\le j\le pm$ we have $p^j(1-p)^{m-j}\ge p^{pm}(1-p)^{(1-p)m}=2^{-mH_2(p)}$. The binomial theorem therefore gives
\[
 2^{-mH_2(p)}
 \sum_{j=0}^{\lfloor pm\rfloor}\binom mj
 \le
 \sum_{j=0}^{\lfloor pm\rfloor}\binom mj p^j(1-p)^{m-j}
 \le (p+(1-p))^m=1,
\]
which proves the bound.
For each distinct vector counted in \eqref{geo:eq-sign-count}, \eqref{geo:eq-hamming-count} bounds the number of $\sigma$ within Hamming distance $m/8$. Summing these bounds and using \eqref{geo:eq-fixed-sample}, we obtain
\[
 2^{m-1}
 \le \left(\frac{em}{v_n}\right)^{v_n}2^{mH_2(1/8)}.
\]

Choose $K_0\ge8$ large enough that $\frac{\log_2(eK_0)}{K_0}+H_2(1/8)\le\frac34$. Since $m/v_n\ge K_0$ and the function $r\mapsto\log_2(er)/r$ is decreasing for $r\ge1$, we obtain
\[
 2^{m-1}
 \le (em/v_n)^{v_n}2^{mH_2(1/8)}
 \le 2^{3m/4}
 <2^{m-1},
\]
where the last inequality uses $m\ge K_0v_n\ge8$. This is a contradiction.

Consequently, for some $\sigma\in\{-1,1\}^m$,
\[
 \inf_{S\in\Sigma_n^K}\|f_\sigma-S\|_q
 \ge c_0A_0(m\tau)^{1/q}
 =c_0m^{-b_q}\tau^{1/q-\eta}.
\]
Since $f_\sigma$ was constructed with a weighted sum of absolute coefficients equal to one, this proves \eqref{geo:eq-packing-lower}.
\end{proof}

We next prove Theorem~\ref{geo:interior-lower} by transferring an $L^2$ approximation lower bound on the Euclidean ball to a ball contained in $K$.

\begin{proof}[Proof of Theorem~\ref{geo:interior-lower}]
Since invertible affine changes of variables preserve $\mathcal E_n^K(w_{K,\eta};q)$, John's ellipsoid theorem \cite{John1948} allows us to assume that $B^d\subset K\subset dB^d$.

For the ReLU dictionary
\[
 \mathbb P_1^d
 =\{y\mapsto(\xi\cdot y-s)_+:|\xi|=1,\ |s|\le1\}
 \quad\text{on }B^d,
\]
\cite[equation~(1.24)]{Siegel2025} implies
\begin{equation}\label{geo:eq-siegel-lower}
 \sup_{g\in B_1(\mathbb P_1^d)}
 \inf_{T\in\Sigma_N(\mathbb P_1^d)}
 \|g-T\|_{L^2(\mu_{B^d})}
 \ge c_d(N\log N)^{-\alpha},
 \qquad N\ge2,\quad \alpha=\frac{d+3}{2d}.
\end{equation}
Here $B_1(\mathbb P_1^d)$ is the $L^2$ closure of finite linear combinations of elements of $\mathbb P_1^d$ with the unweighted sum of absolute coefficients at most one, and $\Sigma_N(\mathbb P_1^d)$ consists of linear combinations of at most $N$ elements of $\mathbb P_1^d$ with unrestricted real coefficients.

To apply \eqref{geo:eq-siegel-lower}, we show that $\Sigma_n^{B^d}\subseteq\Sigma_{n+2}(\mathbb P_1^d)$. For $u\ne0$, positive homogeneity gives
\[
 (u\cdot y+b)_+=|u|(\xi\cdot y-s)_+,
 \qquad \xi=\frac{u}{|u|},\quad s=-\frac{b}{|u|}.
\]
If $|s|\le1$, the normalized atom belongs to $\mathbb P_1^d$. Otherwise, it is zero or affine on $B^d$. Terms with $u=0$ are constant.

All affine and constant contributions combine into a single function $v\cdot y+c$. Choose $e=v/|v|$ if $v\ne0$, and any unit vector otherwise. Since $(\pm e\cdot y+1)_+=1\pm e\cdot y$ on $B^d$, we have
\[
 v\cdot y+c
 =\frac{c+|v|}{2}(e\cdot y+1)_+
  +\frac{c-|v|}{2}(-e\cdot y+1)_+.
\]
Both atoms on the right belong to $\mathbb P_1^d$. There are at most $n$ normalized terms with $|s|\le1$, and the combined affine contribution requires at most two additional atoms. Hence $\Sigma_n^{B^d}\subseteq\Sigma_{n+2}(\mathbb P_1^d)$. Apply \eqref{geo:eq-siegel-lower} with $N=n+2$. Since $\Sigma_n^{B^d}\subseteq\Sigma_{n+2}(\mathbb P_1^d)$, the lower bound also holds for approximation by $\Sigma_n^{B^d}$.

By \eqref{closure:eq-suprema} in Lemma~\ref{lem:closure}(1), applied with $X=L^2(\mu_{B^d})$ and $w\equiv1$, the supremum over $B_1(\mathbb P_1^d)$ is unchanged when restricted to finite sums of elements of $\mathbb P_1^d$ with unweighted sum of absolute coefficients at most one. Combining this with \eqref{geo:eq-siegel-lower} and the
preceding inclusion, and decreasing $c_d$ if necessary,
we can, for each integer $n\ge0$, choose a finite ReLU
sum $g=g_n:\mathbb R^d\to\mathbb R$ of the form
\[
 g(y)=\sum_j a_j(\xi_j\cdot y-s_j)_+,
 \qquad
 \sum_j|a_j|\le1,\quad |\xi_j|=1,\quad |s_j|\le1,
\]
whose restriction to $B^d$ belongs to
$B_1(\mathbb P_1^d)$ and satisfies
\begin{equation}\label{geo:eq-ball-lower}
 \inf_{T\in\Sigma_n^{B^d}}
 \|g-T\|_{L^2(\mu_{B^d})}
 \ge c_d\bigl((n+1)\log(n+2)\bigr)^{-\alpha}.
\end{equation}

We rescale $g$ to obtain a finite ReLU sum on $K$ with weighted sum of absolute coefficients at most one. Recall that $h_{(\xi,t)}=H_K(\xi)-t$ and
$\theta_{(\xi,t)}=|C_{(\xi,t)}|/|K|$.
For $|\xi|=1$ and $|t|\le1/2$, the inclusions
$B^d\subset K\subset dB^d$ imply
\[
 \frac12\le h_{(\xi,t)}\le d+\frac12,
 \qquad
 \frac{|\frac{1}{4}B^d|}{|dB^d|}=(4d)^{-d}\le\theta_{(\xi,t)}\le1.
\]
Indeed, the cap $C_{(\xi,t)}$ contains the ball of radius $1/4$ centered at $3\xi/4$. By \eqref{eq:intrinsic-weight}, there is therefore a constant $W=W_{d,\eta}>0$ such that $w_{K,\eta}(\xi,t)\le W$ for all these parameters. Define
\[
 F(x)=\frac{g(2x)}{2W}
     =\sum_j\frac{a_j}{W}(\xi_j\cdot x-s_j/2)_+,
 \qquad x\in K.
\]
Since $|s_j/2|\le1/2$, its coefficients satisfy
\[
 \sum_j\frac{|a_j|}{W}w_{K,\eta}(\xi_j,s_j/2)
 \le\sum_j|a_j|\le1.
\]
Hence $F\in\mathcal A_{w_{K,\eta}}(K)$ by the definition
of $\mathcal A_w(K)$ in Subsection~\ref{Weighted spaces}.

Set $D=\tfrac12B^d\subset K$. For any $S\in\Sigma_n^K$, restriction to $D$ and the inequality $\|u\|_{L^q(\mu_D)}\ge\|u\|_{L^2(\mu_D)}$ for $q\ge2$ give
\[
 \begin{aligned}
 \|F-S\|_{L^q(\mu_K)}
 &\ge \left(\frac{|D|}{|K|}\right)^{1/q}
       \|F-S\|_{L^q(\mu_D)}\\
 &\ge \left(\frac{|D|}{|K|}\right)^{1/q}
       \|F-S\|_{L^2(\mu_D)}\\
 &=\frac1{2W}\left(\frac{|D|}{|K|}\right)^{1/q}
       \|g-2W S(\,\cdot\,/2)\|_{L^2(\mu_{B^d})}.
 \end{aligned}
\]
The transformed network $2W S(\,\cdot\,/2)$ belongs to $\Sigma_n^{B^d}$, and $|D|/|K|\ge(2d)^{-d}$. Applying \eqref{geo:eq-ball-lower} proves \eqref{geo:eq-interior-lower}. 
\end{proof}

\subsection{Optimal Exponents for \texorpdfstring{$C^2$}{C2} Convex Bodies}
\label{hil:c2}

Recall from \eqref{geo:packing} that $P_K(\tau)$ is
the largest number of caps of $K$ with relative
volume $\tau$ and pairwise disjoint interiors.
The following lemma gives the cap packing estimate
used in the proof of Theorem~\ref{main:c2} and later
in the proof of Theorem~\ref{main:curved}.

\begin{lemma}
\label{geo:c2-packing}
Let $K\subset\mathbb R^d$, $d\ge2$, be a convex body
with $C^2$ boundary, and put $\kappa_d=(d-1)/(d+1)$.
There are constants $c_K,\tau_K>0$ such that, for
$0<\tau<\tau_K$,
\[
 c_K\tau^{-\kappa_d}
 \le P_K(\tau)
 \le\mathcal C_K(\tau)
 \le C_d\tau^{-\kappa_d}.
\]
In particular, $\kappa(K)$ exists and equals $\kappa_d$.
\end{lemma}

\begin{proof}
At each $p\in\partial K$, let $\nu_p$ be the outward
unit normal and write nearby points as
\[
 x=p+y-s\nu_p,\qquad y\in\nu_p^\perp,\quad s\in\mathbb R.
\]
Since $\partial K$ is $C^2$, there is a $C^2$ function
$g_p$, defined on an open neighborhood of $0$ in
$\nu_p^\perp$, such that, for sufficiently small $y$ and $s$,
\begin{equation}\label{geo:eq-local-graph}
\begin{aligned}
 p+y-s\nu_p\in\partial K
 &\quad\Longleftrightarrow\quad s=g_p(y),\\
 p+y-s\nu_p\in K
 &\quad\Longleftrightarrow\quad s\ge g_p(y), 
\end{aligned}
\end{equation}
see \cite[pp.~94 and~354]{GilbargTrudinger2001} or  \cite[Section~2.5, pp.~112--113]{Schneider2014}. 
Thus the boundary is parametrized locally by
$y\mapsto p+y-g_p(y)\nu_p$.
Since $p\in\partial K$, we have $g_p(0)=0$.
The tangent hyperplane at $p$ is $p+\nu_p^\perp$,
so $Dg_p(0)=0$.
Here $Dg_p$ and $D^2g_p$ denote the gradient and Hessian
with respect to the tangential variable $y$, computed
in orthonormal coordinates on $\nu_p^\perp$. In these orthonormal tangent coordinates, the eigenvalues
of $D^2g_p(0)$ are the principal curvatures at $p$,
with the convention that spheres have positive curvature
\cite[Section~14.6, p.~354, Eq.~(14.93)]{GilbargTrudinger2001} or \cite[Section~2.5, pp.~114--115]{Schneider2014}.

\medskip
\noindent\textit{Step 1. A boundary patch with positive curvature.}  We first find an open boundary patch
(a nonempty relatively open subset of $\partial K$)
on which all principal curvatures are positive. 
After a translation, assume that $0\in\operatorname{int}K$.
Set $R=\max_{x\in K}|x|$ and choose $p_0\in\partial K$
with $|p_0|=R$.
Since $K\subset RB^d$, the tangent hyperplane to
$RB^d$ at $p_0$ also supports $K$ at $p_0$.
Thus the outward unit normal there is $\nu=p_0/R$.

For all sufficiently small
$y\in\nu^\perp$, the boundary point
$p_0+y-g_{p_0}(y)\nu$ belongs to $RB^d$, so $|y|^2+(R-g_{p_0}(y))^2\le R^2$. Consequently,
\[
 g_{p_0}(y)\ge R-\sqrt{R^2-|y|^2}.
\]
The difference between the left- and right-hand sides
has a local minimum at $0$, so
\[
 D^2g_{p_0}(0)\ge R^{-1}I.
\]
Thus all principal curvatures at $p_0$ are at least
$R^{-1}>0$. By continuity, they remain positive
on an open neighborhood of $p_0$ in $\partial K$.

\medskip
\noindent\textit{Step 2. Cap volume and radius estimates.} Choose a compact subset $\Gamma$ of positive surface
measure contained in this open patch.
The principal curvatures are continuous and positive
on $\Gamma$, so they have a positive lower bound
and a finite upper bound.
By $C^2$ regularity and compactness, there are
$r>0$ and $0<\lambda\le\Lambda$ such that all tangent
graphs $g_p$ are defined for $|y|<r$ and satisfy
\begin{equation} \label{geo:eq-hessian}
    \lambda I\le D^2g_p(y)\le\Lambda I,
 \qquad
 \forall p\in\Gamma,\ y\in\nu_p^\perp,\ |y|<r.
\end{equation}
Since $g_p(0)=0$ and $Dg_p(0)=0$, Taylor's formula gives
\begin{equation}\label{geo:eq-quadratic}
 \frac{\lambda}{2}|y|^2\le g_p(y)
 \le\frac{\Lambda}{2}|y|^2,
\end{equation}
uniformly for these $p$ and $y$. To estimate the volume of a cap at $p$ using $g_p$, we need the whole cap to lie in the neighborhood where \eqref{geo:eq-local-graph} holds. We will show that this is true when its height is below a fixed positive number, the same for every $p\in\Gamma$.

The supporting hyperplane at each $p\in\Gamma$
meets $K$ only at $p$.
Indeed, suppose that it also contains
$q\in K\setminus\{p\}$.
By convexity, the line segment $[p,q]$ lies in $K$
and in this hyperplane, hence in $\partial K$.
Set $v=q-p\in\nu_p^\perp$.
For sufficiently small $t>0$, the boundary point
$p+tv$ satisfies $s=g_p(y)$ with $y=tv$ and $s=0$.
Thus $g_p(tv)=0$, contradicting
$g_p(tv)\ge \frac{\lambda}{2}t^2|v|^2>0$.

Define the cap of height $h$ at $p$ by
\begin{equation}\label{geo:eq-cap-height}
 C(p,h)=K\cap\{x:\nu_p\cdot(p-x)\le h\}.
\end{equation}
By $C^2$ regularity and compactness of $\Gamma$,
choose a common radius $0<r_0<r$ such that, for each
$p\in\Gamma$, \eqref{geo:eq-local-graph} holds with
the corresponding $g_p$ whenever $|y|^2+s^2<r_0^2$.  Since the supporting hyperplane at $p$ meets $K$
only at $p$, we have $\nu_p\cdot(p-x)>0$ for
$x\in K\setminus\{p\}$.
By continuity on the compact set
$\{(p,x)\in\Gamma\times K:|x-p|\ge r_0\}$,
\begin{equation}\label{geo:eq-cap-gap}
 \inf_{\substack{p\in\Gamma,\ x\in K\\|x-p|\ge r_0}}
 \nu_p\cdot(p-x)>0.
\end{equation}
By \eqref{geo:eq-cap-height} and \eqref{geo:eq-cap-gap},
we have $C(p,h)\subset B(p,r_0)$ for all sufficiently
small $h>0$, uniformly in $p\in\Gamma$. 
Using \eqref{geo:eq-local-graph} in the orthonormal
coordinates $x=p+y-s\nu_p$ and integrating along
the normal direction $\nu_p$, we obtain
\[
 |C(p,h)|
 =\int_{\substack{y\in\nu_p^\perp\\|y|<r}}
       (h-g_p(y))_+\,dy.
\]
By \eqref{geo:eq-quadratic}, for sufficiently small
$h>0$, we have
\[
 \int_{\nu_p^\perp}
 \left(h-\frac{\Lambda}{2}|y|^2\right)_+\,dy
 \le |C(p,h)|
 \le
 \int_{\nu_p^\perp}
 \left(h-\frac{\lambda}{2}|y|^2\right)_+\,dy.
\]
The substitution $y=\sqrt h\,z$ therefore gives
\begin{equation}\label{geo:eq-cap-volume}
 \mu_K(C(p,h))
 =\frac{|C(p,h)|}{|K|}
 \asymp_K h^{(d+1)/2}.
\end{equation}
Moreover, by \eqref{geo:eq-quadratic}, every $x=p+y-s\nu_p\in C(p,h)$ satisfies
\[
 \frac{\lambda}{2}|y|^2\le g_p(y)\le s\le h.
\]
Thus $|x-p|^2=|y|^2+s^2\le 2h/\lambda+h^2$,
which gives $C(p,h)\subset B(p,C_K\sqrt h)$.
Both estimates hold uniformly for $p\in\Gamma$
and all sufficiently small $h>0$.

\medskip
\noindent\textit{Step 3. Construction of disjoint caps.} For each $p\in\Gamma$, the function
$h\mapsto\mu_K(C(p,h))$ is continuous and vanishes
at $h=0$. Together with the uniform estimate
\eqref{geo:eq-cap-volume}, this ensures that,
for every sufficiently small $\tau>0$ and every
$p\in\Gamma$, we can choose
$h(p,\tau)>0$ such that
$\mu_K(C(p,h(p,\tau)))=\tau$.
The preceding estimates then give
\begin{equation}\label{geo:eq-cap-scale}
 h(p,\tau)\asymp_K\tau^{2/(d+1)},
 \qquad
 C(p,h(p,\tau))
 \subset B(p,A_K\tau^{1/(d+1)}),
\end{equation}
uniformly for $p\in\Gamma$, where $A_K>0$ depends
only on $K$.

Fix $L>2A_K$ and set $\delta=L\tau^{1/(d+1)}$.
Choose a maximal $\delta$-separated set
$\{p_1,\ldots,p_N\}\subset\Gamma$.
Thus $|p_i-p_j|\ge\delta$ for $i\ne j$, and the set
is maximal with this property. Consequently, $\Gamma\subset\bigcup_{i=1}^N B(p_i,\delta)$.

For $\tau$ small enough that $\delta<r_0$, every
$x\in\Gamma\cap B(p_i,\delta)$ can be written as
$x=p_i+y-s\nu_{p_i}$, where $y\in\nu_{p_i}^{\perp}$
and $|y|^2+s^2<\delta^2$.
Since $x\in\partial K$, \eqref{geo:eq-local-graph}
gives $s=g_{p_i}(y)$, and we have $|y|<\delta$. 
Let $\sigma$ denote surface measure.
By \eqref{geo:eq-hessian} and $Dg_p(0)=0$, we have
$|Dg_p(y)|\le\Lambda|y|$ for $|y|<r$.
The graph area formula
\cite[Section~16.1, p.~390, (16.11)]{GilbargTrudinger2001}
therefore gives
\[
 \sigma\bigl(\Gamma\cap B(p_i,\delta)\bigr)
 \le
 \int_{\substack{y\in\nu_{p_i}^{\perp}\\|y|<\delta}}
 \sqrt{1+|Dg_{p_i}(y)|^2}\,dy
 \le C_K\delta^{d-1},
\]
where $C_K>0$ depends only on $K$.
Therefore
\[
 0<\sigma(\Gamma)
 \le\sum_{i=1}^N
       \sigma\bigl(\Gamma\cap B(p_i,\delta)\bigr)
 \le C_KN\delta^{d-1}.
\]
Hence $N\ge c_K\delta^{-(d-1)}$, where
$c_K=\sigma(\Gamma)/C_K>0$.

By \eqref{geo:eq-cap-scale}, the caps
$C(p_i,h(p_i,\tau))$ lie in balls centered at $p_i$
with radius $A_K\tau^{1/(d+1)}<\delta/2$. Since the centers are at least $\delta$ apart,
these balls, and hence the caps, are pairwise disjoint.
Each cap has relative volume $\tau$, so
\begin{equation}\label{geo:eq-c2-packing-lower}
 P_K(\tau)\ge N\ge c_K'\tau^{-\kappa_d},
\end{equation}
where $c_K'=c_KL^{-(d-1)}>0$, since
$\delta=L\tau^{1/(d+1)}$ and
$\kappa_d=(d-1)/(d+1)$.

Finally, caps of relative volume $\tau$ lie in the
wet part $\{x\in K:v_K(x)\le\tau\}$.
Since their boundaries have measure zero, caps with
pairwise disjoint interiors have additive volumes,
so $\tau P_K(\tau)\le W_K(\tau)$.
Combining this with \eqref{geo:eq-c2-packing-lower}
and Lemma~\ref{rand:wet-bound}, we obtain
\[
 c_K'\tau^{-\kappa_d}
 \le P_K(\tau)
 \le\mathcal C_K(\tau)
 \le C_d\tau^{-\kappa_d}.
\]
Hence $\kappa(K)$ exists and equals $\kappa_d$.
\end{proof}

\begin{proof}[Proof of Theorem~\ref{main:c2}]
Lemma~\ref{geo:c2-packing} gives $\kappa(K)=\kappa_d$.
Set
\[
 \varepsilon=\frac1q-\eta\ge0,
 \qquad
 r=\min\left\{\alpha,\,
               b_q+\frac{\varepsilon}{\kappa_d}\right\}.
\]

We first prove the lower bound.
For each $n\ge1$, put
\[
 m=\left\lceil C_d(n+1)\log(n+2)\right\rceil,
\]
where $C_d$ is the constant in
Proposition~\ref{geo:packing-lower}.
Let $c_K,\tau_K$ be the constants in
Lemma~\ref{geo:c2-packing}.
Choose $c>0$ such that
$c<\tau_K$ and $c_Kc^{-\kappa_d}\ge1$, and set
$\tau=c\,m^{-1/\kappa_d}$.
Since $m\ge1$, we have $\tau<\tau_K$ and
\[
 P_K(\tau)\ge c_K\tau^{-\kappa_d}
             =c_Kc^{-\kappa_d}m\ge m.
\]
Proposition~\ref{geo:packing-lower} therefore gives
\[
 \mathcal E_n^K(w_{K,\eta};q)
 \gtrsim_{d,q} \tau^\varepsilon m^{-b_q}
 =c^\varepsilon m^{-b_q-\varepsilon/\kappa_d}.
\]
Since $m\asymp_d(n+1)\log(n+2)$, this estimate and
Theorem~\ref{geo:interior-lower} give two lower bounds
with exponents $b_q+\varepsilon/\kappa_d$ and $\alpha$.
Taking the larger of these bounds yields
\begin{equation}\label{geo:eq-c2-error-lower}
 \mathcal E_n^K(w_{K,\eta};q)
 \ge c_{K,q,\eta}
       \bigl((n+1)\log(n+2)\bigr)^{-r}.
\end{equation}
Thus, $r_*(K,\eta,q)\le r$.

For the upper bound, Lemma~\ref{geo:c2-packing} gives
$\mathcal C_K(\tau)\le C_d\tau^{-\kappa_d}$
for all sufficiently small $\tau>0$.
Since $\Gamma_d(q)=b_q$ for $q\le Q_d$,
Theorem~\ref{main:randomized} gives $r_*(K,\eta,q)\ge r$.
Thus \eqref{main:eq-c2-exponent} holds.
In particular, for every $0<s<r$, the definition
of $r_*$ gives a constant $C_s>0$ such that
\begin{equation}\label{geo:eq-c2-error-upper}
 \mathcal E_n^K(w_{K,\eta};q)
 \le C_s(n+1)^{-s}.
\end{equation}

It remains to establish the logarithmic limit.
By \eqref{geo:eq-c2-error-upper} and
\eqref{geo:eq-c2-error-lower}, we have
\[
 s
 \le\liminf_{n\to\infty}
     \frac{-\log\mathcal E_n^K(w_{K,\eta};q)}{\log n}
 \le\limsup_{n\to\infty}
     \frac{-\log\mathcal E_n^K(w_{K,\eta};q)}{\log n}
 \le r,
\]
where the last inequality uses
$\log\log(n+2)/\log n\to0$.
Letting $s\uparrow r$ proves
\eqref{main:eq-c2-log-limit}.
The argument includes the critical weight
$\eta=1/q$ and, when finite, the endpoint $q=Q_d$.
\end{proof}

\begin{remark} \label{rem:c2-local-extension}
    The conclusions of Lemma~\ref{geo:c2-packing} and
Theorem~\ref{main:c2} remain valid if $\partial K$
contains a $C^2$ patch (a nonempty relatively open
subset of $\partial K$) on which all principal
curvatures are positive.
No regularity is required elsewhere on the boundary.

In Lemma~\ref{geo:c2-packing}, the global $C^2$
assumption is used only to obtain such a patch.
On a compact subset of positive surface measure,
bounds of the form \eqref{geo:eq-quadratic},
together with the convexity and compactness of $K$,
give the same construction of disjoint caps.
The upper bound follows from
Lemma~\ref{rand:wet-bound} for arbitrary convex bodies.

The proof of Theorem~\ref{main:c2} is then unchanged,
since Theorems~\ref{main:randomized}
and~\ref{geo:interior-lower} also apply to arbitrary
convex bodies.
The ranges of $q$ and $\eta$ remain unchanged,
with $q\ge2$ required by
Theorem~\ref{geo:interior-lower}.
\end{remark}

\section{Geometric Refinements and Optimal Exponents}
\label{sec:geometric-proofs}
\label{geo:section}

In this section, we refine the general upper bounds
of Theorem~\ref{main:randomized} using cap geometry.
We first establish estimates for combining local errors,
then construct cap covers for positively curved bodies,
planar bodies, and products of planar bodies with fixed
polytopes. Combining the resulting upper bounds with
the lower bounds proved earlier yields the corresponding
optimal approximation exponents.

Throughout this section $1\le q<\infty$, the measure is $\mu_K$,
and $w=w_{K,\eta}=h\theta^\eta$. We use
$\varepsilon=1/q-\eta$, $b_q=1-1/q$, and
$a_q=\alpha-b_q$, where $\alpha=(d+3)/(2d)$.

\subsection{Combining Local Errors}

For each group $B$, Lemma~\ref{geom:local} bounds the local approximation error by a multiple of $\mathbf1_{E_B}$, where $E_B$ contains all caps assigned to $B$. For positively curved bodies, we use \eqref{geo:eq-hardy} to bound the $L^q$ norm of the sum over all groups, including when $\varepsilon=0$. For planar bodies and their products with polytopes, we first apply \eqref{geo:layer} within each volume interval and then add the resulting $L^q$ error bounds by the triangle inequality. We use this second argument under the assumption $\varepsilon>0$.


\begin{definition} \label{geo:atlas}
Recall from Subsection~\ref{Weighted spaces} that
\[
 Z_K=\{(\xi,t)\in\mathbb R^d\times\mathbb R:
 |\xi|=1,\ -H_K(-\xi)\le t<H_K(\xi)\}
\]
is the parameter set of the ReLU atoms under consideration. Partition $Z_K$ into a finite or countable collection $\mathcal B$ of nonempty, pairwise disjoint sets. Each $B\in\mathcal B$ is a group of parameters. For each group $B\in\mathcal B$, choose a convex body $E_B\subset K$ containing all caps $C_z$ associated with parameters in $B$, and require
\begin{equation}\label{geo:eq-atlas}
 C_z\subset E_B,\qquad
 A^{-1}\mu_K(E_B)\le\theta_z\le\mu_K(E_B),
 \quad\forall z\in B,
\end{equation}
where $A\ge1$ is independent of $B$ and $z$.
\end{definition}

The groups partition $Z_K$, whereas the sets $E_B\subset K$ may overlap. Such a grouping exists for every convex body $K$. We briefly recall the cap coverings constructed before Proposition~\ref{rand:multiplicity} and use them to group all parameters in $Z_K$. Indeed, for every sufficiently small $\tau>0$, Lemma~\ref{hil:economic-cover} provides finitely many covering caps such that every cap with $\tau/2<\theta_z\le\tau$ is contained in one of them. Assign each such parameter $z$ to exactly one containing cap $E$. The volume bound in that lemma gives
\[
 \theta_z\le\mu_K(E)\le C_d\tau<2C_d\theta_z.
\]
Choose $0<\tau_0<1$ so that Lemma~\ref{hil:economic-cover} applies whenever $0<\tau\le\tau_0$, and repeat the construction at the scales $\tau=\tau_j=2^{-j}\tau_0$, $j\ge0$. Assign all parameters with $\theta_z>\tau_0$ to one additional group, with $K$ as its containing set. Since there are countably many scales and only finitely many groups at each scale, the resulting collection is at most countable. After deleting empty groups, we obtain a partition $\mathcal B$ of $Z_K$ satisfying \eqref{geo:eq-atlas} with $A=\max\{2C_d,\tau_0^{-1}\}$.

For each group $B\in\mathcal B$, set
\[
 \psi_B=\mu_K(E_B)^{-1/q}\mathbf1_{E_B},
 \qquad
 \lambda_B=\mu_K(E_B)^\varepsilon
          =\mu_K(E_B)^{1/q-\eta}.
\]
Then $\|\psi_B\|_{L^q(\mu_K)}=1$.
The factor $\lambda_B$ is the volume factor in the local
$L^q$ error bound \eqref{geom:eq-localq},
with $\Theta=\mu_K(E_B)$.

Consider a finite network $F=\sum_i c_i\phi_{z_i}$ and, for each $B\in\mathcal B$, write
\[
 F_B=\sum_{i:\,z_i\in B}c_i\phi_{z_i},
 \qquad
 M_B=\sum_{i:\,z_i\in B}|c_i|w(z_i).
\]
Thus $F=\sum_{B\in\mathcal B}F_B$, and $M_B$ is the weighted coefficient sum of the terms in group $B$. Since $F$ is a finite sum, only finitely many groups have a nonzero weighted coefficient sum $M_B$. For every integer $m_B\ge0$, Lemma~\ref{geom:local} gives an approximant $S_B$ using at most $m_B$ original atoms from group $B$, with weighted coefficient sum at most $M_B$ by \eqref{geom:eq-localcost}. Since $w(z_i)=h_i\theta_{z_i}^{\eta}$, \eqref{geo:eq-atlas} gives
\[
 \sum_{i:\,z_i\in B}|c_i|h_i
 =\sum_{i:\,z_i\in B}|c_i|w(z_i)\theta_{z_i}^{-\eta}
 \le C_{A,\eta}\mu_K(E_B)^{-\eta}M_B.
\]
Thus \eqref{geom:eq-localpoint}, together with $\lambda_B\psi_B=\mu_K(E_B)^{-\eta}\mathbf1_{E_B}$, yields
\begin{equation}\label{geo:eq-block}
 |F_B-S_B|
 \le C_{d,A,\eta}\lambda_BM_B
       (m_B+1)^{-\alpha}\psi_B.
\end{equation}

\paragraph{\normalfont\itshape Combining errors over all groups.}
Summing the pointwise bounds in \eqref{geo:eq-block} gives
\[
 \left|F-\sum_{B\in\mathcal B}S_B\right|
 \le C_{d,A,\eta}
 \sum_{B\in\mathcal B}
 \bigl[\lambda_BM_B(m_B+1)^{-\alpha}\bigr]
 \mu_K(E_B)^{-1/q}\mathbf1_{E_B}.
\]
To control the total $L^q$ error, we will apply the following
estimate with $u_B=\lambda_BM_B(m_B+1)^{-\alpha}$.
Each parameter $z_i$ of the finite network $F$ belongs
to exactly one group. Hence $M_B$, and therefore $u_B$,
is nonzero for only finitely many groups.
We assume that the sets $E_B$ chosen in
Definition~\ref{geo:atlas} satisfy
\begin{equation}\label{geo:eq-hardy}
 \left\|
   \sum_{B\in\mathcal B}
   u_B\mu_K(E_B)^{-1/q}\mathbf1_{E_B}
 \right\|_{L^q(\mu_K)}
 \le H\left(\sum_{B\in\mathcal B}u_B^q\right)^{1/q}
\end{equation}
for every choice of nonnegative coefficients $u_B$ with only finitely many nonzero terms. For fixed $K$, $q$, and these sets $E_B$, the constant $H<\infty$ is independent of the coefficients and of which groups have nonzero coefficients. For $q=1$, direct integration gives equality with $H=1$. For the grouping obtained from Lemma~\ref{hil:economic-cover}, \eqref{rand:eq-positive-synthesis} gives \eqref{geo:eq-hardy} when $1\le q<r_d$. For positively curved bodies, Proposition~\ref{curvednew:atlas} constructs coverings with uniformly bounded overlap for caps in each volume range $\tau_j/2<\theta_z\le\tau_j$ and proves \eqref{geo:eq-hardy} for every $1\le q<\infty$. 

Combining \eqref{geo:eq-block} with \eqref{geo:eq-hardy}, using $u_B=\lambda_BM_B(m_B+1)^{-\alpha}$, gives
\[
 \left\|F-\sum_{B\in\mathcal B}S_B\right\|_{L^q(\mu_K)}
 \le C_{d,A,\eta}H
 \left(
   \sum_{B\in\mathcal B}
   \bigl[\lambda_BM_B(m_B+1)^{-\alpha}\bigr]^q
 \right)^{1/q}.
\]
In the next subsection, we use this bound to choose the integers $m_B$ with $\sum_{B\in\mathcal B}m_B\le n$, so that the approximating network uses at most $n$ neurons in total.

\paragraph{\normalfont\itshape Combining errors within each volume range.}
For planar bodies and their products with fixed polytopes, Propositions~\ref{intr:planar-cover} and~\ref{intr:product-cover} bound the number and overlap of covering sets for caps with volumes in $(\tau_j/2,\tau_j]$, where $\tau_j=\tau_0 2^{-j}$. The overlap bounds allow us to combine the local estimates in \eqref{geo:eq-block} into an $L^q$ error bound for each volume interval. When $\eta<1/q$, we then apply Proposition~\ref{geo:layer-upper} to sum the resulting estimates over $j$ by the triangle inequality, without using \eqref{geo:eq-hardy}.

For each $j$, choose finitely many convex bodies $E_{ji}\subset K$ such that every cap with $\tau_j/2<\theta_z\le\tau_j$ is contained in at least one of them and $\mu_K(E_{ji})\le C_{d}\tau_j$. Apply the grouping procedure following Definition~\ref{geo:atlas}. Let $N_j$ be the number of nonempty groups at scale $j$, and relabel their containing sets as $E_{j1},\ldots,E_{jN_j}$. The same volume comparison gives
\[
 \tau_j/2<\mu_K(E_{ji})\le C_{d}\tau_j,
 \qquad 1\le i\le N_j.
\]
Including the group for $\theta_z>\tau_0$ with containing set $K$, the resulting partition satisfies \eqref{geo:eq-atlas} with $A=\max\{2C_d,\tau_0^{-1}\}$.

For each $j$, let $S_j(q)$ be the smallest constant such that
\begin{equation}\label{geo:layer}
 \left\|
   \sum_{i=1}^{N_j}
   u_i\mu_K(E_{ji})^{-1/q}\mathbf1_{E_{ji}}
 \right\|_{L^q(\mu_K)}
 \le S_j(q)\left(\sum_{i=1}^{N_j}u_i^q\right)^{1/q}
\end{equation}
for all nonnegative coefficients $u_i$, one for each set $E_{ji}$. With $u_i=\lambda_BM_B(m_B+1)^{-\alpha}$ for the group $B$ corresponding to $E_{ji}$, \eqref{geo:layer} combines the local bounds in \eqref{geo:eq-block} into an $L^q$ error estimate for this volume interval.

It remains to bound $S_j(q)$ in terms of the overlap of the covering sets $E_{ji}$. Let $L_j=\operatorname*{ess\,sup}_{x\in K}
       \sum_{i=1}^{N_j}\mathbf1_{E_{ji}}(x)$, so that almost every point of $K$ belongs to at most $L_j$ of the sets $E_{ji}$. Then
\begin{equation}\label{geo:eq-simple-synthesis}
 1\le S_j(q)\le L_j^{b_q}\le N_j^{b_q}.
\end{equation}
Taking one nonzero coefficient gives the lower bound. For $q=1$, direct integration gives $S_j(1)=1$. For $q>1$, at most $L_j$ summands are nonzero at almost every $x\in K$, so H\"older's inequality gives
\[
 \left(
   \sum_{i=1}^{N_j}
   u_i\mu_K(E_{ji})^{-1/q}\mathbf1_{E_{ji}}(x)
 \right)^q
 \le L_j^{q-1}
   \sum_{i=1}^{N_j}
   u_i^q\mu_K(E_{ji})^{-1}\mathbf1_{E_{ji}}(x).
\]
Integrating over $K$ makes the right-hand side equal to $L_j^{q-1}\sum_i u_i^q$. Taking $q$th roots therefore gives $S_j(q)\le L_j^{1-1/q}=L_j^{b_q}$. Finally, $L_j\le N_j$ gives the last inequality. 

The coverings in Propositions~\ref{intr:planar-cover} and~\ref{intr:product-cover} satisfy $L_j\le C(j+1)^{d-1}$, where $C$ is independent of $j$. Thus \eqref{geo:eq-simple-synthesis} gives
\[
 S_j(q)\le C^{b_q}(j+1)^{(d-1)b_q}.
\]
In particular, $\log S_j(q)/j\to0$, which verifies the condition on $S_j(q)$ in \eqref{geo:eq-subpower}.

\subsection{Upper Bounds in Terms of Cap Geometry}

We now derive upper bounds for $\mathcal E_n^K(w_{K,\eta};q)$
from the estimates for combining local errors in the preceding
subsection.
These bounds describe how the approximation error depends on
the volumes of the covering sets and their number at each
volume scale.
Together with the lower bounds, they will be used in
Subsections~\ref{curvednew:section} and~\ref{intr:section}
to determine the optimal approximation exponents for positively
curved bodies and for planar bodies and their products with
polytopes.

For a bounded nonnegative family $(\lambda_B)_{B\in\mathcal B}$, define
\begin{equation} \label{geo:eq-rearrangement}
    \lambda_k^*=\inf\bigl\{t\ge0:
             \#\{B\in\mathcal B:\lambda_B>t\}\le k-1\bigr\},
 \qquad k\ge1.
\end{equation}
For a finite family, these are its values in nonincreasing order, followed by zeros. 

\begin{proposition}
\label{geo:hardy-upper}
\label{geo:profile}
Let $\mathcal B$ and $(E_B)_{B\in\mathcal B}$ be as in Definition~\ref{geo:atlas}, and recall that $b_q=1-1/q$ and $a_q=\alpha-b_q$. Suppose that \eqref{geo:eq-hardy} holds with constant $H$ and that $\sup_{B\in\mathcal B}\lambda_B<\infty$. Then, for every integer $n\ge0$,
\begin{equation}\label{geo:eq-profile-upper}
 \mathcal E_n^K(w_{K,\eta};q)\le CH(n+1)^{-\alpha}
 \begin{cases}
 \displaystyle
 \left(\sum_{k=1}^{n+1}(\lambda_k^*)^{1/a_q}\right)^{a_q},
                  &a_q>0,\\[3pt]
 \displaystyle\sup_{B\in\mathcal B}\lambda_B,&a_q\le0.
 \end{cases}
\end{equation}
Here $C$ depends only on $d,q,A,\eta$. The following conclusions also hold.

\begin{enumerate}
\item[(1)] \emph{Finite networks.}
For every finite network $F=\sum_i c_i\phi_{z_i}$ with $\sum_i|c_i|w_{K,\eta}(z_i)\le1$, there is an approximant whose $L^q(\mu_K)$ error is bounded by the right-hand side of \eqref{geo:eq-profile-upper}. It uses at most $n$ atoms appearing in $F$ and does not increase the weighted coefficient sum.

\item[(2)] \emph{Weighted variation unit ball.}
After enlarging $C$ if necessary, every function in the weighted variation unit ball admits an approximant with at most $n$ atoms, weighted coefficient sum at most one, and $L^q(\mu_K)$ error bounded by the right-hand side of \eqref{geo:eq-profile-upper}.

\item[(3)] \emph{The endpoint $\eta=1/q$.}
In this case, $\lambda_B=1$ for every $B\in\mathcal B$.  Under \eqref{geo:eq-hardy}, the estimate \eqref{geo:eq-profile-upper} gives
\begin{equation}\label{geo:eq-hardy-endpoint}
 \mathcal E_n^K(h\theta^{1/q};q)
 \le CH(n+1)^{-\min\{\alpha,b_q\}}.
\end{equation}
\end{enumerate}
\end{proposition}

\begin{proof}
By \eqref{eq:domain-error} and Lemma~\ref{lem:closure}(1), it suffices to prove a uniform bound for finite networks 
$F=\sum_i c_i\phi_{z_i}$ with
\[
 \sum_{B\in\mathcal B}M_B
 =\sum_i|c_i|w_{K,\eta}(z_i)\le1.
\]
Only finitely many groups have $M_B>0$.

Let $I\subset\mathcal B$ be a finite collection of groups with $M_B>0$,
and let $N\ge1$ be an integer.
We estimate the error of approximating $\sum_{B\in I}F_B$
using at most $N$ neurons. Fix $0<\gamma\le\alpha$.
For any nonnegative integers $(m_B)_{B\in I}$,
\eqref{geo:eq-block} gives
\[
 |F_B-S_B|
 \le C\lambda_BM_B(m_B+1)^{-\alpha}\psi_B
 \le C\lambda_BM_B(m_B+1)^{-\gamma}\psi_B,
 \qquad B\in I.
\]
Here we used $m_B+1\ge1$ and $\gamma\le\alpha$.

Applying Lemma~\ref{lem:allocation} to
$(\lambda_BM_B)_{B\in I}$, choose nonnegative integers $m_B$
with $\sum_{B\in I}m_B\le N$.
For these integers, take the local approximants $S_B$ from
\eqref{geo:eq-block} and set $S_I=\sum_{B\in I}S_B$.
Using \eqref{geo:eq-hardy} with $u_B=\lambda_BM_B(m_B+1)^{-\gamma}$ and then \eqref{core:eq-allocation}, we obtain
\begin{equation}\label{geo:eq-allocation-r}
 \begin{aligned}
 \left\|\sum_{B\in I}F_B-S_I\right\|_{L^q(\mu_K)}
 &\le CH
 \left(\sum_{B\in I}
   \bigl[\lambda_BM_B(m_B+1)^{-\gamma}\bigr]^q
 \right)^{1/q}\\
 &\le CHN^{-\gamma}
 \left(\sum_{B\in I}(\lambda_BM_B)^r\right)^{1/r},
 \end{aligned}
\end{equation}
where $r=(\gamma+1/q)^{-1}$.
Since each $S_B$ uses at most $m_B$ original atoms,
$S_I$ uses at most $\sum_{B\in I}m_B\le N$ atoms appearing in $F$. The allocation lemma allows $m_B=0$. For such a group we take $S_B=0$, and the error is still bounded by \eqref{geo:eq-block}.

We next estimate $\left(\sum_{B\in I}(\lambda_BM_B)^r\right)^{1/r}$
in terms of the weighted coefficient sum $\sum_{B\in I}M_B$.
Set $a=\gamma-b_q=1/r-1$.
If $a>0$, then $0<r<1$, and H\"older's inequality with exponents
$1/r$ and $1/(1-r)$ gives
\begin{equation}\label{geo:eq-mass-holder}
 \left(\sum_{B\in I}(\lambda_BM_B)^r\right)^{1/r}
 \le \left(\sum_{B\in I}M_B\right)
       \left(\sum_{B\in I}\lambda_B^{1/a}\right)^a.
\end{equation}
If $a\le0$, then $r\ge1$, so
\[
 \left(\sum_{B\in I}(\lambda_BM_B)^r\right)^{1/r}
 \le \sum_{B\in I}\lambda_BM_B
 \le \left(\sup_{B\in\mathcal B}\lambda_B\right)
       \sum_{B\in I}M_B.
\]

When $a_q\le0$ and $n\ge1$, apply
\eqref{geo:eq-allocation-r} with $\gamma=\alpha$, $N=n$,
and $I=\{B\in\mathcal B:M_B>0\}$.
Then $a=a_q\le0$ and $\sum_{B\in I}M_B\le1$, so the preceding
inequality gives
\[
 \|F-S_I\|_{L^q(\mu_K)}
 \le CHn^{-\alpha}\sup_{B\in\mathcal B}\lambda_B.
\]
This proves \eqref{geo:eq-profile-upper} in this case,
after adjusting $C$ to replace $n$ by $n+1$.

Now suppose $a_q>0$ and $n\ge2$.
List the groups with $M_B>0$ as $B_1,\ldots,B_J$ so that
\[
 \lambda_{B_1}\ge\cdots\ge\lambda_{B_J}.
\]
Since $B_1,\ldots,B_J$ form a subcollection of $\mathcal B$,
the definition of $\lambda_k^*$ gives
$\lambda_{B_k}\le\lambda_k^*$ for $1\le k\le J$.

Apply \eqref{geo:eq-allocation-r} and
\eqref{geo:eq-mass-holder} with
$I=\{B_1,\ldots,B_{\min\{n,J\}}\}$,
$\gamma=\alpha$, and $N=\lfloor n/2\rfloor$.
The resulting approximant $S_I$ uses at most
$\lfloor n/2\rfloor$ atoms appearing in $F$.
Since $\sum_{B\in I}M_B\le1$,
$\lambda_{B_k}\le\lambda_k^*$, and
$\lfloor n/2\rfloor\ge n/3$, we obtain
\[
 \left\|\sum_{B\in I}F_B-S_I\right\|_{L^q(\mu_K)}
 \le CHn^{-\alpha}
 \left(\sum_{k=1}^{n}(\lambda_k^*)^{1/a_q}\right)^{a_q}.
\]

Let $I'=\{B_k:n<k\le J\}$ denote the remaining groups.
Every $B\in I'$ satisfies $\lambda_B\le\lambda_{n+1}^*$.

If $q>1$, then $a_q>0$ implies $0<b_q<\alpha$.
Apply \eqref{geo:eq-allocation-r} to $I'$ with
$\gamma=b_q$ and $N=n-\lfloor n/2\rfloor$.
Since $r=(b_q+1/q)^{-1}=1$, the resulting approximant
$S_{I'}$ uses at most $n-\lfloor n/2\rfloor$ atoms appearing in $F$ and satisfies
\[
 \begin{aligned}
 \left\|\sum_{B\in I'}F_B-S_{I'}\right\|_{L^q(\mu_K)}
 &\le CHn^{-b_q}\sum_{B\in I'}\lambda_BM_B\\
 &\le CHn^{-b_q}\lambda_{n+1}^*
       \sum_{B\in I'}M_B\\
 &\le CHn^{-b_q}\lambda_{n+1}^*,
 \end{aligned}
\]
where we used $N\ge n/2$ and $\sum_{B\in I'}M_B\le1$. 

If $q=1$, then $b_q=0$. Take $S_{I'}=0$.
Using $\|\phi_z\|_{L^1(\mu_K)}\le h_z\theta_z$ and
$\theta_z\asymp\mu_K(E_B)$ for $z\in B$ by
\eqref{geo:eq-atlas}, we obtain
\[
 \|F_B\|_{L^1(\mu_K)}
 \le \sum_{i:z_i\in B}|c_i|w_{K,\eta}(z_i)
                  \theta_{z_i}^{1-\eta}
 \le C\mu_K(E_B)^{1-\eta}M_B
 =C\lambda_BM_B,
\]
where $\lambda_B=\mu_K(E_B)^{1-\eta}$ since $q=1$. Thus
\[
 \left\|\sum_{B\in I'}F_B\right\|_{L^1(\mu_K)}
 \le C\sum_{B\in I'}\lambda_BM_B
 \le C\lambda_{n+1}^*.
\]

Since $\lambda_k^*\ge\lambda_{n+1}^*$ for $k\le n$
and $\alpha-a_q=b_q$,
\[
 n^{-b_q}\lambda_{n+1}^*
 \le n^{-\alpha}
 \left(\sum_{k=1}^{n}(\lambda_k^*)^{1/a_q}\right)^{a_q}.
\]
The approximant $S=S_I+S_{I'}$ uses at most $n$ original atoms.
The triangle inequality therefore gives
\[
 \|F-S\|_{L^q(\mu_K)}
 \le CHn^{-\alpha}
 \left(\sum_{k=1}^{n}(\lambda_k^*)^{1/a_q}\right)^{a_q}.
\]
Using $n^{-\alpha}\le2^\alpha(n+1)^{-\alpha}$ yields \eqref{geo:eq-profile-upper},
after adjusting $C$. For $n=0,1$, take $S=0$.
By the triangle inequality, \eqref{geo:eq-atlas}, and
$\sum_i|c_i|w_{K,\eta}(z_i)\le1$, we have $\|F\|_{L^q(\mu_K)}
 \le \sum_i|c_i|h_{z_i}\theta_{z_i}^{1/q}
 \le C\lambda_1^*$. This gives \eqref{geo:eq-profile-upper} after adjusting $C$.

For a finite network $F$, the approximant uses only atoms
appearing in $F$ and does not increase the weighted coefficient sum.
Lemma~\ref{lem:closure}(2) gives an approximant for every
function in the weighted variation unit ball, using at most $n$
atoms, with weighted coefficient sum at most one and error at most
twice the right-hand side of \eqref{geo:eq-profile-upper}.

Finally, if $\eta=1/q$, then $\lambda_B=1$ for every $B$.
Using $a_q=\alpha-b_q$, \eqref{geo:eq-profile-upper} gives
\[
 \mathcal E_n^K(h\theta^{1/q};q)
 \le CH(n+1)^{-\alpha+\max\{a_q,0\}}
 =CH(n+1)^{-\min\{\alpha,b_q\}},
\]
which proves \eqref{geo:eq-hardy-endpoint}.
\end{proof}
For $1/q>\eta$, the next proposition gives upper bounds
in terms of the growth of $N_j$ and $S_j(q)$.
We use \eqref{geo:layer} within each volume layer and combine
the resulting approximation errors by the triangle inequality.

\begin{proposition}
\label{geo:layer-upper}
Consider the grouping by cap volumes constructed before
\eqref{geo:layer}, with $\tau_j=\tau_0 2^{-j}$ for $j\ge0$. 
Let $N_j$ be the number of groups for
$\tau_j/2<\theta_z\le\tau_j$, and let $S_j(q)$ be defined
by \eqref{geo:layer}.
Assume $\varepsilon=1/q-\eta>0$.

Suppose first that $N_j\le C_0 2^{\kappa j}$ and $S_j(q)\le S$
for all $j\ge0$, where $C_0,S>0$ and $\kappa>0$.
Then, for every integer $n\ge0$,
\begin{equation}\label{geo:eq-layer-rates}
 \mathcal E_n^K(w_{K,\eta};q)\le C
 \begin{cases}
 (n+1)^{-(b_q+\varepsilon/\kappa)},
       &a_q>0,\ 0<\varepsilon<\kappa a_q,\\
 (n+1)^{-\alpha}\ell_n^\alpha,
       &a_q>0,\ \varepsilon=\kappa a_q,\\
 (n+1)^{-\alpha},
       &\varepsilon>\kappa\max\{a_q,0\},
 \end{cases}
\end{equation}
where $\ell_n=1+\log(n+1)$. If \eqref{geo:eq-hardy} also holds for $(E_B)_{B\in\mathcal B}$,
the second case of \eqref{geo:eq-layer-rates} holds with
$\ell_n^{a_q}$ in place of $\ell_n^\alpha$.

More generally, suppose that, for some $0\le\kappa<\infty$,
\begin{equation}\label{geo:eq-subpower}
 \limsup_{j\to\infty}\frac{\log N_j}{j\log2}\le\kappa,
 \qquad
 \lim_{j\to\infty}\frac{\log S_j(q)}{j}=0.
\end{equation}
Then, for every $\zeta>0$ and every integer $n\ge0$,
\begin{equation}\label{geo:eq-subpower-upper}
 \mathcal E_n^K(w_{K,\eta};q)
 \le C_\zeta(n+1)^{-r+\zeta},
 \qquad
 r=\min\{\alpha,b_q+\varepsilon/\kappa\},
\end{equation}
with the convention $\varepsilon/0=+\infty$.
If $\kappa=0$ or
$\varepsilon>\kappa\max\{a_q,0\}$, the stronger estimate
$\mathcal E_n^K(w_{K,\eta};q)\le C(n+1)^{-\alpha}$ holds.

For every function in the weighted variation unit ball,
these bounds hold with approximants using at most $n$ atoms
and weighted coefficient sum at most one.
For a finite network with weighted coefficient sum at most one,
the approximant uses only atoms appearing in that network
and does not increase the weighted coefficient sum.
The constants are independent of $n$ and of the function
being approximated.
\end{proposition}

\begin{proof}
By \eqref{eq:domain-error} and Lemma~\ref{lem:closure}, it suffices to consider a finite sum $F=\sum_i c_i\phi_{z_i}$ satisfying $\sum_i|c_i|w_{K,\eta}(z_i)\le1$. Let $F_j$ be the sum of its terms with $\tau_{j+1}<\theta_{z_i}\le\tau_j$, and let $M_j$ be their weighted sum of absolute coefficients. Only finitely many $M_j$ are nonzero.

Let $\mathcal B_j$ be the $N_j$ groups in layer $j$.
Then $F_j=\sum_{B\in\mathcal B_j}F_B$ and
$M_j=\sum_{B\in\mathcal B_j}M_B$.
By \eqref{geo:eq-atlas},
$\lambda_B\asymp\tau_j^\varepsilon$ for $B\in\mathcal B_j$.
Within this layer, \eqref{geo:layer} allows the argument for
\eqref{geo:eq-allocation-r} to be applied with $H=S_j(q)$.

For $m\ge1$, take $N=m$ and $\gamma=\alpha$ in
\eqref{geo:eq-allocation-r}.
The exponent $r$ in that estimate then satisfies $1/r-1=a_q$.

If $a_q>0$, apply \eqref{geo:eq-mass-holder}.
If $a_q\le0$, then $r\ge1$ and
\[
 \left(\sum_{B\in\mathcal B_j}(\lambda_BM_B)^r\right)^{1/r}
 \le \sum_{B\in\mathcal B_j}\lambda_BM_B
 \le C\tau_j^\varepsilon M_j.
\]
Using $\#\mathcal B_j=N_j$ and
$\lambda_B\asymp\tau_j^\varepsilon$, we obtain in either case,
after adjusting $C$, an approximant $S_j^{(m)}$ to $F_j$
satisfying
\begin{equation}\label{geo:eq-layer-fast}
 \|F_j-S_j^{(m)}\|_q
 \le C S_j(q)\tau_j^\varepsilon
      N_j^{\max\{a_q,0\}}M_j(m+1)^{-\alpha}.
\end{equation}

If $q>1$ and $a_q>0$, take $\gamma=b_q\in(0,\alpha)$
in \eqref{geo:eq-allocation-r}, so that $r=1$.
Since $\sum_{B\in\mathcal B_j}\lambda_BM_B
\le C\tau_j^\varepsilon M_j$, we obtain an approximant $T_j^{(m)}$ to $F_j$ satisfying
\begin{equation}\label{geo:eq-layer-base}
 \|F_j-T_j^{(m)}\|_q
 \le C S_j(q)\tau_j^\varepsilon M_j(m+1)^{-b_q}.
\end{equation}
Both $S_j^{(m)}$ and $T_j^{(m)}$ use at most $m$ atoms
appearing in $F_j$ and have weighted coefficient sum
at most $M_j$. For $m=0$, take zero approximants.
Since $\|\phi_z\|_q\le h_z\theta_z^{1/q}
=w_{K,\eta}(z)\theta_z^\varepsilon$ and
$\theta_z\le\tau_j$ in this layer, the triangle inequality gives
$\|F_j\|_q\le\tau_j^\varepsilon
\sum_{B\in\mathcal B_j}M_B=\tau_j^\varepsilon M_j$.
Thus both estimates remain valid for $m=0$.

Suppose the error in layer $j$ using at most $m_j$ atoms
is bounded by $D_jM_j(m_j+1)^{-\gamma}$, where
$\gamma>0$ and $D_j\ge0$.
If $0<\sum_jD_j^{1/\gamma}<\infty$, fix an integer $N\ge1$
and set
\[
 m_j=\left\lfloor
 \frac{ND_j^{1/\gamma}}{\sum_kD_k^{1/\gamma}}
 \right\rfloor.
\]
Then $\sum_jm_j\le N$, and the definition of $m_j$ gives
\begin{equation} \label{geo:eq-layer-allocation}
    \sum_jD_jM_j(m_j+1)^{-\gamma}
 \le N^{-\gamma}
 \left(\sum_jD_j^{1/\gamma}\right)^\gamma
 \sum_jM_j.
\end{equation}
Only finitely many $m_j$ are positive, and their choice
is independent of $M_j$.
If all $D_j=0$, take $m_j=0$ for every $j$.

Recall that $\tau_j=\tau_0 2^{-j}$ for $j\ge0$. Assume $N_j\le C_0 2^{\kappa j}$ and $S_j(q)\le S$.
First let $a_q>0$ and $n\ge2$, and choose $J\ge0$ so that
$2^{\kappa J}\le n+1<2^{\kappa(J+1)}$.

Use \eqref{geo:eq-layer-fast} for $0\le j\le J$ and
\eqref{geo:eq-block} for the group with $\theta_z>\tau_0$,
whose containing set is $K$ and hence has $\lambda_B=1$.
Applying \eqref{geo:eq-layer-allocation} with $\gamma=\alpha$
and $N=\lfloor n/2\rfloor$, and using the fact that their
total weighted coefficient sum is at most one, gives a
combined error at most
\begin{equation}\label{geo:eq-layer-head}
 C(n+1)^{-\alpha}
 \left(1+\sum_{j=0}^J
       2^{j(\kappa a_q-\varepsilon)/\alpha}\right)^\alpha.
\end{equation}
The term $1$ accounts for the group with $\theta_z>\tau_0$.

For $q>1$, combine \eqref{geo:eq-layer-base} over $j>J$
using \eqref{geo:eq-layer-allocation} with $\gamma=b_q$
and $N=n-\lfloor n/2\rfloor$.
The resulting error is at most
\begin{equation}\label{geo:eq-layer-tail}
 C(n+1)^{-b_q}
 \left(\sum_{j>J}2^{-j\varepsilon/b_q}\right)^{b_q}
 \le C(n+1)^{-b_q}2^{-J\varepsilon}.
\end{equation}
For $q=1$, approximate layers with $j>J$ by zero.
The bounds $\|F_j\|_1\le\tau_j^\varepsilon M_j$ and
$\sum_{j>J}M_j\le1$ give a combined error at most
$C2^{-J\varepsilon}$.

The two approximants together use at most $n$ atoms.
For $0<\varepsilon<\kappa a_q$, summing the geometric series
and adding the two errors gives the first case of
\eqref{geo:eq-layer-rates}, since
$2^{\kappa J}\asymp n+1$ and $\alpha-a_q=b_q$.
For $\varepsilon=\kappa a_q$, \eqref{geo:eq-layer-head}
is bounded by
$C(n+1)^{-\alpha}(J+2)^\alpha
 \le C(n+1)^{-\alpha}\ell_n^\alpha$,
which gives the second case.

For the third case of \eqref{geo:eq-layer-rates},
assume $\varepsilon>\kappa\max\{a_q,0\}$.
Using $N_j\le C_0 2^{\kappa j}$, $S_j(q)\le S$, and
$\tau_j=\tau_0 2^{-j}$, we have
\[
 \sum_{j=0}^\infty
 \left(S_j(q)\tau_j^\varepsilon
       N_j^{\max\{a_q,0\}}\right)^{1/\alpha}
 \le C\sum_{j=0}^\infty
 2^{j(\kappa\max\{a_q,0\}-\varepsilon)/\alpha}
 <\infty.
\]
Apply \eqref{geo:eq-layer-allocation} with $\gamma=\alpha$
and $N=n$ to the bounds in \eqref{geo:eq-layer-fast}
for all layers and \eqref{geo:eq-block} for the group
with $\theta_z>\tau_0$, obtaining $C(n+1)^{-\alpha}$.

At the critical value $\varepsilon=\kappa a_q>0$, suppose
in addition that $(E_B)_{B\in\mathcal B}$ satisfies
\eqref{geo:eq-hardy}.
The bounds $1+\sum_{j=0}^J N_j\le C2^{\kappa J}$ and
$\lambda_B\asymp2^{-j\kappa a_q}$ for $B\in\mathcal B_j$
imply $\lambda_k^*\le Ck^{-a_q}$.
Hence
$\sum_{k=1}^{n+1}(\lambda_k^*)^{1/a_q}
\le C\sum_{k=1}^{n+1}k^{-1}\le C\ell_n$,
and \eqref{geo:eq-profile-upper} gives the improved bound
$C(n+1)^{-\alpha}\ell_n^{a_q}$.

Finally, \eqref{geo:eq-subpower} gives
$N_j\le C_\delta2^{(\kappa+\delta)j}$ and
$S_j(q)\tau_j^\varepsilon
\le C_\delta2^{-j(\varepsilon-\delta)}$
for every $0<\delta<\varepsilon$.
Given $\zeta>0$, choose $\delta$ so that
\[
 \min\left\{\alpha,b_q+
 \frac{\varepsilon-\delta}{\kappa+\delta}\right\}
 >r-\frac{\zeta}{2},
\]
where $r$ is defined in \eqref{geo:eq-subpower-upper}. Repeat the derivation of \eqref{geo:eq-layer-rates} with
$(\varepsilon,\kappa)$ replaced by
$(\varepsilon-\delta,\kappa+\delta)$, choosing $J$ so that
$2^{(\kappa+\delta)J}\asymp n+1$ in the first two cases.
This gives, in all three cases, an error at most
\[
 C_\delta(n+1)^{
 -\min\{\alpha,b_q+(\varepsilon-\delta)/(\kappa+\delta)\}}
 \ell_n^\alpha.
\]
The choice of $\delta$ and
$\ell_n^\alpha\le C_\zeta(n+1)^{\zeta/2}$
then prove \eqref{geo:eq-subpower-upper}. If $\kappa=0$ or $\varepsilon>\kappa\max\{a_q,0\}$,
choose $\delta>0$ with
$\varepsilon-\delta>(\kappa+\delta)\max\{a_q,0\}$.
The argument for the third case of \eqref{geo:eq-layer-rates}
then gives $C(n+1)^{-\alpha}$. The cases $n=0,1$ follow by taking the zero approximant
and enlarging the constant, since $\|F\|_q\le1$.

Lemma~\ref{lem:closure}(2) extends the constructions to
the weighted variation unit ball, after adjusting the constants.
\end{proof}

\subsection{Approximation Rates for Positively Curved Bodies}
\label{curvednew:section}

We prove Theorem~\ref{main:curved}, which gives upper bounds
and the optimal approximation exponent for convex bodies
with $C^2$ boundary and strictly positive principal curvatures.
We construct cap coverings to apply
Proposition~\ref{geo:hardy-upper}, and combine the resulting
upper bounds with the lower bounds in
Proposition~\ref{geo:packing-lower} and
Theorem~\ref{geo:interior-lower}. 
We begin with a lemma showing that caps with uniformly
bounded overlap within each volume layer satisfy
\eqref{geo:eq-hardy}.

\begin{lemma}
\label{curvednew:depth-hardy}
Let $K\subset\mathbb R^d$ be a convex body.
For each $j\ge0$, let $(E_{ji})_i$ be a finite family of caps
of $K$ satisfying
\[
 a2^{-j}\le \mu_K(E_{ji})\le A2^{-j},
 \qquad
 \sum_i\mathbf1_{E_{ji}}\le L
 \quad\text{almost everywhere on }K,
\]
where $a,A,L>0$ are independent of $j$ and $i$.
Then, for every $1\le q<\infty$, every scalar $c_*$,
and every finitely supported coefficient family $(c_{ji})$,
\[
 \left\|
 c_*\mathbf1_K+
 \sum_{j,i}c_{ji}\mu_K(E_{ji})^{-1/q}\mathbf1_{E_{ji}}
 \right\|_{L^q(\mu_K)}
 \le C_{d,q,a,A,L}
 \left(|c_*|^q+\sum_{j,i}|c_{ji}|^q\right)^{1/q}.
\]
\end{lemma}
\begin{proof}
For $q=1$, the triangle inequality gives the result with
constant one. For $q>1$, we verify the weighted overlap condition
\eqref{geom:eq-weighted-overlap} in Lemma~\ref{geom:depth-synthesis}.

By Definition~\ref{def:wet-part-depth}, every cap $E$ of $K$ satisfies
\begin{equation}\label{curvednew:depth-comparison}
 v_K(x)\le\mu_K(E), \qquad\forall x\in E.
\end{equation}
Thus $x\in E_{ji}$ implies $v_K(x)\le A2^{-j}$.
For almost every $x\in K$, $v_K(x)>0$ and at most $L$
caps in each layer contain $x$. Using
$\mu_K(E_{ji})\ge a2^{-j}$, we obtain, for every $\gamma>0$, 
\[
 \begin{aligned}
 \sum_{j,i}
 \left(\frac{v_K(x)}{\mu_K(E_{ji})}\right)^\gamma
 \mathbf1_{E_{ji}}(x)
 &\le La^{-\gamma}
 \sum_{\substack{j\ge0\\v_K(x)2^j\le A}}
       \bigl(v_K(x)2^j\bigr)^\gamma\\
 &\le La^{-\gamma}\sum_{\ell=0}^{\infty}(A2^{-\ell})^{\gamma}
 =\frac{L(A/a)^\gamma}{1-2^{-\gamma}}.
 \end{aligned}
\]

The preceding sum is bounded for every $\gamma>0$. For any $1<q<\infty$, choose
\[
 \gamma=\frac{\beta_d}{2(q-1)},
 \qquad
 \gamma(q-1)=\frac{\beta_d}{2}<\beta_d.
\]
Since $\mu_K(K)=1$, adding $K$ to the family adds $v_K(x)^\gamma\le1$ to the sum in \eqref{geom:eq-weighted-overlap}. Lemma~\ref{geom:depth-synthesis}, applied with $s=q$ to the sets $E_{ji}$ together with $K$, now gives the required bound.
\end{proof}

\begin{proposition}
\label{curvednew:atlas}
Let $K\subset\mathbb R^d$ be a convex body with $C^2$
boundary and strictly positive principal curvatures.
Set $\tau_j=2^{-j}$ and recall that
$\kappa_d=(d-1)/(d+1)$. 
There exists an integer $j_0\ge0$ such that, for every
$j\ge j_0$, one can choose caps $E_{j1},\ldots,E_{jN_j}$
of $K$ so that every cap $C_z$ with
$\tau_j/2<\theta_z\le\tau_j$ is contained in some $E_{ji}$.
These caps satisfy
\begin{equation}\label{curvednew:count}
 N_j\asymp_K\tau_j^{-\kappa_d},\qquad
 \mu_K(E_{ji})\asymp_K\tau_j.
\end{equation}
Their overlap satisfies
$\sum_{i=1}^{N_j}\mathbf1_{E_{ji}}\le C_K$
almost everywhere on $K$, uniformly in $j\ge j_0$.

Caps with $\theta_z>\tau_{j_0}$ may be assigned to one
additional group with containing set $K$.
For every $1\le q<\infty$, every scalar $c_*$, and every
finitely supported coefficient family $(c_{ji})$,
\begin{equation}\label{curvednew:hs}
 \left\|
 c_*\mathbf1_K+
 \sum_{j,i}c_{ji}\mu_K(E_{ji})^{-1/q}\mathbf1_{E_{ji}}
 \right\|_{L^q(\mu_K)}
 \le C_{K,q}
 \left(|c_*|^q+\sum_{j,i}|c_{ji}|^q\right)^{1/q}.
\end{equation}
Moreover, the cap packing number $P_K(\tau)$ defined in
\eqref{geo:packing} satisfies
\begin{equation}\label{curvednew:packing}
 P_K(\tau)\asymp_K\tau^{-\kappa_d},
 \qquad\text{as }\tau\downarrow0.
\end{equation}
\end{proposition}

\begin{proof}
Let $\nu(p)$ be the outward unit normal at $p\in\partial K$.
The principal curvatures are continuous and positive, so
compactness of $\partial K$ gives a uniform positive lower
bound and a finite upper bound.
Thus we may apply the tangent graph argument in the proof of
Lemma~\ref{geo:c2-packing}, taking the compact boundary subset
used there to be $\partial K$. 
For the caps $C(p,h)$ defined in \eqref{geo:eq-cap-height},
there exists $h_0>0$, depending only on $K$, such that
\begin{equation}\label{curvednew:quadratic}
 C(p,h)\subset B(p,C_K\sqrt h),\qquad
 \mu_K(C(p,h))\asymp_K h^{(d+1)/2},
\end{equation}
for every $p\in\partial K$ and $0<h<h_0$.
By $C^2$ regularity and compactness, there exists
$r_0>0$, depending only on $K$, such that
\begin{equation}\label{curvednew:normal-lipschitz}
 |\nu(p')-\nu(p)|\le C_K|p'-p|,
\end{equation}
for all $p,p'\in\partial K$ with $|p'-p|<r_0$.
For these points, the quadratic estimate
\eqref{geo:eq-quadratic}, applied in tangent coordinates
at $p'$, also gives
\begin{equation}\label{curvednew:support-deficit}
 0\le\nu(p')\cdot(p'-p)\le C_K|p'-p|^2.
\end{equation}

Set $h_j=\tau_j^{2/(d+1)}$ and choose a maximal
$\sqrt{h_j}$-separated set
$\{p_{ji}\}_{i=1}^{N_j}\subset\partial K$.
Maximality gives
$\partial K\subset\bigcup_i B(p_{ji},\sqrt{h_j})$,
while the balls $B(p_{ji},\sqrt{h_j}/2)$ are disjoint.
For sufficiently small $r>0$, the surface area of
$\partial K\cap B(p,r)$ is comparable to $r^{d-1}$,
uniformly in $p\in\partial K$.
The covering gives the lower bound for $N_j$, and
the disjointness gives the upper bound. Thus
$N_j\asymp_K h_j^{-(d-1)/2}=\tau_j^{-\kappa_d}$.

For sufficiently large $j$, consider any cap $C(p,h)$
in layer $j$, that is,
$\tau_j/2<\mu_K(C(p,h))\le\tau_j$.
By \eqref{curvednew:quadratic}, its height satisfies
$h\asymp_K h_j$.
Assign this cap to a point $p_{ji}$ with
$|p-p_{ji}|\le\sqrt{h_j}$, which exists by maximality. For $x\in C(p,h)$, applying
\eqref{curvednew:support-deficit} and
\eqref{curvednew:normal-lipschitz} with $p'=p_{ji}$,
and using \eqref{curvednew:quadratic}, we obtain
\[
 \begin{aligned}
 \nu(p_{ji})\cdot(p_{ji}-x)
 &=\nu(p_{ji})\cdot(p_{ji}-p)
   +(\nu(p_{ji})-\nu(p))\cdot(p-x)\\
 &\qquad+\nu(p)\cdot(p-x)\\
 &\le C_K|p_{ji}-p|^2
   +C_K|p_{ji}-p|\,|p-x|+h\\
 &\le C_Kh_j.
 \end{aligned}
\]
Thus, for a sufficiently large fixed $A_K\geq C_K$, the caps
$E_{ji}=C(p_{ji},A_Kh_j)$ contain their assigned caps.
Taking $j_0$ sufficiently large, \eqref{curvednew:quadratic}
also gives
$\mu_K(E_{ji})\asymp_K\tau_j$ and
$E_{ji}\subset B(p_{ji},C_K\sqrt{h_j})$.
This proves \eqref{curvednew:count}.

For any $x\in K$, the balls $B(p_{ji},\sqrt{h_j}/2)$
with $x\in E_{ji}$ are disjoint and lie in
$B(x,(C_K+1/2)\sqrt{h_j})$.
The number of these disjoint balls equals
$\sum_{i=1}^{N_j}\mathbf1_{E_{ji}}(x)$.
Comparing their total volume with that of the containing
ball gives
\[
 \sum_{i=1}^{N_j}\mathbf1_{E_{ji}}(x)
 \le
 \left(
 \frac{(C_K+1/2)\sqrt{h_j}}{\sqrt{h_j}/2}
 \right)^d
 =(2C_K+1)^d.
\]
This bound is independent of $x$ and $j$. Lemma~\ref{curvednew:depth-hardy}, applied to these caps
together with $K$, now proves \eqref{curvednew:hs}.

Finally, \eqref{curvednew:packing} follows from
Lemma~\ref{geo:c2-packing}.
\end{proof}

\begin{proof}[Proof of Theorem~\ref{main:curved}]
Set $\kappa_d=(d-1)/(d+1)$ and recall that
$\varepsilon=1/q-\eta\ge0$.
Take the groups from Proposition~\ref{curvednew:atlas},
with containing sets $E_{ji}$ in layer $j$ and $K$
for the additional group.
By \eqref{curvednew:hs}, these sets satisfy
\eqref{geo:eq-hardy} with $H=C_{K,q}$.

For $j\ge j_0$, \eqref{curvednew:count} gives
$1+\sum_{\ell=j_0}^jN_\ell\le C_K2^{\kappa_d j}$,
while
$\lambda_B=\mu_K(E_B)^\varepsilon
\le C_{K,q,\eta}2^{-j\varepsilon}$
for groups $B$ in layer $j$.
Thus at most $C_K2^{\kappa_dj}$ groups $B$ satisfy
$\lambda_B>C_{K,q,\eta}2^{-j\varepsilon}$.
For sufficiently large $k$, choose $j\ge j_0$ such that
$C_K2^{\kappa_dj}\le k-1<C_K2^{\kappa_d(j+1)}$.
By \eqref{geo:eq-rearrangement},
\[
 \lambda_k^*
 \le C_{K,q,\eta}2^{-j\varepsilon}
 \le C_{K,q,\eta}k^{-\varepsilon/\kappa_d}.
\]
After enlarging the constant to cover the finitely many
smaller $k$, the bound holds for every $k\ge1$.

If $a_q>0$, Proposition~\ref{geo:hardy-upper} therefore gives
\[
 \mathcal E_n^K(w_{K,\eta};q)
 \le C_{K,q,\eta}(n+1)^{-\alpha}
 \left(
  \sum_{k=1}^{n+1}
       k^{-\varepsilon/(\kappa_d a_q)}
 \right)^{a_q}.
\]
Therefore, we have
\[
 \sum_{k=1}^{n+1}k^{-\varepsilon/(\kappa_d a_q)}
 \le C
 \begin{cases}
  1,&\varepsilon>\kappa_d a_q,\\
  \ell_n,&\varepsilon=\kappa_d a_q,\\
  (n+1)^{1-\varepsilon/(\kappa_d a_q)},
      &0\le\varepsilon<\kappa_d a_q,
 \end{cases}
\]
where $\ell_n=1+\log(n+1)$. Substituting these bounds into the preceding error estimate
and using $\alpha-a_q=b_q$ proves
\eqref{main:eq-curved-rates}.

When $a_q\le0$, \eqref{geo:eq-profile-upper}
gives \eqref{main:eq-curved-classical}.
Setting $\varepsilon=0$ in the cases $a_q>0$ and
$a_q\le0$ gives the stated bound at $\eta=1/q$.

The conclusion for finite networks in
Proposition~\ref{geo:hardy-upper} gives
\eqref{main:eq-compression} for these upper bounds,
with the factor $M$ obtained by rescaling \(F\).

For the lower bound, take
\[
 m=\left\lceil C_d(n+1)\log(n+2)\right\rceil,
 \qquad
 \tau=c_Km^{-1/\kappa_d},
\]
where $C_d$ is the constant in
Proposition~\ref{geo:packing-lower}.
Choose $c_K>0$ sufficiently small that
\eqref{curvednew:packing} gives $P_K(\tau)\ge m$.
Proposition~\ref{geo:packing-lower} then yields
\[
 \mathcal E_n^K(w_{K,\eta};q)
 \ge c_{d,q}m^{-b_q}\tau^\varepsilon
 =c_{K,q,\eta}m^{-b_q-\varepsilon/\kappa_d}.
\]
Since $m\asymp_d(n+1)\log(n+2)$, this estimate and
\eqref{geo:eq-interior-lower} give
\[
 \mathcal E_n^K(w_{K,\eta};q)
 \ge c_{K,q,\eta}
 \bigl((n+1)\log(n+2)\bigr)^{
 -\min\{\alpha,b_q+\varepsilon/\kappa_d\}}.
\]
Comparing this with the upper bounds proves
\eqref{main:eq-c2-exponent} and
\eqref{main:eq-c2-log-limit}, since
$\log\log(n+2)/\log n\to0$.
The argument also covers $\varepsilon=0$.
\end{proof}

\subsection{Planar Geometry and Products with Polytopes}
\label{intr:section}

We prove Theorem~\ref{main:intrinsic} for planar convex
bodies and invertible affine images of their products
with polytopes, without any boundary regularity assumption
on the planar body.
We construct containing sets for caps and bound their
number and overlap at each volume scale.
We also compare $\mathcal C_K$ with $P_K$, up to powers
of a logarithm.
These estimates allow us to apply
Propositions~\ref{geo:layer-upper} and~\ref{geo:packing-lower}.

Recall from \eqref{intr:eq-profile} that
$\mathcal C_K(\tau)=W_K(\tau)/\tau$, where
$W_K(\tau)=\mu_K\{x:v_K(x)\le\tau\}$.
The packing number $P_K(\tau)$, defined in
\eqref{geo:packing}, is the maximum number of caps
of relative volume $\tau$ with pairwise disjoint interiors.
Write $L(\tau)=1+\log(1/\tau)$ for $0<\tau<1$.

\subsubsection{Cap Coverings for Planar Convex Bodies}

\begin{lemma}
\label{intr:arc-facts}
Let $D$ be a planar convex body.
For every cap $C$ of $D$ with $0<\mu_D(C)<1$,
the set $\Gamma=C\cap\partial D$ is a closed boundary arc,
and $C=\operatorname{conv}\Gamma$.
Thus inclusion of such arcs implies inclusion of the
corresponding caps.
If two such arcs are disjoint except possibly at their
endpoints, the corresponding caps have disjoint interiors.

For a finite family of such caps, if each point of
$\partial D$ belongs to at most $L$ of their boundary arcs,
then each point of $\operatorname{int}D$ belongs to
at most $2L$ caps.
\end{lemma}

\begin{proof}
The boundary of $C$ consists of $\Gamma$ and the chord
joining its endpoints. Hence $C=\operatorname{conv}\Gamma$,
which also proves the inclusion statement.
If two such arcs are disjoint except possibly at their
endpoints, the cutting line of one cap separates the arcs
into opposite closed half-planes. Their convex hulls
therefore have disjoint interiors.

For the last assertion, fix $x\in\operatorname{int}D$
and let $k$ be the number of caps containing $x$.
For each such cap $C=D\cap\{y:u\cdot y\ge t\}$,
the ray $x+r\omega$, $r\ge0$, exits $D$ through $\Gamma$
whenever $\omega$ is a unit vector with $u\cdot\omega\ge0$.
These directions form a semicircle of angular length $\pi$.
Let $\Gamma_1,\ldots,\Gamma_k$ be the arcs of the caps
containing $x$, and let $y(\omega)$ denote the exit point
of the ray $x+r\omega$ for $\omega\in\mathbb S^1$.
For each $i$, the set of directions $\omega$ with
$y(\omega)\in\Gamma_i$ contains a semicircle and hence
has arc length at least $\pi$.
For each fixed $\omega$, the exit point $y(\omega)$
belongs to at most $L$ of the arcs
$\Gamma_1,\ldots,\Gamma_k$. Integrating with respect to arc length gives
\[
 \begin{aligned}
 k\pi\le \sum_{i=1}^k\int_{\mathbb S^1}
       \mathbf1_{\Gamma_i}(y(\omega))\,d\omega=\int_{\mathbb S^1}\sum_{i=1}^k
       \mathbf1_{\Gamma_i}(y(\omega))\,d\omega
 \le 2\pi L.
 \end{aligned}
\]
Hence $k\le2L$.
\end{proof}

\begin{proposition}
\label{intr:planar-cover}
There are absolute constants $A,C>0$ and
$\tau_0\in(0,1/2)$ such that, for every planar convex body
$D$ and $0<\tau<\tau_0$, there exist caps
$E_1,\ldots,E_N$ of $D$ with the following properties.
Every cap of relative area in $[\tau/2,\tau]$ is contained
in some $E_i$, and
\begin{equation}\label{intr:eq-planar-cover}
 \begin{aligned}
 \frac{\tau}{2}\le\mu_D(E_i)
 &\le A\tau, \quad\forall 1\le i\le N,\\
 N
 &\le C\bigl(1+P_D(2\tau)\bigr)L(\tau),\\
 \operatorname*{ess\,sup}_{x\in D}
 \sum_{i=1}^N\mathbf1_{E_i}(x)
 &\le CL(\tau).
 \end{aligned}
\end{equation}
Moreover,
\begin{equation}\label{intr:eq-planar-profile}
 \begin{aligned}
 P_D(\tau)\le\frac{W_D(\tau)}{\tau}
 \le C\bigl(1+P_D(2\tau)\bigr)L(\tau)\le C\bigl(1+P_D(\tau)\bigr)L(\tau).
 \end{aligned}
\end{equation}
In particular, for every $1\le q<\infty$ and all
coefficients $c_1,\ldots,c_N$,
\[
 \left\|
 \sum_{i=1}^N c_i\mu_D(E_i)^{-1/q}\mathbf1_{E_i}
 \right\|_{L^q(\mu_D)}
 \le C_qL(\tau)^{b_q}
 \left(\sum_{i=1}^N|c_i|^q\right)^{1/q},
\]
where $b_q=1-1/q$.
\end{proposition}

\begin{proof}
By scaling, assume $|D|=1$, and choose $0<\tau_0<1/28$.
For a closed boundary arc $\Gamma\subset\partial D$, set
$E(\Gamma)=\operatorname{conv}\Gamma$.
If $0<|E(\Gamma)|<1$, then $E(\Gamma)$ is a cap of $D$,
cut off by the line through the endpoints of $\Gamma$.

\medskip
\noindent\textit{Step 1. A local covering construction.} 
Fix $z\in\partial D$ and put $T=2\tau$.
All boundary arcs are taken in the clockwise direction. By continuity, choose $A,B\in\partial D$ so that the arcs
$[A,z]$ and $[z,B]$ each cut off area $T$.
These arcs meet only at $z$: otherwise they cover
$\partial D$, and their caps cover $D$, contradicting $2T<1$.

Now let $C$ be a cap with $0<|C|\le\tau$ whose boundary
arc contains $z$. Denote the endpoints of this arc by $L,R$,
ordered clockwise along the arc. Then
\[
 L\in[A,z],\qquad R\in[z,B].
\]
Otherwise, the boundary arc of $C$ contains
one of $[A,z]$ and $[z,B]$, giving $|C|\ge T=2\tau$,
which contradicts $|C|\le\tau$.
Figure~\ref{intr:fig-anchor} illustrates this configuration.
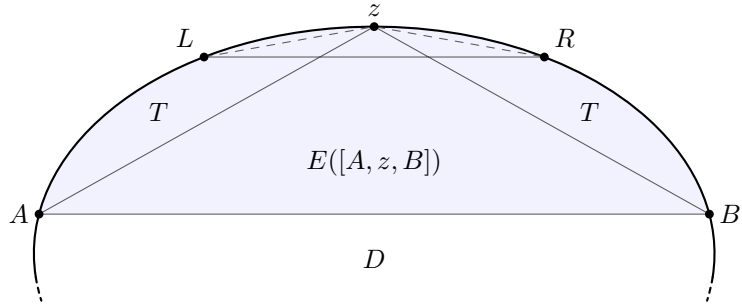
\begin{figure}[!htbp]
\centering
\begin{tikzpicture}[
  scale=1.5,
  line cap=round,
  line join=round,
  font=\small
]
\coordinate (A) at ({3*cos(170)},{2*sin(170)});
\coordinate (L) at ({3*cos(120)},{2*sin(120)});
\coordinate (z) at (0,2);
\coordinate (R) at ({3*cos(60)},{2*sin(60)});
\coordinate (B) at ({3*cos(10)},{2*sin(10)});

\fill[blue!5]
  (A) arc[start angle=170,end angle=10,
          x radius=3cm,y radius=2cm]
  -- cycle;

\draw[line width=0.8pt]
  ({3*cos(186)},{2*sin(186)})
  arc[start angle=186,end angle=-6,
      x radius=3cm,y radius=2cm];

\draw[line width=0.8pt,dash pattern=on 2pt off 2pt]
  ({3*cos(186)},{2*sin(186)})
  arc[start angle=186,end angle=192,
      x radius=3cm,y radius=2cm];
\draw[line width=0.8pt,dash pattern=on 2pt off 2pt]
  ({3*cos(-6)},{2*sin(-6)})
  arc[start angle=-6,end angle=-12,
      x radius=3cm,y radius=2cm];

\draw[line width=0.4pt,black!65]
  (A)--(B)
  (A)--(z)--(B)
  (L)--(R);

\draw[line width=0.4pt,black!65,dashed]
  (L)--(z)--(R);

\foreach \P in {A,L,z,R,B}
  \fill (\P) circle (1.1pt);

\node[left]        at (A) {$A$};
\node[above left]  at (L) {$L$};
\node[above]       at (z) {$z$};
\node[above right] at (R) {$R$};
\node[right]       at (B) {$B$};

\node at (-1.9,1.25) {$T$};
\node at (1.9,1.25)  {$T$};
\node at (0,0.8)     {$E([A,z,B])$};
\node at (0,-0.05)   {$D$};
\end{tikzpicture}
\caption{The local cap construction, showing only part of $D$.
The arcs $[A,z]$ and $[z,B]$ each cut off area $T=2\tau$.
A cap of area at most $\tau$ whose boundary arc contains $z$
has endpoints $L\in[A,z]$ and $R\in[z,B]$
and contains the triangle $zLR$.}
\label{intr:fig-anchor}
\end{figure}

Translate $z$ to the origin and regard $A$ and $B$
as column vectors in $\mathbb R^2$. Define $\Delta=|\det(A,B)|$, where $\det(A,B)$ denotes the determinant of the
$2\times2$ matrix with columns $A$ and $B$.
Thus $\Delta/2$ is the area of the triangle $AzB$. The set $E([A,z,B])$ is the union of the caps
$E([A,z])$, $E([z,B])$ and the triangle $AzB$,
with pairwise disjoint interiors.
If $\Delta\le T$, then
\[
 0<|E([A,z,B])|
 =2T+\frac{\Delta}{2}
 \le\frac{5T}{2}=5\tau<1.
\]
Thus $E([A,z,B])$ is a cap containing every cap of area
at most $\tau$ whose boundary arc contains $z$. This single cap gives the required local cover when $\Delta\le T$.

It remains to consider $\Delta>T$, where the area of
$E([A,z,B])$ may be too large compared with $\tau$.
Write
\[
 L=aA-uB,\qquad R=-vA+bB,\qquad \delta=T/\Delta<1.
\]
Convexity and the clockwise boundary order $A,L,z,R,B$
give $a,b,u,v\ge0$.
The triangles $zAL$ and $zRB$ lie in $E([A,z])$ and
$E([z,B])$, respectively, each of area $T$.
Their areas therefore satisfy
\[
 \frac12|\det(A,L)|=\frac{u\Delta}{2}\le T,\qquad
 \frac12|\det(R,B)|=\frac{v\Delta}{2}\le T.
\]
The line $AB$ consists of the points $sA+tB$ with $s+t=1$,
and the closed half-plane containing $z=0$ is given by
$s+t\le1$. Since $L=aA-uB$ and $R=-vA+bB$ lie in this
half-plane, we obtain $a-u\le1$ and $b-v\le1$. Therefore
\begin{equation}\label{intr:eq-anchor-coordinates}
 0\le u,v\le2\delta,\qquad
 0\le a,b\le1+2\delta\le3.
\end{equation}
The triangle $zLR$ has area $(ab-uv)\Delta/2$,
with $ab-uv\ge0$ by the clockwise order of $L,z,R$.
Since this triangle lies in $C$ and $|C|\le\tau$, we obtain
\begin{equation}\label{intr:eq-anchor-product}
 0\le ab-uv\le\frac{2\tau}{\Delta}=\delta,
 \qquad ab\le\delta+4\delta^2\le5\delta.
\end{equation}
The inequality $ab\le5\delta$ applies to the two endpoints of the same cap. When a left endpoint and a right endpoint come from different caps, their coordinates need not satisfy this inequality. We therefore group the caps with $a>\delta$ so that any two values of $a$ in the same group differ by at most a factor of two.

Since $0\le a\le3$, partition $[0,3]$ into $[0,\delta]$ and the intervals
$(2^k\delta,\min\{2^{k+1}\delta,3\}]$
for integers $k\ge0$ satisfying $2^k\delta<3$.
Since $\Delta/2\le |D|=1$, we have $\delta=T/\Delta\ge\tau$,
so there are at most $CL(\tau)$ intervals.

For each interval $I$, let $\mathcal F_I$ consist of all
caps $C$ with $0<|C|\le\tau$ whose boundary arc contains $z$
and whose left endpoint $L=aA-uB$ satisfies $a\in I$.
If $\mathcal F_I\ne\varnothing$, its boundary arcs all
contain $z$ and lie in $[A,z,B]$. The closure of their union
is therefore a closed arc, which we denote by
\[
 [L_*,z,R_*]
 =\overline{\bigcup_{C\in\mathcal F_I}(C\cap\partial D)}.
\]
The endpoints $L_*$ and $R_*$ are limits of left and right
endpoints, respectively, of caps in $\mathcal F_I$.
By construction, $E([L_*,z,R_*])$ contains every cap
in $\mathcal F_I$.

Write
\[
 L_*=a_*A-u_*B,\qquad R_*=-v_*A+b_*B,
\]
with $a_*,b_*,u_*,v_*\ge0$. For $I=(2^k\delta,\min\{2^{k+1}\delta,3\}]$,
\eqref{intr:eq-anchor-product} gives
$b\le5\delta/a\le5\cdot2^{-k}$ for every cap in $\mathcal F_I$.
Passing to the endpoint limits therefore yields
\[
 a_*\le2^{k+1}\delta,\qquad b_*\le5\cdot2^{-k}.
\]
For $I=[0,\delta]$, the same limiting argument gives
$a_*\le\delta$ and $b_*\le3$.
Thus $a_*b_*\le10\delta$ in both cases.

Since $[L_*,z]\subset[A,z]$ and $[z,R_*]\subset[z,B]$,
the corresponding caps each have area at most $T$. Hence
\[
 |E([L_*,z,R_*])|
 \le2T+\frac{a_*b_*\Delta}{2}
 \le7T=14\tau<1.
\]
Since $0<|E([L_*,z,R_*])|<1$, this set is a cap.
Taking one such cap for each nonempty $\mathcal F_I$
gives the required local cover, with at most $CL(\tau)$ caps.

\medskip
\noindent\textit{Step 2. The global cover and its overlap bound.} 
Fix $z_0\in\partial D$ and successively choose points
$z_1,\ldots,z_m$ in clockwise order so that
\[
 \Gamma_j=[z_{j-1},z_j],\qquad
 |E(\Gamma_j)|=T,\qquad 1\le j\le m.
\]
Stop when the remaining arc $\Gamma_{\mathrm{rem}}=[z_m,z_0]$,
possibly degenerate, satisfies $0\le |E(\Gamma_{\mathrm{rem}})|<T$. By Lemma~\ref{intr:arc-facts}, the caps $E(\Gamma_1),\ldots,E(\Gamma_m)$ have disjoint interiors. Hence
\[
 m\le P_D(T)\le\frac1T.
\]
Thus the process stops after selecting at most
$\lfloor1/T\rfloor$ full arcs. 

Let $C$ be a cap with $0<|C|\le\tau$. Its boundary arc
is either contained in one of the partition arcs
$\Gamma_1,\ldots,\Gamma_m,\Gamma_{\mathrm{rem}}$
or contains a partition endpoint $z_j$.
Apply Step 1 at each partition endpoint $z_j$ and collect
all caps so obtained. Add $E(\Gamma_1),\ldots,E(\Gamma_m)$,
as well as $E(\Gamma_{\mathrm{rem}})$ if its area is positive. The resulting caps $E_1,\ldots,E_N$ satisfy
\[
 N\le C(m+1)L(\tau),\qquad
 0<|E_i|\le14\tau,\quad\forall 1\le i\le N,
\]
and every cap of area in $(0,\tau]$ is contained in
at least one $E_i$.

To bound overlap, fix $y\in\partial D$.
Since $E(\Gamma_1),\ldots,E(\Gamma_m)$ have disjoint
interiors and each has area $T$,
\[
 \#\{1\le j\le m:\Gamma_j\subset E_i\}
 \le \frac{|E_i|}{T}\le7.
\]
Suppose $E_i$ comes from Step 1 at $z_j$ and contains $y$.
The subarc joining $z_j$ to $y$ within $E_i\cap\partial D$
consists of at most seven of the arcs $\Gamma_1,\ldots,\Gamma_m$,
possibly $\Gamma_{\mathrm{rem}}$, and at most one partial
partition arc. It is therefore contained in the union of at most nine
consecutive partition arcs, each with convex hull of area
at most $T$. For $z_j\ne y$, the subarc from $y$ to $z_j$ meets
successive partition endpoints after $y$, at most one new endpoint per partition arc. Hence $z_j$ is among the first nine
endpoints in that direction.
Allowing both directions and $z_j=y$ gives at most
$19$ possible endpoints, each contributing at most
$CL(\tau)$ caps by Step 1. 
Among $E(\Gamma_1),\ldots,E(\Gamma_m)$ and
$E(\Gamma_{\mathrm{rem}})$ when included, at most two
contain $y$. Consequently, Lemma~\ref{intr:arc-facts} gives
\[
 \left\|\sum_{i=1}^N\mathbf1_{E_i}\right\|_{L^\infty(D)}
 \le
 2\sup_{y\in\partial D}\sum_{i=1}^N\mathbf1_{E_i}(y)
 \le CL(\tau).
\]

Finally, retain only those $E_i$ that contain a cap of area
in $[\tau/2,\tau]$. Every cap in this area range is still
contained in a retained $E_i$, and each retained $E_i$ satisfies $\frac{\tau}{2}\le |E_i|\le14\tau$. The bounds on their number and overlap remain valid.
Together with $m\le P_D(2\tau)$, this proves
\eqref{intr:eq-planar-cover}.

\medskip
\noindent\textit{Step 3. Bounds for $W_D(\tau)$ and $L^q$ estimates for sums of cap indicators.} 
Caps of area $\tau$ with disjoint interiors lie in the wet
part, so $\tau P_D(\tau)\le W_D(\tau)$.
The caps constructed in Step 2 before deletion cover
the wet part. Summing their areas therefore gives
\[
 W_D(\tau)
 \le C\tau\bigl(1+P_D(2\tau)\bigr)L(\tau).
\]
Together with $P_D(2\tau)\le P_D(\tau)$, this proves
\eqref{intr:eq-planar-profile}.

For the retained caps $E_i$, let $c_i$ be arbitrary
coefficients. Since
$\|\mu_D(E_i)^{-1/q}\mathbf1_{E_i}\|_q=1$,
pointwise H\"older's inequality followed by integration
with respect to $\mu_D$ gives
\[
\begin{aligned}
 \left\|\sum_i c_i\mu_D(E_i)^{-1/q}\mathbf1_{E_i}\right\|_q
 \le\left\|\sum_i\mathbf1_{E_i}\right\|_\infty^{b_q}
 \left(\sum_i|c_i|^q\right)^{1/q}\le CL(\tau)^{b_q}
 \left(\sum_i|c_i|^q\right)^{1/q}.
\end{aligned}
\]
The last inequality uses the overlap bound from Step 2.
This proves the claimed $L^q$ estimate.
\end{proof}

\begin{remark}
The logarithmic factor in~\eqref{intr:eq-planar-profile}
cannot in general be removed: $W_D(\tau)/\tau$ can grow
like $L(\tau)$ even when $P_D(\tau)$ is bounded
independently of $\tau$.

To see this, take the unit square $D=[0,1]^2$.
Every cap of positive area contains a neighbourhood in $D$
of at least one vertex. Caps with disjoint interiors cannot
contain such neighbourhoods of the same vertex, so
$P_D(\tau)\le4$.

For $0<x,y<1/2$, the triangle
\[
 \operatorname{conv}\{(0,0),(2x,0),(0,2y)\}
\]
is a cap of area $2xy$ containing $(x,y)$.
Thus the wet part of $D$ at level $\tau$ contains
$\{(x,y)\in(0,1/2)^2:xy\le\tau/2\}$. For $0<\tau<1/2$, this gives
\[
 W_D(\tau)
 \ge \int_0^{1/2}
       \min\left\{\frac12,\frac{\tau}{2x}\right\}\,dx
 =\frac{\tau}{2}\left(1+\log\frac1{2\tau}\right)
 \asymp\tau L(\tau).
\]
The matching upper bound follows from
\eqref{intr:eq-planar-profile} and $P_D(2\tau)\le4$.
Hence $W_D(\tau)\asymp\tau L(\tau)$ as $\tau\downarrow0$.
\end{remark}

\subsubsection{Products with a Fixed Polytope}

Let $P\subset\mathbb R^p$ be a fixed polytope with nonempty
interior, where $p\ge1$. The relative volume of its wet part
satisfies
\begin{equation}\label{intr:eq-polytope-wet}
 W_P(\tau)\le C_P\tau\bigl(1+\log(1/\tau)\bigr)^{p-1},
 \qquad\forall 0<\tau<1/2.
\end{equation}
For $p\ge2$, this follows from
\cite[equation~(0.2)]{BesauSchuettWerner2018}
after normalization and an adjustment of $C_P$.
For $p=1$, it follows from $W_P(\tau)=2\tau$.

For $p\ge2$ and all sufficiently large integers $\ell$,
Lemma~\ref{hil:economic-cover} provides caps
$E_{\ell,1},\ldots,E_{\ell,N_\ell}\subset P$ such that
every cap $C$ with
\[
 2^{-\ell-1}\le\mu_P(C)\le2^{-\ell}
\]
is contained in some $E_{\ell,k}$, and
\[
 \mu_P(E_{\ell,k})\asymp_p2^{-\ell},
 \qquad
 N_\ell\le C_p2^\ell W_P(2^{-\ell})
       \le C_P(\ell+1)^{p-1}.
\]
For $p=1$ and $\ell\ge1$, set $N_\ell=2$ and choose
$E_{\ell,1}$ and $E_{\ell,2}$ to be the subintervals of $P$
of relative length $2^{-\ell}$ sharing its left and right
endpoints, respectively. Every cap in the layer
$[2^{-\ell-1},2^{-\ell}]$ is contained in one of them. Choose $\ell_0=\ell_0(p)\ge1$ so that the above constructions
apply for $\ell\ge\ell_0$. For $0\le\ell<\ell_0$, set
$N_\ell=1$ and $E_{\ell,1}=P$.
Since $1\asymp_p2^{-\ell}$ on these finitely many levels,
the same volume and number bounds hold after adjusting
the constants.

For each fixed $\ell$, the overlap is bounded by the
number of sets:
\[
 \left\|\sum_{k=1}^{N_\ell}\mathbf1_{E_{\ell,k}}
 \right\|_{L^\infty(P)}
 \le N_\ell
 \le C_P(\ell+1)^{p-1}.
\]
We use these volume, number, and overlap bounds in the
covering construction for $D\times P$ below.

\begin{proposition}
\label{intr:product-cover}
Let $K=D\times P$, where $D\subset\mathbb R^2$ is a convex
body and $P\subset\mathbb R^p$ is a fixed polytope with
nonempty interior, $p\ge1$.
There exist constants $A,C\ge1$, $c\in(0,1)$,
$\tau_0\in(0,1/2)$, and an integer $M\ge1$, depending
only on $p$ and $P$, such that the following statements
hold for every $0<\tau<\tau_0$.

There are convex bodies $E_1,\ldots,E_N\subset K$ such
that every cap of $K$ with relative volume in
$[\tau/2,\tau]$ is contained in some $E_i$, and
\begin{equation}\label{intr:eq-product-cover}
 \begin{aligned}
  \frac{\tau}{2}\le\mu_K(E_i)
  &\le A\tau,\qquad \forall 1\le i\le N,\\
  N
  &\le C\bigl(1+P_D(c\tau)\bigr)L(\tau)^{p+1},\\
  \left\|\sum_{i=1}^N\mathbf1_{E_i}\right\|_{L^\infty(K)}
  &\le CL(\tau)^{p+1}.
 \end{aligned}
\end{equation}
Moreover,
\begin{equation}\label{intr:eq-product-profile}
 1+P_D(\tau)\le1+P_K(\tau)
 \le2\mathcal C_K(\tau)
 \le C\bigl(1+P_D(\tau)\bigr)L(\tau)^M.
\end{equation}
Consequently,
\[
 \limsup_{\tau\downarrow0}
 \frac{\log\mathcal C_K(\tau)}{\log(1/\tau)}
 =
 \limsup_{\tau\downarrow0}
 \frac{\log\mathcal C_D(\tau)}{\log(1/\tau)},
\]
and the same equality holds with $\liminf$ in place
of $\limsup$.
\end{proposition}

\begin{proof}
Put $d=p+2$. We use normalized volume measures, so
$\mu_K=\mu_D\otimes\mu_P$.
All implicit constants depend only on $p$ and $P$.

\medskip
\noindent\textit{Step 1. A volume comparison.}
Write
\[
 C=\{(y,z)\in K:u\cdot y+v\cdot z\ge t\},
\]
where $u\in\mathbb R^2$, $v\in\mathbb R^p$, and
$(u,v)\ne0$. Set $h=H_D(u)+H_P(v)-t$ and
\[
 A(y)=H_D(u)-u\cdot y,\qquad
 B(z)=H_P(v)-v\cdot z.
\]
Then $A\ge0$ on $D$ and $B\ge0$ on $P$.
Since $C$ has positive volume,
$t<H_D(u)+H_P(v)$, so $h>0$.
Consequently,
\[
 C=\{(y,z)\in K:A(y)+B(z)\le h\}.
\]
Set
\[
 C_D(h)=\{y\in D:A(y)\le h\},\qquad
 C_P(h)=\{z\in P:B(z)\le h\}.
\]
Choose $y_0\in D$ and $z_0\in P$ with
$A(y_0)=B(z_0)=0$.
Since $D$ and $P$ are convex and $A$ and $B$ are affine,
the map
\[
 (y,z)\longmapsto
 \left(\frac{y+y_0}{2},\frac{z+z_0}{2}\right)
\]
sends $C_D(h)\times C_P(h)$ into $C$ and multiplies
volume by $2^{-d}$.
Since $C\subset C_D(h)\times C_P(h)$, we obtain
\begin{equation}\label{intr:eq-product-volume}
 2^{-d}\mu_D(C_D(h))\mu_P(C_P(h))
 \le\mu_K(C)
 \le\mu_D(C_D(h))\mu_P(C_P(h)).
\end{equation}
This argument also applies when $u=0$ or $v=0$, or when
$C_D(h)=D$ or $C_P(h)=P$.

\medskip
\noindent\textit{Step 2. Construction of the cover.}
Apply Proposition~\ref{intr:planar-cover} to $D$ with
$\tau=2^{-i}$ for all sufficiently large integers $i$,
and denote the resulting family by $\mathcal E_i$.
For the finitely many remaining $i\ge0$, set
$\mathcal E_i=\{D\}$.
Then every cap of $D$ with relative area in
$[2^{-i-1},2^{-i}]$ is contained in some $E\in\mathcal E_i$.

For $P$, use the sets $E_{\ell,k}\subset P$ constructed
before this proposition, and write
\[
 \mathcal F_\ell=\{E_{\ell,1},\ldots,E_{\ell,N_\ell}\},
 \qquad \ell\ge0.
\]
By construction, every cap of $P$ with relative volume
in $[2^{-\ell-1},2^{-\ell}]$ is contained in some
$F\in\mathcal F_\ell$.

For all $i,\ell\ge0$, the volume bounds are
\[
 \mu_D(E)\asymp2^{-i},\quad\forall E\in\mathcal E_i,
 \qquad
 \mu_P(F)\asymp2^{-\ell},\quad\forall F\in\mathcal F_\ell.
\]
The covering estimates and the monotonicity of $P_D$
give, for some fixed $c_0\in(0,1)$,
\begin{equation}\label{intr:eq-product-family-bounds}
 \begin{aligned}
 \#\mathcal E_i
 &\le C\bigl(1+P_D(c_0 2^{-i})\bigr)(i+1),\\
 \left\|\sum_{E\in\mathcal E_i}\mathbf1_E
 \right\|_{L^\infty(D)}
 &\le C(i+1),\\
 \left\|\sum_{F\in\mathcal F_\ell}\mathbf1_F
 \right\|_{L^\infty(P)}
 &\le \#\mathcal F_\ell
 \le C(\ell+1)^{p-1}.
 \end{aligned}
\end{equation}
Fix a cap $C$ of $K$ with $\tau/2\le\mu_K(C)\le\tau$,
and use the notation of Step 1. Choose integers $i,\ell\ge0$ such that
\[
 2^{-i-1}<\mu_D(C_D(h))\le2^{-i},\qquad
 2^{-\ell-1}<\mu_P(C_P(h))\le2^{-\ell}.
\]
Equation~\eqref{intr:eq-product-volume} and these bounds give
$\frac{\tau}{2}\le \mu_D(C_D(h))\mu_P(C_P(h))\le2^d\tau$ and
\[
 \mu_D(C_D(h))\mu_P(C_P(h))\le2^{-(i+\ell)}<4\mu_D(C_D(h))\mu_P(C_P(h)).
\]
Thus $(i,\ell)$ belongs to the finite set
\[
 \mathcal I_\tau
 =
 \left\{(i,\ell)\in\mathbb Z_{\ge0}^2:
 \frac{\tau}{2}\le2^{-(i+\ell)}\le2^{d+2}\tau\right\}.
\]
The covering properties give $E\in\mathcal E_i$ and
$F\in\mathcal F_\ell$ containing $C_D(h)$ and $C_P(h)$,
respectively. Hence
\[
 C\subset C_D(h)\times C_P(h)\subset E\times F.
\]

Form all products $E\times F$ with
$(i,\ell)\in\mathcal I_\tau$, $E\in\mathcal E_i$,
and $F\in\mathcal F_\ell$.
These sets are convex, and their relative volumes satisfy
\[
 \mu_K(E\times F)
 =\mu_D(E)\mu_P(F)
 \le C2^{-(i+\ell)}
 \le A\tau,
\]
where $A$ depends only on $p$ and $P$.
Retain only those products containing a cap of $K$
with relative volume in $[\tau/2,\tau]$, and denote
them by $E_1,\ldots,E_N$.
Every cap in this range is still contained in some $E_j$,
and
\[
 \frac{\tau}{2}\le\mu_K(E_j)\le A\tau,
 \qquad 1\le j\le N.
\]

For $(i,\ell)\in\mathcal I_\tau$, we have
$2^{-i}\ge2^{-(i+\ell)}\ge\tau/2$.
Since $P_D$ is nonincreasing, taking $c=c_0/2$ gives $P_D(c_0 2^{-i})\le P_D(c\tau)$. The definition of $\mathcal I_\tau$ also gives
\[
 \log_2(1/\tau)-d-2
 \le i+\ell\le\log_2(1/\tau)+1.
\]
Thus $s=i+\ell$ takes at most $d+4$ integer values.
For each such $s$, there are $s+1$ pairs of nonnegative
integers $(i,\ell)$ with $i+\ell=s$, and each term
$(i+1)(\ell+1)^{p-1}$ is at most $(s+1)^p$. Consequently,
\begin{equation}\label{intr:eq-product-index-sum}
 \sum_{(i,\ell)\in\mathcal I_\tau}
 (i+1)(\ell+1)^{p-1}
 \le CL(\tau)^{p+1}.
\end{equation}

For each fixed pair $(i,\ell)$, there are
$\#\mathcal E_i\,\#\mathcal F_\ell$ products, and
\[
 \sum_{E\in\mathcal E_i}\sum_{F\in\mathcal F_\ell}
 \mathbf1_{E\times F}(y,z)
 =
 \left(\sum_{E\in\mathcal E_i}\mathbf1_E(y)\right)
 \left(\sum_{F\in\mathcal F_\ell}\mathbf1_F(z)\right).
\]
Applying~\eqref{intr:eq-product-family-bounds} and using
$P_D(c_0 2^{-i})\le P_D(c\tau)$, we bound the number
and overlap of the products for each
$(i,\ell)\in\mathcal I_\tau$.
Summing these bounds and
using~\eqref{intr:eq-product-index-sum}, we obtain
\[
\begin{aligned}
 N
 &\le \sum_{(i,\ell)\in\mathcal I_\tau}
       \#\mathcal E_i\,\#\mathcal F_\ell\\
 &\le C\bigl(1+P_D(c\tau)\bigr)
       \sum_{(i,\ell)\in\mathcal I_\tau}
       (i+1)(\ell+1)^{p-1}\\
 &\le C\bigl(1+P_D(c\tau)\bigr)L(\tau)^{p+1}.
\end{aligned}
\]
Similarly, the preceding product identity gives
\[
\begin{aligned}
 \left\|\sum_{j=1}^N\mathbf1_{E_j}\right\|_{L^\infty(K)}
 &\le \sum_{(i,\ell)\in\mathcal I_\tau}
 \left\|\sum_{E\in\mathcal E_i}\mathbf1_E
 \right\|_{L^\infty(D)}
 \left\|\sum_{F\in\mathcal F_\ell}\mathbf1_F
 \right\|_{L^\infty(P)}\\
 &\le C\sum_{(i,\ell)\in\mathcal I_\tau}
       (i+1)(\ell+1)^{p-1}\\
 &\le CL(\tau)^{p+1}.
\end{aligned}
\]
Together with the volume bounds, this
proves~\eqref{intr:eq-product-cover}.

\medskip
\noindent\textit{Step 3. Packing and wet part estimates.}
For every cap $Q$ of $D$, $Q\times P$ is a cap of $K$
with the same relative volume.
If two caps of $D$ have disjoint interiors, so do
their products with $P$.
Moreover, the product of the wet part of $D$ at level
$\tau$ with $P$ lies in the wet part of $K$ at the same level.
Thus
\[
 P_D(\tau)\le P_K(\tau),\qquad
 W_D(\tau)\le W_K(\tau).
\]

Any cap of relative volume below $\tau/2$ is contained in a cap of relative volume exactly $\tau/2$, obtained by moving the cutting hyperplane parallel to itself. Such a position exists by continuity of cap volume. The larger cap is contained in one of the sets $E_j$ retained in Step 2, so the original cap is contained in the same set. Thus $E_1,\ldots,E_N$ cover the wet part of $K$ at level $\tau$, and
\[
\begin{aligned}
 \mathcal C_K(\tau)
 &=\frac{W_K(\tau)}{\tau}
 \le\frac1\tau\sum_{j=1}^N\mu_K(E_j)\\
 &\le AN
 \le C\bigl(1+P_D(c\tau)\bigr)L(\tau)^{p+1}.
\end{aligned}
\]
By~\eqref{intr:eq-planar-profile},
\[
 1+P_D(s)\le C\bigl(1+P_D(2s)\bigr)L(s)
\]
for sufficiently small $s$.
Choose the smallest integer $r\ge1$ such that $2^rc\ge1$.
Iterating the preceding inequality at $s=2^kc\tau$,
$k=0,\ldots,r-1$, and using $L(2^kc\tau)\le CL(\tau)$,
we obtain
\[
\begin{aligned}
 1+P_D(c\tau)
 &\le C\bigl(1+P_D(2^rc\tau)\bigr)L(\tau)^r\\
 &\le C\bigl(1+P_D(\tau)\bigr)L(\tau)^r,
\end{aligned}
\]
where the last inequality uses $2^rc\tau\ge\tau$
and the monotonicity of $P_D$.
Substituting this estimate into the bound for
$\mathcal C_K(\tau)$ proves the upper bound
in~\eqref{intr:eq-product-profile} with $M=p+1+r$.

Since $P_K(\tau)\le\mathcal C_K(\tau)$ and
$\mathcal C_K(\tau)\ge1$, the bound
$P_D(\tau)\le P_K(\tau)$ completes the proof
of~\eqref{intr:eq-product-profile}. Finally, $W_D(\tau)\le W_K(\tau)$ and
$1+P_D(\tau)\le2\mathcal C_D(\tau)$ give
\[
 \mathcal C_D(\tau)
 \le\mathcal C_K(\tau)
 \le C\mathcal C_D(\tau)L(\tau)^M.
\]
Since $\log L(\tau)/\log(1/\tau)\to0$, the asserted
equalities of the upper and lower limits follow.
\end{proof}

Propositions~\ref{intr:planar-cover} and~\ref{intr:product-cover}
give the cap covering and wet part estimates needed to apply
Proposition~\ref{geo:layer-upper}.
We now combine the resulting upper bounds with the cap packing
lower bound in Proposition~\ref{geo:packing-lower} and the
interior lower bound in Theorem~\ref{geo:interior-lower}
to prove Theorem~\ref{main:intrinsic}.

\begin{proof}[Proof of Theorem~\ref{main:intrinsic}]
By the affine invariance of relative cap volumes and
Proposition~\ref{geom:atoms}, it suffices to consider planar
bodies $K$ and products $K=D\times P$.
Put $\varepsilon=1/q-\eta>0$.

For planar $K$, apply~\eqref{intr:eq-planar-profile}
with $D=K$. Since $P_K(\tau)\ge1$, we obtain
\[
\begin{aligned}
 1+P_K(\tau)
 &\le2P_K(\tau)\le2\mathcal C_K(\tau)\\
 &\le2C\bigl(1+P_K(2\tau)\bigr)L(\tau)\le2C\bigl(1+P_K(\tau)\bigr)L(\tau),
\end{aligned}
\]
where the last inequality uses the monotonicity of $P_K$.

For $K=D\times P$, \eqref{intr:eq-product-profile} gives
\[
\begin{aligned}
 1+P_K(\tau)
 &\le2\mathcal C_K(\tau)\le C\bigl(1+P_D(\tau)\bigr)L(\tau)^M\le C\bigl(1+P_K(\tau)\bigr)L(\tau)^M,
\end{aligned}
\]
where the last inequality uses
$P_D(\tau)\le P_K(\tau)$, also given by
\eqref{intr:eq-product-profile}.
Since $P_K(\tau)\ge1$, we have
$\log(1+P_K(\tau))=\log P_K(\tau)+O(1)$.
Taking logarithms in the preceding comparison and using
$\log L(\tau)=o(\log(1/\tau))$ therefore gives
\[
 \frac{\log P_K(\tau)}{\log(1/\tau)}
 =
 \frac{\log\mathcal C_K(\tau)}{\log(1/\tau)}+o(1).
\]
Thus $P_K$ can replace $\mathcal C_K$ in the definitions
in~\eqref{intr:eq-kappa}.

Choose $\tau_0>0$ sufficiently small. At scales
$\tau_j=\tau_0 2^{-j}$, $j\ge0$, use the covers from
Proposition~\ref{intr:planar-cover} when $K$ is planar
and from Proposition~\ref{intr:product-cover} when
$K=D\times P$.
Assign each cap with relative volume in $(\tau_j/2,\tau_j]$
to one covering set containing it.
For each covering set $E$, the caps assigned to $E$
form one group. Caps with relative volume greater than $\tau_0$ form
one additional group with containing set $K$.

Since $L(\tau_j)\asymp j+1$, the number bounds in
\eqref{intr:eq-planar-cover} and~\eqref{intr:eq-product-cover}
give, for some fixed $c\in(0,1)$,
\[
\begin{aligned}
 N_j
 &\le C\bigl(1+P_K(c\tau_j)\bigr)(j+1)^{d-1}\\
 &\le C\mathcal C_K(c\tau_j)(j+1)^{d-1}.
\end{aligned}
\]
Here we use monotonicity of the packing numbers,
$P_D\le P_K$ in the product case, and
$1+P_K\le2\mathcal C_K$.
Since $\log(1/(c\tau_j))=j\log2+O(1)$ and
$\log(j+1)=o(j)$, \eqref{intr:eq-kappa} gives
\[
 \limsup_{j\to\infty}\frac{\log N_j}{j\log2}
 \le
 \limsup_{j\to\infty}
 \frac{\log\mathcal C_K(c\tau_j)}{\log(1/(c\tau_j))}
 \le\overline\kappa(K).
\]
The overlap bounds in~\eqref{intr:eq-planar-cover} and
\eqref{intr:eq-product-cover}, together with
\eqref{geo:eq-simple-synthesis}, give
\[
 1\le S_j(q)\le C(j+1)^{(d-1)b_q}.
\]
Consequently, $\log S_j(q)/j\to0$. Thus~\eqref{geo:eq-subpower} holds with
$\kappa=\overline\kappa(K)$.

Set
\[
 R=\min\left\{\alpha,\,
 b_q+\frac{\varepsilon}{\overline\kappa(K)}\right\},
\]
with $\varepsilon/0=+\infty$.
Proposition~\ref{geo:layer-upper} therefore gives, for every
$0\le r<R$,
\[
 \mathcal E_n^K(w_{K,\eta};q)
 \le C_r(n+1)^{-r},
 \qquad n\ge0.
\]
By~\eqref{intr:eq-order}, this implies
$r_*(K,\eta,q)\ge R$.
Under the strict condition
$\varepsilon>\overline\kappa(K)\max\{\alpha-b_q,0\}$,
Proposition~\ref{geo:layer-upper}
gives~\eqref{main:eq-intrinsic-classical}.

The conclusion for finite networks in
Proposition~\ref{geo:layer-upper} gives
\eqref{main:eq-compression} for every fixed $0\le r<R$
and for the bound \eqref{main:eq-intrinsic-classical},
with the factor $M$ obtained by rescaling \(F\).
The transport identities \eqref{geom:eq-transport}
preserve the correspondence with the original atoms
and the weighted coefficient sum, so these conclusions
also hold for invertible affine images.

For the reverse inequality,
Theorem~\ref{geo:interior-lower} gives
$r_*(K,\eta,q)\le\alpha$.
If $\overline\kappa(K)>0$, choose $s_j\downarrow0$ such that,
with $m_j=P_K(s_j)$,
\[
 \frac{\log m_j}{\log(1/s_j)}
 \longrightarrow\overline\kappa(K).
\]
Then $m_j\to\infty$. Let $C_d$ be the constant in Proposition~\ref{geo:packing-lower}. Choose $0<c_d<(2C_d)^{-1}$ and set
\[
 n_j=\left\lfloor\frac{c_d m_j}{\log(m_j+2)}\right\rfloor,
 \qquad
 \widetilde m_j=
 \left\lceil C_d(n_j+1)\log(n_j+2)\right\rceil.
\]
Since $n_j\asymp c_d m_j/\log m_j$, we have
$\widetilde m_j/m_j\to C_dc_d<1$.
Thus $P_K(s_j)=m_j\ge\widetilde m_j$ for all sufficiently large $j$.
Applying Proposition~\ref{geo:packing-lower} with
$n=n_j$, $\tau=s_j$, and $m=\widetilde m_j$ gives
\[
 \mathcal E_{n_j}^K(w_{K,\eta};q)
 \ge c\,\widetilde m_j^{-b_q}s_j^\varepsilon
 \ge c\,m_j^{-b_q}s_j^\varepsilon,
\]
where the last inequality uses $b_q\ge0$.
Moreover, $\log(n_j+1)/\log m_j\to1$.
If $\mathcal E_n^K(w_{K,\eta};q)\le C_r(n+1)^{-r}$
for all $n\ge0$, the preceding lower bound gives
\[
 r\le
 \frac{b_q\log m_j+\varepsilon\log(1/s_j)+\log(C_r/c)}
      {\log(n_j+1)}
 \longrightarrow
 b_q+\frac{\varepsilon}{\overline\kappa(K)}.
\]
Taking the supremum over such $r\ge0$ in
\eqref{intr:eq-order} yields
$r_*(K,\eta,q)\le b_q+\varepsilon/\overline\kappa(K)$. 
Thus $r_*(K,\eta,q)=R$.
If $\overline\kappa(K)=0$, then $R=\alpha$.
The upper estimates from Proposition~\ref{geo:layer-upper}
give $r_*(K,\eta,q)\ge\alpha$, while
Theorem~\ref{geo:interior-lower} gives
$r_*(K,\eta,q)\le\alpha$.
Hence $r_*(K,\eta,q)=\alpha$. Thus~\eqref{main:eq-intrinsic-order} holds in both cases.

Now suppose that $\kappa=\kappa(K)>0$ exists.
Fix $0<\delta<\kappa$.
Since $\log P_K(\tau)/\log(1/\tau)\to\kappa$, $P_K(\tau)\ge\tau^{-(\kappa-\delta)}$ for all sufficiently small $\tau$.
For each sufficiently large $n$, set
\[
 m_n=\left\lceil C_d(n+1)\log(n+2)\right\rceil,
 \qquad
 \tau_n=m_n^{-1/(\kappa-\delta)}.
\]
Then $P_K(\tau_n)\ge\tau_n^{-(\kappa-\delta)}=m_n$,
so Proposition~\ref{geo:packing-lower} gives
\[
 \mathcal E_n^K(w_{K,\eta};q)
 \ge c\,m_n^{-b_q}\tau_n^\varepsilon
 =c\,m_n^{-b_q-\varepsilon/(\kappa-\delta)}.
\]
Since $\log m_n/\log n\to1$, this estimate and
Theorem~\ref{geo:interior-lower} imply
\[
 \limsup_{n\to\infty}
 \frac{-\log\mathcal E_n^K(w_{K,\eta};q)}{\log n}
 \le
 \min\left\{\alpha,\,
 b_q+\frac{\varepsilon}{\kappa-\delta}\right\}.
\]
Letting $\delta\downarrow0$ gives the upper bound $R$,
since $\overline\kappa(K)=\kappa$.

For every $0\le r<R$, the bound
$\mathcal E_n^K(w_{K,\eta};q)\le C_r(n+1)^{-r}$
proved above yields
\[
 \liminf_{n\to\infty}
 \frac{-\log\mathcal E_n^K(w_{K,\eta};q)}{\log n}
 \ge r.
\]
Letting $r\uparrow R$ proves that the limit equals
$R=r_*(K,\eta,q)$.
If $\kappa(K)=0$, \eqref{main:eq-intrinsic-classical}
and Theorem~\ref{geo:interior-lower} give the limit
$\alpha=r_*(K,\eta,q)$.

Finally, Proposition~\ref{intr:product-cover} gives
$\overline\kappa(D\times P)=\overline\kappa(D)$
and shows that $\kappa(D\times P)$ exists if and only if
$\kappa(D)$ exists, in which case their values agree.
\end{proof}

\appendix

\section{Proofs of Auxiliary Results}
\label{app:auxiliary-proofs}
\begin{proof}[Proof of Lemma~\ref{lem:wet-part-depth}]
Fix $x\in K$.
We first show that the infimum defining $v_K(x)$
is attained whenever $v_K(x)>0$.
Choose caps $C_{z_j}$, with $z_j=(\xi_j,t_j)$, such that
\[
 x\in C_{z_j},
 \qquad
 \theta_{z_j}\longrightarrow v_K(x).
\]
Since $|\xi_j|=1$ and 
\[
-\max_{y\in K}|y|\leq\min_{y\in K}\xi_j\cdot y\leq t_j<\max_{y\in K}\xi_j\cdot y\le\max_{y\in K}|y|,
\] 
we pass to a subsequence for which
$\xi_j\to\xi$ and $t_j\to t$.
Continuity of the support function gives
\[
 |\xi|=1,
 \qquad
 \min_{y\in K}\xi\cdot y\le t\le H_K(\xi).
\]
Define the limiting cap by
$C_z=K\cap\{y:\xi\cdot y\ge t\}$, with $z=(\xi,t)$.
At this stage it may have zero volume. Passing to the limit in
$\xi_j\cdot x\ge t_j$ gives $\xi\cdot x\ge t$, so $x\in C_z$.

For every $y\in K$ outside the hyperplane $\xi\cdot y=t$,
the sign of $\xi_j\cdot y-t_j$ agrees with
that of $\xi\cdot y-t$ when $j$ is sufficiently large.
Thus
\[
 \mathbf{1}_{C_{z_j}}(y)\longrightarrow\mathbf{1}_{C_z}(y)
 \qquad\text{for almost every }y\in K.
\]
The exceptional hyperplane has zero volume, so dominated
convergence yields
\[
 \mu_K(C_z)
 =\lim_{j\to\infty}\mu_K(C_{z_j})
 =v_K(x).
\]
If $v_K(x)>0$, then $\mu_K(C_z)>0$. Since a cap with
$t=H_K(\xi)$ lies in a supporting hyperplane and has
zero volume, $t<H_K(\xi)$. Hence $C_z$ is an admissible cap attaining the infimum.

Now let $0<\tau<1$.
If $v_K(x)<\tau$, the definition of the infimum provides
a cap containing $x$ with relative volume below $\tau$.
If $v_K(x)=\tau$, then $v_K(x)>0$, and the preceding
argument provides a cap containing $x$ with relative
volume exactly $\tau$.
In either case, $x$ belongs to the wet part at threshold $\tau$.
Conversely, if $x\in C_z$ for some $\theta_z\le\tau$,
then
\[
 v_K(x)\le\theta_z\le\tau.
\]
This proves the set identity.
Taking normalized volumes gives the formula for $W_K(\tau)$.
The same compactness argument, now with $x_j\to x$, shows that
the sublevel set $\{x:v_K(x)\le\tau\}$ is closed and hence measurable.
\end{proof}

\subsection{Auxiliary Approximation Lemmas}
\label{app:closure-allocation}

Let $X$ be a Banach space, and assume that the normalized dictionary $\{\phi_z/w(z):z\in Z\}$ is bounded in $X$, with $w(z)>0$. By definition, its closed absolutely convex hull $B$ is the $X$-closure of finite sums $\sum_i c_i\phi_{z_i}$ satisfying $\sum_i|c_i|w(z_i)\le1$. Define
\[
 \mathcal V_w=\bigcup_{R>0}RB,
 \qquad
 \|f\|_{\mathcal V_w}=\inf\{R>0:f\in RB\}.
\]
Then $\|\cdot\|_{\mathcal V_w}$ is a norm on $\mathcal V_w$, with closed unit ball $B$. 

To verify completeness, let $(f_j)$ be Cauchy in $\|\cdot\|_{\mathcal V_w}$. Since $B$ is bounded in $X$, let $C=\sup_{g\in B}\|g\|_X<\infty$. If $h\in RB$, then $\|h\|_X\le CR$. Taking the infimum over $R$ gives
\[
 \|h\|_X\le C\|h\|_{\mathcal V_w},
 \qquad h\in\mathcal V_w.
\]
Thus $\mathcal V_w$ is continuously embedded in $X$. In particular,
\[
 \|f_j-f_k\|_X
 \le C\|f_j-f_k\|_{\mathcal V_w}
 \longrightarrow0
 \qquad\text{as }j,k\to\infty.
\]
Thus $(f_j)$ is Cauchy in $X$ and, by completeness of $X$, converges to some $f\in X$. For every $\delta>0$, we have $f_j-f_k\in\delta B$ for all sufficiently large $j,k$. Letting $k\to\infty$ and using the closedness of $\delta B$ gives $f_j-f\in\delta B$. Thus $f\in\mathcal V_w$ and $f_j\to f$ in the variation norm.

Let $\Sigma_n\subset X$ denote the class of networks of width at most $n$. When $X=L^q(\mu_K)$, this is $\Sigma_n^K$ as defined in \eqref{eq:network-class}. Set
\[
 E_n(f)_X=\inf_{S\in\Sigma_n}\|f-S\|_X.
\]

The following lemma extends approximation error bounds for finite atomic sums to all functions in $\mathcal V_w$ by taking limits in $X$.

\begin{lemma}
\label{lem:closure}
Let $\mathcal A_w$ be the set of finite sums $\sum_i c_i\phi_{z_i}$ satisfying $\sum_i|c_i|w(z_i)\le1$, so that $B=\overline{\mathcal A_w}^{\,X}$.

\textup{(1)} For any $f,g\in X$,
\begin{equation}\label{closure:eq-lipschitz}
 |E_n(f)_X-E_n(g)_X|\le\|f-g\|_X,
 \qquad f,g\in X.
\end{equation}
Consequently,
\begin{equation}\label{closure:eq-suprema}
 \sup_{f\in B}E_n(f)_X
 =\sup_{F\in\mathcal A_w}E_n(F)_X.
\end{equation}
Suppose there is a constant $A_n>0$, independent of $N$, $c_i$, and $z_i$, such that every finite sum $F=\sum_{i=1}^N c_i\phi_{z_i}$ satisfies $E_n(F)_X\le A_n\sum_{i=1}^N|c_i|w(z_i)$. Then
\[
 E_n(f)_X\le A_n\|f\|_{\mathcal V_w}
 \qquad\text{for every }f\in\mathcal V_w.
\]

\textup{(2)} Assume, in addition, that every finite sum $F$ in \textup{(1)} admits a network $S_F=\sum_{j=1}^n a_j\phi_{z_j'}\in\Sigma_n$ satisfying
\[
 \|F-S_F\|_X\le A_n\sum_{i=1}^N|c_i|w(z_i),
 \qquad
 \sum_{j=1}^n|a_j|w(z_j')\le \sum_{i=1}^N|c_i|w(z_i).
\]
Then every $f\in\mathcal V_w$ admits a network $S=\sum_{j=1}^n a_j\phi_{z_j'}\in\Sigma_n$ satisfying
\[
 \|f-S\|_X\le2A_n\|f\|_{\mathcal V_w},
 \qquad
 \sum_{j=1}^n|a_j|w(z_j')\le\|f\|_{\mathcal V_w}.
\]
\end{lemma}
\begin{proof}
For every $S\in\Sigma_n$, the triangle inequality gives
\[
 E_n(f)_X\le\|f-S\|_X
 \le\|f-g\|_X+\|g-S\|_X.
\]
Taking the infimum over $S$ yields $E_n(f)_X\le\|f-g\|_X+E_n(g)_X$. Interchanging $f$ and $g$ proves \eqref{closure:eq-lipschitz}.

For every $f\in B$, choose $F_k\in\mathcal A_w$ with $F_k\to f$ in $X$. By \eqref{closure:eq-lipschitz}, $E_n(F_k)_X\to E_n(f)_X$, so $E_n(f)_X\le\sup_{F\in\mathcal A_w}E_n(F)_X$. Taking the supremum over $f\in B$ proves \eqref{closure:eq-suprema}.

If $f=0$, the remaining conclusions hold with the zero network. Suppose $f\ne0$. Since $B$ is the closed unit ball of $\mathcal V_w$ and $B=\overline{\mathcal A_w}^{\,X}$, $f\in\|f\|_{\mathcal V_w}B
 =\overline{\|f\|_{\mathcal V_w}\mathcal A_w}^{\,X}$. Thus there are finite representations $F_k=\sum_i c_{k,i}\phi_{z_{k,i}}$ such that
\[
 F_k\longrightarrow f\quad\text{in }X,
 \qquad
 \sum_i|c_{k,i}|w(z_{k,i})\le\|f\|_{\mathcal V_w}.
\]
The hypothesis in \textup{(1)} gives $E_n(F_k)_X\le A_n\sum_i|c_{k,i}|w(z_{k,i})\le A_n\|f\|_{\mathcal V_w}$. Passing to the limit by \eqref{closure:eq-lipschitz}, we obtain
\[
 E_n(f)_X=\lim_{k\to\infty}E_n(F_k)_X
 \le A_n\|f\|_{\mathcal V_w}.
\]

For \textup{(2)}, since $A_n\|f\|_{\mathcal V_w}>0$, choose $k$ such that $\|f-F_k\|_X\le A_n\|f\|_{\mathcal V_w}$. Apply the hypothesis in \textup{(2)} to this representation of $F_k$, obtaining a network $S=\sum_{j=1}^n a_j\phi_{z_j'}\in\Sigma_n$. Then we have
\[
 \begin{aligned}
 \|f-S\|_X
 &\le\|f-F_k\|_X+\|F_k-S\|_X\\
 &\le A_n\|f\|_{\mathcal V_w}
      +A_n\sum_i|c_{k,i}|w(z_{k,i})
 \le2A_n\|f\|_{\mathcal V_w},
 \end{aligned}
\]
and $\sum_{j=1}^n|a_j|w(z_j')
 \le\sum_i|c_{k,i}|w(z_{k,i})
 \le\|f\|_{\mathcal V_w}$. This proves \textup{(2)}.
\end{proof}

The following lemma distributes at most $N$ neurons among finitely many groups and controls the $\ell^q$ norm of the resulting local error bounds. We use it in the proof of Proposition~\ref{geo:hardy-upper}.
\begin{lemma} \label{lem:allocation}
Let $\gamma>0$, $1\le q<\infty$, and $r=(\gamma+1/q)^{-1}$. For every finite family of nonnegative numbers $(B_I)_{I\in\mathcal I}$ and every integer $N\ge1$, there exist nonnegative integers $m_I$ with $\sum_{I\in\mathcal I}m_I\le N$ such that
\begin{equation}\label{core:eq-allocation}
 \left[\sum_{I\in\mathcal I} B_I^q(m_I+1)^{-\gamma q}\right]^{1/q}
 \le N^{-\gamma}
 \left(\sum_{I\in\mathcal I} B_I^r\right)^{1/r}.
\end{equation}
\end{lemma}

\begin{proof}
Set $T=\sum_{I\in\mathcal I}B_I^r$. If $T=0$, all $B_I$ vanish, and taking $m_I=0$ proves the claim. If $T>0$, choose $m_I=\left\lfloor\frac{NB_I^r}{T}\right\rfloor$. Then $\sum_{I\in\mathcal I}m_I
 \le \sum_{I\in\mathcal I}\frac{NB_I^r}{T}
 =N$. For each $I$ with $B_I>0$, we have $m_I+1>NB_I^r/T$. Since $q-\gamma qr=r$, it follows that
\[
 \begin{aligned}
 \sum_{I\in\mathcal I}B_I^q(m_I+1)^{-\gamma q}
 &\le \left(\frac{T}{N}\right)^{\gamma q}
       \sum_{I:\,B_I>0}B_I^{q-\gamma qr}\\
 &=N^{-\gamma q}T^{\gamma q+1}.
 \end{aligned}
\]
Taking $q$-th roots and using $\gamma+1/q=1/r$ gives \eqref{core:eq-allocation}. 
\end{proof}

\begin{proof}[Proof of Proposition~\ref{prop:closure-exponent}]
The case $p=q$ is immediate, so assume $1\le p<q$. Since $\mu_K(K)=1$, the embedding $X_q(K)\hookrightarrow L^p(\mu_K)$ is continuous. Taking closures of $\mathcal A_w(K)$ therefore gives
\[
 \mathbb B_w^{(q)}(K)\subseteq\mathbb B_w^{(p)}(K).
\]
For the reverse inclusion, let $f\in\mathbb B_w^{(p)}(K)$ and choose $F_m\in\mathcal A_w(K)$ such that $F_m\to f$ in $L^p(\mu_K)$.

\emph{Case $q<\infty$.}
By the definition of $\mathcal A_w(K)$ and the boundedness assumption, $\sup_m\|F_m\|_{L^q(\mu_K)}
 \le \sup_{z\in Z_K}\|\phi_z\|_{L^q(\mu_K)}/w(z)
 <\infty$. Since $1<q<\infty$, the space $L^q(\mu_K)$ is reflexive, and its closed unit ball is weakly sequentially compact. Hence a subsequence converges weakly to some $g\in L^q(\mu_K)$. Every continuous linear functional on $L^p(\mu_K)$ restricts to a continuous linear functional on $L^q(\mu_K)$, so the subsequence also converges weakly to $g$ in $L^p(\mu_K)$. Since $F_m\to f$ strongly in $L^p(\mu_K)$, uniqueness of the weak limit yields $g=f$. The set $\mathbb B_w^{(q)}(K)$ is closed and convex in $L^q(\mu_K)$, hence weakly closed by the separation theorem for convex sets. Therefore $f\in\mathbb B_w^{(q)}(K)$.

\emph{Case $q=\infty$.}
Here $X_\infty(K)=C(K)$, so we must show that $f$ has a continuous representative in the uniform closure of $\mathcal A_w(K)$. Since $\|\phi_z\|_\infty=h_z$, we have 
\[
 M=\sup_{z\in Z_K}\frac{h_z}{w(z)}<\infty.
\]
To obtain a compact parameter space, we describe each normalized atom $\phi_z/w(z)$ by its direction $\xi$, cap height $h_z$, and maximum value $h_z/w(z)$ on $K$. Let $P$ be the closure of
\[
 \{(\xi,h_z,h_z/w(z)):z=(\xi,t)\in Z_K\}
\]
in the compact set $S^{d-1}\times[0,\operatorname{diam}K]\times[0,M]$. For $s=(\xi,h,a)\in P$, define
\[
 k_s(x)=
 \begin{cases}
 \displaystyle
 a\,\frac{(\xi\cdot x-H_K(\xi)+h)_+}{h},
   & h>0,\\[4pt]
 0,& h=0,
 \end{cases}
 \qquad x\in K.
\]
For $z=(\xi,t)\in Z_K$, taking $s=(\xi,h_z,h_z/w(z))$ gives $k_s=\phi_z/w(z)$.

For $s=(\xi,h,a)\in P$ with $h>0$, choose $z_n=(\xi_n,t_n)\in Z_K$ such that $(\xi_n,h_{z_n},h_{z_n}/w(z_n))\to s$. Since $h_{z_n}\to h>0$ and $H_K$ is continuous, we have
\[
 \left\|\frac{\phi_{z_n}}{w(z_n)}-k_s\right\|_\infty
 \longrightarrow0.
\]
Thus $k_s$ is a uniform limit of normalized atoms. For $h=0$, we have $k_s=0$. By the definition of $\mathbb B_w^{(\infty)}(K)$ and the uniform bound on the normalized atoms,
\[
 k_s\in\mathbb B_w^{(\infty)}(K),
 \qquad
 \|k_s\|_\infty\le M,
 \qquad\forall s\in P.
\]

Recall that $(F_m)$ is the sequence of finite networks chosen to approximate $f$ in $L^p(\mu_K)$. For each $m$, choose a representation $F_m=\sum_i c_i\phi_{z_i}$ with $\sum_i|c_i|w(z_i)\le1$. For $z_i=(\xi_i,t_i)$, set $s_i=(\xi_i,h_{z_i},h_{z_i}/w(z_i))$ and $\nu_m=\sum_i c_iw(z_i)\,\delta_{s_i}$, where $\delta_{s_i}$ denotes the unit point mass at $s_i$. Since $k_{s_i}=\phi_{z_i}/w(z_i)$,
\[
 F_m(x)=\int_P k_s(x)\,d\nu_m(s),
 \qquad
 |\nu_m|(P)\le\sum_i|c_i|w(z_i)\le1.
\]

By the Riesz representation theorem, every $\Lambda\in C(P)^*$ is represented by a unique finite signed Radon measure $\rho$ on $P$, with
\[
 \Lambda(u)=\int_P u(s)\,d\rho(s),
 \ \forall u\in C(P),
 \qquad
 \|\Lambda\|_{C(P)^*}=|\rho|(P).
\]
Under this identification, $\|\nu_m\|_{C(P)^*}=|\nu_m|(P)\le1$. Since $P$ is compact and metrizable, $C(P)$ is separable. The Banach--Alaoglu theorem and separability imply that the closed unit ball of $C(P)^*$ is compact and metrizable in the weak-* topology, hence sequentially compact. We may therefore pass to a subsequence, still indexed by $m$, such that
\[
 \nu_m\rightharpoonup^*\nu
 \quad\text{in }C(P)^*,
 \qquad |\nu|(P)\le1,
\]
where $\nu$ is a finite signed Radon measure on $P$. The function $(s,x)\mapsto k_s(x)$ is Borel on $P\times K$, since it is continuous on $\{h>0\}\times K$ and zero on $\{h=0\}\times K$. Since $|k_s(x)|\le M$ and $|\nu|(P)\le1$, the integral
\[
 g(x)=\int_P k_s(x)\,d\nu(s)
\]
is well defined for every $x\in K$ and satisfies $|g(x)|\le M$.

We first show that $g=f$ almost everywhere. To deduce $F_m(x)\to g(x)$ from $\nu_m\rightharpoonup^*\nu$, we need the function $s\mapsto k_s(x)$ to be continuous on $P$. For $x\in\operatorname{int}K$, choose $r>0$ such that $\overline B(x,r)\subset K$. Then
\[
 H_K(\xi)-\xi\cdot x\ge r,
 \qquad\forall\xi\in S^{d-1},
\]
so $k_s(x)=0$ whenever $0\le h<r$. Note that the map $s\mapsto k_s(x)$ is continuous at points with $h>0$, and at points with $h=0$ because it vanishes whenever $0\le h<r$. Thus it is continuous on $P$. The weak-* convergence $\nu_m\rightharpoonup^*\nu$ in $C(P)^*$ therefore gives $F_m(x)\to g(x)$ for every $x\in\operatorname{int}K$. Since $\partial K$ has zero volume and $|F_m|,|g|\le M$, dominated convergence yields $F_m\to g$ in $L^p(\mu_K)$. Hence $f=g$ almost everywhere.

It remains to show that $g$ belongs to the uniform closure of $\mathcal A_w(K)$. For $\delta>0$, set
\[
 P_\delta=\{s\in P:h\ge\delta\},
 \qquad
 g_\delta(x)=\int_{P_\delta}k_s(x)\,d\nu(s).
\]
The map $s\mapsto k_s$ is continuous from the compact parameter set $P_\delta$ into $C(K)$ equipped with the uniform norm. By compactness of the image of this map, for every $\epsilon>0$ we can choose a finite Borel partition $\{A_i\}$ of $P_\delta$ and points $s_i\in A_i$ such that
\[
 \sup_{s\in A_i}\|k_s-k_{s_i}\|_\infty\le\epsilon.
\]
It follows that
\[
 \sup_{x\in K}
 \left|
 g_\delta(x)-\sum_i\nu(A_i)k_{s_i}(x)
 \right|
 \le\epsilon|\nu|(P_\delta)\le\epsilon.
\]
Moreover, $\sum_i|\nu(A_i)|\le|\nu|(P_\delta)\le1$. By symmetry, both $k_{s_i}$ and $-k_{s_i}$ belong to $\mathbb B_w^{(\infty)}(K)$. Since $\sum_i|\nu(A_i)|\le1$, convexity gives
\[
 \sum_i\nu(A_i)k_{s_i}\in\mathbb B_w^{(\infty)}(K).
\]
Taking their uniform limit gives $g_\delta\in\mathbb B_w^{(\infty)}(K)$.

Finally, since $\|k_s\|_\infty\le M$ for all $s\in P$ and $k_s=0$ when $h=0$, we have
\[
 \sup_{x\in K}|g(x)-g_\delta(x)|
 \le M|\nu|(\{s\in P:0<h<\delta\})
 \longrightarrow0, 
 \qquad\text{as }\delta\downarrow0.
\]
The convergence follows because $|\nu|$ is finite and the sets $\{0<h<\delta\}$ decrease to the empty set. Thus $g$ is a uniform limit of functions in $\mathbb B_w^{(\infty)}(K)$. It is therefore continuous and, by closedness, belongs to $\mathbb B_w^{(\infty)}(K)$. Since $g=f$ almost everywhere, it is the required continuous representative of $f$, proving the reverse inclusion.

The equality of the unit balls gives equality of the spaces and their variation norms by \eqref{eq:space-definition} and \eqref{eq:variation-norm}.
\end{proof}

\subsection{Epsilon Nets for Finite Families of Sets}
\label{app:epsilon-net}

The following lemma gives conditions for choosing a small set of points that intersects every set of measure at least $\epsilon$ in a given finite family. This is a version of the epsilon-net theorem of Haussler and Welzl~\cite{HausslerWelzl1987} for finite families. We give a sampling proof that applies directly to normalized surface measure on the sphere.

\begin{lemma}
\label{rand:epsilon-net}
Let $(X,\mathcal F,\sigma)$ be a probability space, and let $\mathcal R\subset\mathcal F$ be a finite family. Suppose that, for every $n\ge1$ and every list $x_1,\ldots,x_n\in X$, with repetitions allowed, the index sets
\[
 \{i\in\{1,\ldots,n\}:x_i\in R\},
 \qquad R\in\mathcal R,
\]
take at most $A(n+1)^D$ distinct values, where $A,D$ are fixed constants. These index sets are called the traces of $\mathcal R$ on the list. Then, for every $0<\epsilon\le1/2$, there is a finite set $T\subset X$ such that
\[
 \#T\le C_{A,D}\epsilon^{-1}\log(2/\epsilon)
\]
and $T\cap R\ne\varnothing$ for every $R\in\mathcal R$ with $\sigma(R)\ge\epsilon$. The constant $C_{A,D}$ depends only on $A,D$ and is independent of the cardinality of $\mathcal R$.
\end{lemma}

\begin{proof}
If no member of $\mathcal R$ has measure at least $\epsilon$, take $T=\varnothing$. Otherwise, by increasing $A$ and $D$ in the upper bound $A(n+1)^D$ on the number of distinct traces if necessary, we may assume that $A\ge1$ and $D\ge0$.

Let $N$ be a positive integer with $N\epsilon\ge8$, to be specified below, and take mutually independent random points
\[
 X_1,\ldots,X_N,\qquad Y_1,\ldots,Y_N
\]
with distribution $\sigma$. The first sample will provide the desired set of points. We say that it misses a set $R\in\mathcal R$ if none of $X_1,\ldots,X_N$ belongs to $R$. As shown below, if the first sample misses such a set $R$ with $\sigma(R)\ge\epsilon$, then, conditional on the first sample, the independent second sample contains at least $\epsilon N/2$ points in $R$ with probability at least $1/2$. This observation will allow us to estimate the probability that the first sample misses some set of measure at least $\epsilon$ by counting the distinct traces on the combined list of $2N$ points.

Let $E$ be the event that some $R\in\mathcal R$ with $\sigma(R)\ge\epsilon$ contains none of $X_1,\ldots,X_N$. Let $B$ be the event that some $R\in\mathcal R$ contains none of $X_1,\ldots,X_N$ and contains $Y_j$ for at least $\epsilon N/2$ indices $j\in\{1,\ldots,N\}$. Unlike the definition of $E$, the definition of $B$ imposes no lower bound on $\sigma(R)$. We first show that
\[
 \mathbb P(E)\le2\mathbb P(B).
\]

Fix an enumeration of $\mathcal R$. Whenever $E$ occurs, let $R_*$ be the first member in this enumeration that has measure at least $\epsilon$ and is missed by the first sample. Conditional on any realization of the first sample for which $E$ occurs, the set $R_*$ is fixed. Put $p=\sigma(R_*)\ge\epsilon$. Independence of the second sample implies that
\[
 W=\#\{i:Y_i\in R_*\}
\]
has a binomial distribution with mean $Np$ and variance $Np(1-p)\le Np$. Since $\epsilon N/2\le Np/2$, Chebyshev's inequality gives
\[
 \begin{aligned}
 \mathbb P\left(W<\frac{\epsilon N}{2}\,\middle|\,X_1,\ldots,X_N\right)
 &\le
 \mathbb P\left(|W-Np|\ge\frac{Np}{2}\,\middle|\,X_1,\ldots,X_N\right)\\
 &
 \le\frac{\mathbb{E}[|W-Np|^2]}{\left|\frac{Np}{2}\right|^2} 
 \le\frac{4}{Np}
 \le\frac12.
 \end{aligned}
\]
Whenever $E$ occurs, the selected set $R_*$ contains none of $X_1,\ldots,X_N$. Conditional on the first sample, the preceding estimate shows that at least $\epsilon N/2$ of $Y_1,\ldots,Y_N$ lie in $R_*$ with probability at least $1/2$. If $\mathbb P(E)=0$, the desired inequality is immediate. Otherwise, averaging over realizations of the first sample for which $E$ occurs gives
\[
 \mathbb P(B\mid E)\ge\frac12.
\]
Hence
\[
 \mathbb P(B)\ge\mathbb P(B\cap E)
 =\mathbb P(B\mid E)\mathbb P(E)
 \ge\frac12\mathbb P(E).
\]

For any list of $2N$ points, record for each $R\in\mathcal R$ the indices of the points that lie in $R$. By assumption, at most $A(2N+1)^D$ different index sets can arise as $R$ varies. We will use this bound to estimate $\mathbb P(B)$.

For this purpose, we equivalently generate the two samples as follows. Draw $2N$ independent points $Z_1,\ldots,Z_{2N}$ from $\sigma$, and independently choose $N$ of their indices uniformly at random. Assign the points with the chosen indices to the first sample and the remaining points to the second. Every fixed choice of indices gives two independent samples of size $N$ from $\sigma$, so randomizing the choice preserves the original joint distribution.

Now condition on $Z_1,\ldots,Z_{2N}$, so that all sampled points are fixed. Let $I$ denote the set of $N$ indices assigned to the first sample. Only the choice of $I$ remains random, and all $\binom{2N}{N}$ choices are equally likely because the indices were chosen independently of the sampled points.

For each $R\in\mathcal R$, define
\[
 J_R=\{i\in\{1,\ldots,2N\}:Z_i\in R\}.
\]
These index sets are now fixed, and by assumption there are at most $A(2N+1)^D$ distinct possibilities as $R$ varies. Since the second sample uses the indices outside $I$, the event $B$ occurs precisely when some $J_R$ satisfies $\#J_R\ge\frac{\epsilon N}{2}$ and $I\cap J_R=\varnothing$.

If no $J_R$ has at least $\epsilon N/2$ elements, then $\mathbb P(B\mid Z_1,\ldots,Z_{2N})=0$. Otherwise, fix any such index set $J$ and write $k=\#J\ge\epsilon N/2$. We estimate the probability that the randomly chosen first-sample index set $I$ avoids $J$. If $k>N$, this is impossible because the second sample has only $N$ indices. If $k\le N$, all $N$ indices of $I$ must be chosen from the $2N-k$ indices outside $J$. There are exactly $\binom{2N-k}{N}$ such choices, out of $\binom{2N}{N}$ equally likely choices in total. Therefore,
\[
 \mathbb P\bigl(I\cap J=\varnothing\mid Z_1,\ldots,Z_{2N}\bigr)
 =\frac{\binom{2N-k}{N}}{\binom{2N}{N}}
 =\prod_{j=0}^{k-1}\frac{N-j}{2N-j}
 \le2^{-k}
 \le2^{-\epsilon N/2}.
\]

Taking the union bound over the distinct index sets $J_R$ with at least $\epsilon N/2$ elements gives
\[
 \mathbb P(B\mid Z_1,\ldots,Z_{2N})
 \le A(2N+1)^D2^{-\epsilon N/2}.
\]
Averaging over $Z_1,\ldots,Z_{2N}$ and using the previously established inequality $\mathbb P(E)\le2\mathbb P(B)$ yields
\[
 \mathbb P(E)
 \le2\mathbb P(B)
 \le2A(2N+1)^D2^{-\epsilon N/2}.
\]

It remains to choose $N$ so that this bound is less than one, uniformly for $0<\epsilon\le1/2$. Set
\[
 L=\log(2/\epsilon),
 \qquad
 N=\left\lceil K\epsilon^{-1}L\right\rceil,
\]
where $K\ge1$ will depend only on $A,D$. Since $L\ge\log4$ and $\epsilon^{-1}L\ge1$, we have
\[
 \epsilon N\ge KL,
 \qquad
 2N+1\le(2K+3)\epsilon^{-1}L.
\]
Using $\log(1/\epsilon)\le L$ and $\log L\le L$, we obtain
\[
 \log(2N+1)\le\log(2K+3)+2L.
\]
Consequently,
\[
 \begin{aligned}
 \log\left(2A(2N+1)^D2^{-\epsilon N/2}\right)
 &\le \log(2A)+D\log(2K+3)
       +\left(2D-\frac{K\log2}{2}\right)L.
 \end{aligned}
\]
For $K$ sufficiently large depending only on $A,D$, the coefficient of $L$ is negative and the expression on the right is negative at $L=\log4$, hence for every $L\ge\log4$. Such a choice is possible because the negative term grows linearly in $K$, whereas $D\log(2K+3)$ grows only logarithmically. We also choose $K$ large enough that $K\log4\ge8$, which ensures $N\epsilon\ge8$.

With this choice, $\mathbb P(E)<1$. Choose a realization outside $E$ and let $T$ be the set of distinct points in its first sample. Removing repeated points does not affect which members of $\mathcal R$ are met, and
\[
 \#T\le N
 \le(K+1)\epsilon^{-1}\log(2/\epsilon).
\]
Taking $C_{A,D}=K+1$ proves the claim.
\end{proof}

In the proof of Lemma~\ref{rand:cap-cover}, the map
\[
 \Phi(x)=\bigl((x_i)_{1\le i\le m},
              (x_ix_j)_{1\le i\le j\le m}\bigr)
 \in\mathbb R^{M_m},
 \qquad
 M_m=m+\frac{m(m+1)}2,
\]
turns the linear and quadratic inequalities defining $G_C$ into two halfspace conditions in $\mathbb R^{M_m}$. For a fixed list of points, different halfspaces may contain different subsets of the list. The following lemma bounds the number of subsets that can arise.

\begin{lemma}
\label{rand:halfspace-count}
Fix an integer $d\ge1$, and let $p_1,\ldots,p_n\in\mathbb R^d$ be a list of $n\ge1$ points, with repetitions allowed. For each open or closed affine halfspace $H\subset\mathbb R^d$, consider the index set
\[
 \{i\in\{1,\ldots,n\}:p_i\in H\}.
\]
As $H$ varies over all open or closed affine halfspaces in $\mathbb R^d$, at most $C_d(n+1)^{d+1}$ distinct index sets occur, where $C_d$ depends only on $d$.
\end{lemma}

\begin{proof}
Fix the list $p_1,\ldots,p_n$. Write the halfspace as
\[
 \{x:a\cdot x>b\}
 \quad\text{or}\quad
 \{x:a\cdot x\ge b\},
 \qquad a\in\mathbb R^d\setminus\{0\},\quad b\in\mathbb R.
\]
Choose $\eta>0$ smaller than every nonzero value of $|a\cdot p_i-b|$. If all these values are zero, choose any $\eta>0$. For a closed halfspace, replace $b$ by $b-\eta$. For an open halfspace, replace $b$ by $b+\eta$. The choice of $\eta$ ensures that points strictly on either side of the original boundary remain on the same side. Points on the original boundary remain included in the closed case and excluded in the open case. Thus the selected index set is unchanged, and none of the listed points lies on the shifted boundary hyperplane. 

We now keep the points $p_i$ fixed and allow the coefficients $(a,b)$ to vary. For each $i$, the equation $a\cdot p_i-b=0$ defines a hyperplane
\[
 L_i=\{(a,b)\in\mathbb R^{d+1}:a\cdot p_i-b=0\}
\]
in the space of coefficients. The boundary adjustment above ensures that every index set obtained from an open or closed halfspace can also be obtained with coefficients $(a,b)$ outside $\bigcup_{i=1}^n L_i$. On each connected component of $\mathbb R^{d+1}\setminus\bigcup_{i=1}^n L_i$, the signs of all expressions $a\cdot p_i-b$ are constant. Each component therefore corresponds to a single index set, so it suffices to bound the number of these connected components.

A collection of at most $n$ affine hyperplanes divides $\mathbb R^q$ into at most
\[
 \sum_{j=0}^q\binom nj
 \le(q+1)(n+1)^q
\]
regions~\cite[Section~6.1]{Matousek2002}. Applying this bound with $q=d+1$ shows that the coefficient space has at most $(d+2)(n+1)^{d+1}$ regions. Hence the halfspaces select at most this many distinct index sets. Repeated points give repeated hyperplanes $L_i$, and repetitions do not increase the number of regions. The claim follows with $C_d=d+2$.
\end{proof}

For the application to Lemma~\ref{rand:cap-cover}, take $d=M_m$ and apply Lemma~\ref{rand:halfspace-count} to the fixed points $\Phi(\theta_1),\ldots,\Phi(\theta_n)$. Each of the two halfspace conditions selects at most $C_{M_m}(n+1)^{M_m+1}$ different index sets. The indices satisfying both conditions form the intersection of one index set from each condition. The number of possible intersections is at most the number of such pairs, so the sets $G_C$ select at most
\[
 C_{M_m}^2(n+1)^{2(M_m+1)}
 =C_m(n+1)^{2(M_m+1)}
\]
different index sets, where $C_m$ depends only on $m$.

\bibliographystyle{plain}
\bibliography{references}

\end{document}